\documentclass[11pt,letter]{article}

\usepackage[hyphens]{url}
\usepackage{amsmath}
\usepackage{amssymb}
\usepackage{mathrsfs}
\usepackage{graphicx} 
\usepackage{setspace}
\usepackage{url}
\usepackage{enumitem} 
\setlist[enumerate]{itemsep=0mm}
\usepackage{fullpage}
\usepackage[hidelinks]{hyperref} 
\usepackage{stmaryrd}  
\usepackage{tikz}
\usepackage{tikz-qtree}
\usepackage{tikz-qtree-compat} 
\usepackage{textcomp} 
\usepackage[authoryear,round]{natbib} 
\usepackage{bib entry} 
\bibpunct{(}{)}{;}{a}{}{,} 
\nobibliography* 
\usepackage{longtable} 
\usepackage{gb4e} 

\usepackage{todo} 

\begin{document}

\title{Character Entropy in Modern and Historical Texts: Comparison Metrics for an Undeciphered Manuscript} 
\author{Luke Lindemann and Claire Bowern}
\date{\today}
\maketitle

\begin{small}
Note: This is an updated version of the article that we uploaded on October 27, 2020. For this update we developed an improved method for extracting language text from Wikipedia, removing metadata and wikicode, and we have rebuilt our corpus based on current wikipedia dumps. The new methodology is described in Section~\ref{wikicorpus}, and we have updated the statistics and graphs in Section~\ref{sec:h2} and Appendices B and D. None of our results have changed substantially, suggesting that character-level entropy was not greatly affected by the inclusion of stray metadata. However, our forthcoming work will use these corpora for word-level statistics including measures of repetition and type token ratio, for which metadata would have a much greater effect on results. We also made a minor alteration to the Maximal Voynich transcription system as described in Section~\ref{vts}. This also has a negligible effect on character entropy. The statistics in Section~\ref{sec:h2} and Appendix A have been updated accordingly. 
\end{small}

\bigskip

\begin{abstract}

This paper outlines the creation of three corpora for multilingual comparison and analysis of the Voynich manuscript: a corpus of Voynich texts partitioned by Currier language, scribal hand, and transcription system, a corpus of 311 language samples compiled from Wikipedia, and a corpus of eighteen transcribed historical texts in eight languages. These corpora will be utilized in subsequent work by the Voynich Working Group at Yale University. 

We demonstrate the utility of these corpora for studying characteristics of the Voynich script and language, with an analysis of conditional character entropy in Voynichese. We discuss the interaction between character entropy and language, script size and type, glyph compositionality, scribal conventions and abbreviations, positional character variants, and bigram frequency. 

This analysis characterizes the interaction between script compositionality, character size, and predictability. We show that substantial manipulations of glyph composition are not sufficient to align conditional entropy levels with natural languages. The unusually predictable nature of the Voynichese script is not attributable to a particular script or transcription system, underlying language, or substitution cipher. Voynichese is distinct from every comparison text in our corpora because character placement is highly constrained within the word, and this may indicate the loss of phonemic distinctions from the underlying language.

\end{abstract} 

Corpus materials and code are available from \texttt{github.com/chirila/Voynich-public}.
\pagebreak

\tableofcontents
\pagebreak

\section{Introduction} 

\begin{figure}[h!]
\begin{centering}
\includegraphics[scale=0.75,width=\linewidth]{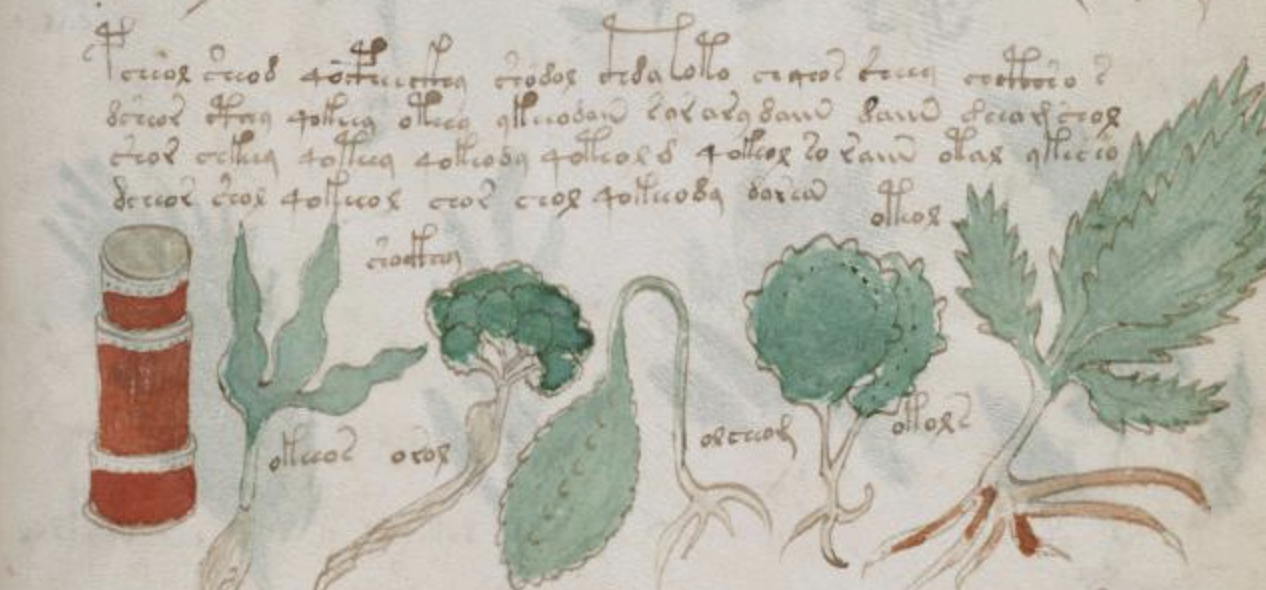}
\caption{A paragraph of text and labelled figures from the ``Recipes'' section of the Voynich Manuscript (f100r).}
\end{centering}
\label{voyex}
\end{figure}

The Voynich Manuscript (e.g.~Figure \ref{voyex}) is an early 15th Century illustrated manuscript written by multiple unknown scribes \citep{davis20} in an unknown cipher or language. It contains about 38,000 words of text. The Voynich alphabet, which is not found in any other known work, has resisted nearly 110 years of modern attempts at decipherment (see \citealt{bowernlindemann20} and \url{http://www.voynich.nu} for overviews). This is despite the fact that there is clear evidence of language-like structure in the text, at least at the paragraph level  \citep{reddy2011we,amancio2013probing,landini2001evidence}.\footnote{Thanks to members of the ``Mystery of the Voynich Manuscript'' class at Yale for discussion of some of these points. Division of labor: LL and CB planned the analyses; LL compiled the corpora and wrote the scripts; LL and CB analyzed the data; LL wrote the paper with input from CB.}

The major unsolved question of the Voynich text is whether it represents meaningful language.\footnote{In previous decades, there was the separate question of whether the manuscript is a modern hoax, i.e. a 20th century forgery of a medieval manuscript. This has been fairly decisively disproven by chemical analysis, though see \citet{barlow1986} for some earlier discussion that predates the results from carbon dating.} It could be a medieval hoax that is designed to look like an esoteric alchemical text, in which no sequences of letters correspond to any meaningful words or concepts (see \citealt{rugg2004elegant,TimmSchinner2020}, amongst others). If so, the creators did an incredible job of imitating the patterns of an authentic language text considering what was known about the structure of language at the time. In so doing, they must have modeled their fake language after a real language that they were familiar with, imbuing it with familiar, language-like patterns. If this is the case, it may still be possible to take clues from the structure of Voynichese to pinpoint a language or region of origin.

We find it more likely that Voynichese does represent meaningful language, and this opens the possibility that Voynichese may ultimately be deciphered. It is possible that the text was created to encode meaning, but the nature of the encipherment obscured it in such a way that the original meaning is permanently irrecoverable. Even if this is the case, we may be able to glean information about language and content with a careful analysis of Voynichese structure. Note that others, such as \citet{TimmSchinner2020}, argue strongly that the Voynich manuscript does not encode meaningful text, and so cannot be compared with natural language.\footnote{\citealt{TimmSchinner2021} make this point especially forcefully, even going as far as accusing us of a lack of scientific rigor for not finding their arguments convincing. While the aim of this paper is to make  language comparisons across natural and constructed languages, rather than to make arguments about any particular theory, we reiterate a point we have made several times elsewhere (including in the review article that they criticize): that Voynichese appears unnatural only below the word level. At the level of page and paragraph, Voynichese is comparable to natural language and structured text. This finding must be taken into account when proposing that Voynichese does not encode meaningful text, as most methods of creating meaningless text will not exhibit this property. See \citet{Zandbergen2021} for discussion of the Cardan Grille cypher method, which could be used to create meaningless text or to encode natural language. }

The goal of this project is to analyze the structure and patterning of the Voynich manuscript, and to compare it to known texts. By comparing Voynichese to known languages and texts, we reduce the set of possible hypotheses about language, origin, and the question of meaningfulness. This paper describes the creation of text corpora for conducting experiments on the Voynich text. Note that our aim is not to advance a strong claim about exclusive identification (of the form ``Voynichese is Hebrew'' or ``Voynichese is Occitan'') but rather to explore the relationships between the morphological and phonological profiles of Voynichese with a typologically broad range of natural and constructed languages.\footnote{This is why we include samples of languages which are extremely unlikely to underlie the Voynich manuscript, including Indigenous languages of the Americas and modern constructed languages. We include them not because we think they are likely Voynichese candidates (far from it), but rather so that we can better study the interaction of morphological and lexical typology and the statistical profiles of different languages.} 

We require as many examples of languages and scripts as possible in order to understand the range of possibilities in the structure of texts. We would also like to be able to classify these texts by language and language family in order to see whether closely related languages share affinities of structure, allowing us to narrow down the range of possible languages (or at least better understand possible encryption processes). For this reason, our first comparison corpus consists of Wikipedia articles written in 311 languages, representing thirty-nine language families.  

The Voynich manuscript is also the product of a particular historical context. This includes the medieval scribal traditions which produced it, as well as the herbalogical, alchemical, and astrological knowledge which informed its content. To take a particular relevant example, medieval scribes made much more frequent use of abbreviations than we find in modern writing. We therefore include a second comparison corpus of historical manuscripts in English, Georgian, Hebrew, Icelandic, Italian, Latin, Persian, and Spanish. 

This paper consists of three sections. In Section~\ref{sec:structure}, we give an overview of the structure and content of the Voynich manuscript, focusing on what is known about the text. We give a brief description of each section and discuss the evidence for multiple Voynich languages and scribal hands.

In Section \ref{sec:corpora}, we outline in detail the process of creating the three corpora used for this project: the Voynich Corpus, the Wikipedia Corpus, and the Historical Corpus. We discuss the issue of transcription in the Voynich text, and define the three transcriptions we use: Maximal, Maximal Simplified, and Minimal. 

In Section \ref{sec:h2}, we demonstrate the utility of the corpora by comparing character-level properties of the Voynich manuscript with the Wikipedia and historical text samples. We discuss in detail the unusually low conditional character entropy of Voynichese, and compare the effect that language, script, transcription system, usage of abbreviations, and typographical convention has on this value.

\section{Structure and Content of the Voynich Manuscript}\label{sec:structure}

The Voynich manuscript contains 102 folios in its current form. There is evidence that some of the pages have been removed and rearranged from their original ordering. There are Arabic numerals 1-116 in the top right corner of each recto folio.\footnote{The fourteen missing folios are f12, f59, f60, f61, f62, f63, f64, f74, f91, f92, f97, f98, f109, and f110.} These numerals were probably not written by the original authors, but were added at a later date. Ten of the folios fold out to reveal additional diagrams and text, the largest of which is the ``Rose,'' a complex six-page foldout. See \citet{davis20} for a discussion of manuscript hands and foliation.

While the Voynich document does not appear to have section or chapter titles, it can be divided into five sections based upon the drawings and figures in each section:\footnote{For ease of reference, we have labelled sequentially coherent Section boundaries based on the current order of the manuscript. Therefore, those pages which consist of only text (and are therefore not obviously classifiable) are classified as part of the section in which they appear. Furthermore, the isolated Herbal pages that are found in the Cosmology and Recipe sections are classified as part of the Cosmology and Recipe sections respectively. This differs from the section coding schema employed by the interlinear gloss file available at \url{http://www.voynich.nu}. It is very likely that some pages in the manuscript are now in a different order than the order in which they were first composed.}

\begin{enumerate}

\item The Herbal section is the first and largest section, taking up approximately half of the entire manuscript. Each folio contains an illustration of an herb or flower. One or more paragraphs of text are written around the illustration. There are no labels on the illustrations themselves in this section.

\item The Cosmological section consists of circular diagrams and charts that appear to be astrological in nature. Most of them include drawings of stars and stylized suns, with text written in spirals and copious labels. A few of the characters are recognizable medieval astrological symbols.\footnote{The number of divisions or points on many of these charts also suggest astrological concepts: twelve representing the houses of the Zodiac, seven representing the planets, four representing the elements (also humors, directions, qualities, or triplicities), eight representing the monastic hours of the day.} There is also a twelve-page sequence of Zodiac illustrations within concentric circles of text and pictures of women with labels. In most cases, there are exactly thirty women per illustration. The Zodiac signs are correctly ordered and have been labelled at the center with corresponding month names in the Occitan dialect of French (which is probably a later addition and should not be mistaken for Voynichese).\footnote{Though it takes up twelve folios, the Zodiac is incomplete because it only depicts ten out of the twelve signs. Capricorn and Aquarius, the first and last signs, are missing. Ares and Taurus are each depicted twice on separate folios. Their charts depict only fifteen women, which suggests that each folio represents a half-month. The only chart with neither thirty nor fifteen women is Gemini, which has twenty-nine.} There is less running paragraph text in this section, but there are many labels and text written in a circular pattern.

\item The Balneological section contains pictures of what appear to be stylized women bathing in large basins and interconnected ornamental tubes. Each folio contains multiple paragraphs of running text, and many of the women are labelled. This is followed by the six-page Rose foldout, which on one side depicts nine interconnected circular diagrams, many centered with suns and containing stars or tubes. The other side contains text and more circular diagrams. 

\item The Recipes section is distinguished by pages with paragraphs of text separated by assortments of labelled herbs, leaves, or roots. To the left of the paragraphs there are what appear to be ornate jars. In between these pages of ``recipes'' there is a central section of herbals in the same style as the Herbal section. 

\item The Stars section contains no illustrations and consists of densely packed short paragraphs of text. Each page contains ten to twenty paragraphs which are marked on the lefthand side by a seven-pointed star symbol. 
 
\end{enumerate}

\begin{figure}[t!]
\centering
\includegraphics[scale=0.16,width=\linewidth]{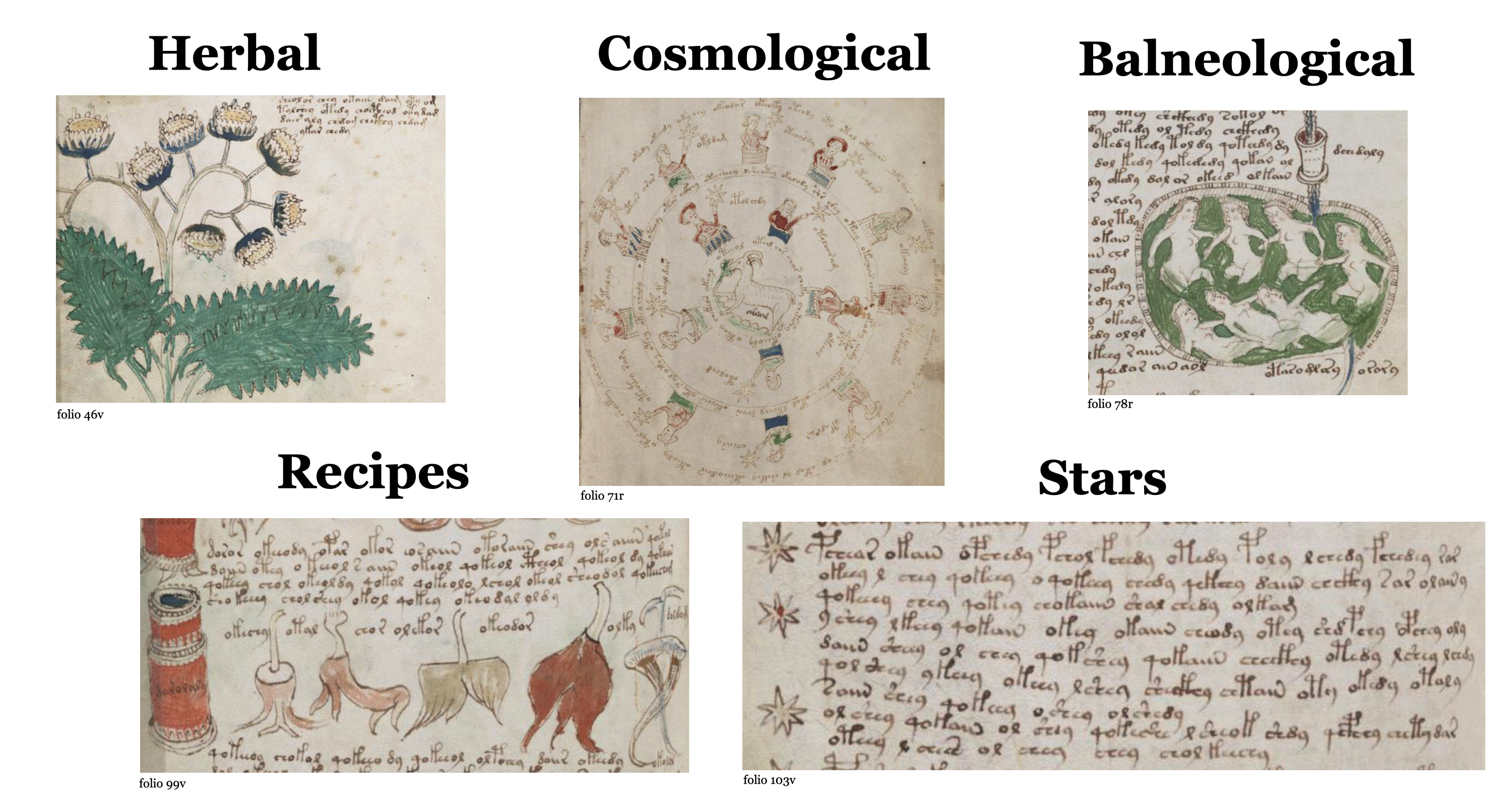}\centering
\caption{Selected examples from each section of the Voynich Manuscript}
\label{voysections}

\smallskip

\includegraphics[scale=0.25,width=\linewidth]{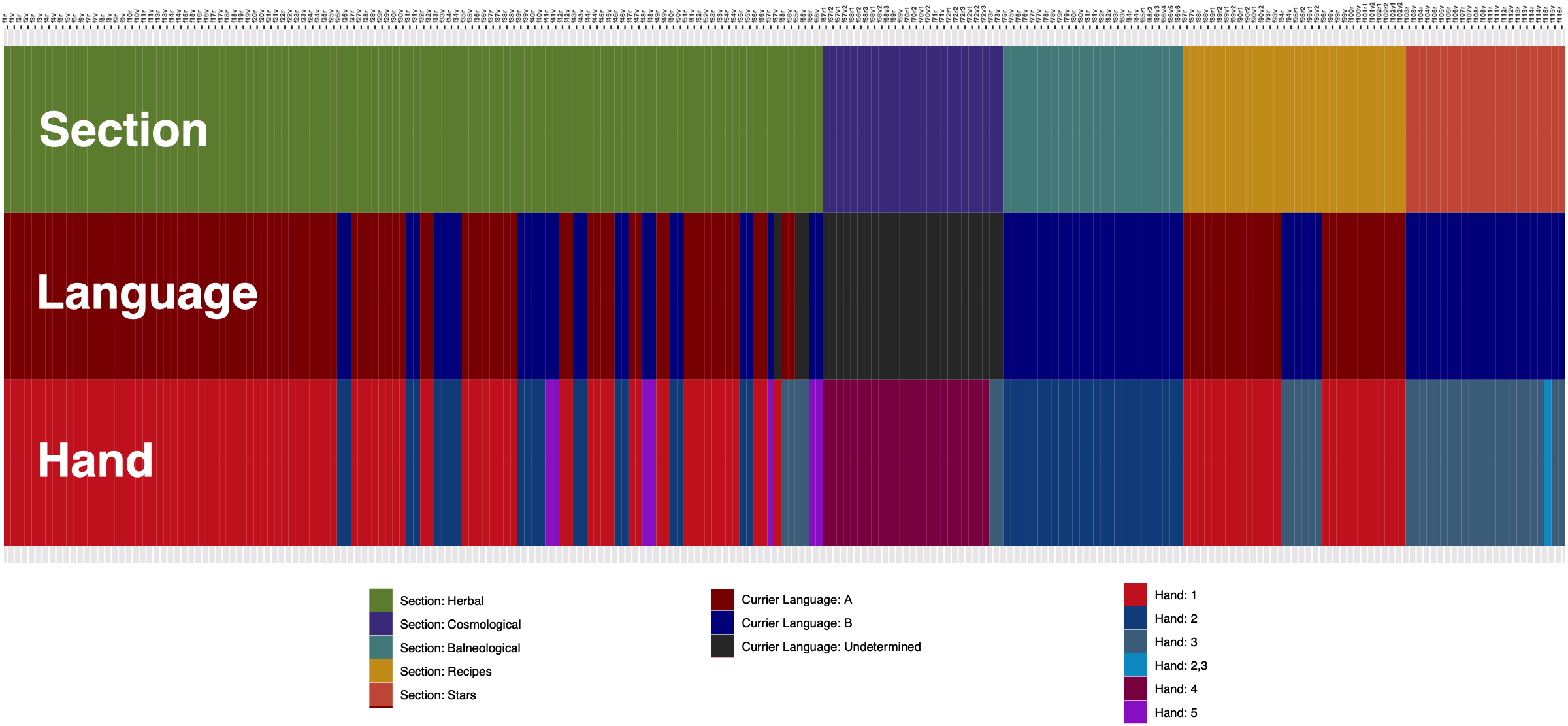}\centering
\caption{Map of Sections, Languages (Currier 1976) and Scribal Hands (Davis 2020)}
\label{voymap}
\end{figure}

There is evidence that more than one scribe produced the text. \citet{currier1976papers} noted the existence of multiple scribal hands, and he also classified pages of the text into two different ``languages'' (Currier Language A and B) based upon consistent and marked differences in the frequency of certain words and glyph combinations. The usage of the term ``language'' is misleading, because Language A and Language B do not necessarily represent different natural languages. There are substantial similarities of structure and vocabulary. They may represent different dialects of the same language, or they may represent the same dialect but use a slightly different encoding scheme. With a small number exceptions, every folio is written in only one Language and Scribal Hand, and each Scribal Hand employs only one language. This implies to us that the scribes who made the text were also its authors. If they were copying a previous work, we should not expect to find such a close correlation between the language and scribal hand.

The first half of the Herbal Section is written in Language A, and the second half alternates between Languages A and B. The Balneological and Stars sections are written entirely in B. The ``recipes'' of the Recipes section are all written in A, while the ``herbals'' in the Recipes section are written in A or B (suggesting that it was originally part of the Herbals section). The Cosmological section, which contains mostly labelled diagrams rather than running text, was left unclassified by Currier, although it most closely resembles Language B.

The recent analysis of \citet{davis20} demonstrates evidence for five different scribal hands based on variations in the formation of several glyphs. She finds that Language A is written entirely by Hand 1 (with the exception of 58r), while the other hands write in Language B. Hand 2 is found in the second half of the Herbal section, the entire Balneological section, and thirty-three lines on a folio in the Stars section (115r) which is shared with Hand 3. Hand 3 is found at the end of the Herbal and Cosmological sections, the ``herbal'' portion of the Recipes, and every folio of the Stars section. The Cosmological section is written almost entirely in Hand 4. Hand 5 is found only in the second half of the Herbal section. 

The amount of text written by Hands 1, 2, and 3 is approximately equal: 10-12 thousand words each. Hands 4 and 5 are found mostly on diagram labels rather than running text, and account for less than four thousand words between them. Overall, 87\% of the Voynich text is written in paragraphs and 13\% consists of labels on diagrams or drawings.

\section{Description of the Corpora}\label{sec:corpora}

The following sections describe the corpus materials used in the current study.

\subsection{The Voynich Corpus}

The Voynich corpus consists of digitally transcribed copies of the manuscript itself (see Section~\ref{vdp} below for more details on the transcription used). We also created separate documents for each Voynich Language, Scribal Hand, and separated running text from the text found in labels and diagrams. The Voynich corpus used for this project consists of the following documents:

\begin{enumerate}

	\item \textbf{Full Voynich}: The entire Voynich text, including running text, labels, and diagrams. 
	\item \textbf{Full Voynich Text}: The Voynich text written in paragraphs, without text in labels and diagrams.
	\item \textbf{Voynich A}: Voynich written in Currier Language A, including running text, labels, and diagrams.
	\item \textbf{Voynich A Text}: Voynich written in Currier Language A, without text in labels and diagrams.
	\item \textbf{Voynich B}: Voynich written in Currier Language B, including running text, labels, and diagrams.
	\item \textbf{Voynich B Text}: Voynich written in Currier Language B, without text in labels and diagrams.
	\item \textbf{Voynich 1}: Voynich written in Hand 1, including running text, labels, and diagrams.
	\item \textbf{Voynich 1 Text}: Voynich written in Hand 1, without text in labels and diagrams.
	\item \textbf{Voynich 2}: Voynich written in Hand 2, including running text, labels, and diagrams.
	\item \textbf{Voynich 2 Text}: Voynich written in Hand 2, without text in labels and diagrams.
	\item \textbf{Voynich 3}: Voynich written in Hand 3, including running text, labels, and diagrams.
	\item \textbf{Voynich 3 Text}: Voynich written in Hand 3, without text in labels and diagrams.
	\item \textbf{Voynich 4}: Voynich written in Hand 4, including running text, labels, and diagrams.
	\item \textbf{Voynich 4 Text}: Voynich written in Hand 4, without text in labels and diagrams.
	\item \textbf{Voynich 5}: Voynich written in Hand 5, including running text, labels, and diagrams.
	\item \textbf{Voynich 5 Text}: Voynich written in Hand 5, without text in labels and diagrams.
\end{enumerate}

The particular transcription system used to write Voynichese can have a measurable effect on the statistical properties of the text itself. All of the documents above have been converted into three different transcription systems: Simplified Maximal, Full Maximal, and Minimal. The important issue of transcription is discussed in Section~\ref{vts}.

\subsubsection{Voynich Transcription Systems}\label{vts}

Scholars of the Voynich manuscript have proposed several transcription systems to assign characters to particular Voynich glyphs. These transcription systems include FSG, Bennett, Currier, Frogguy, EVA (Extensible Voynich Alphabet), and V101 (see \citealt{zandbergen2010} for an overview). The most commonly used of these systems is EVA. The systems differ in the assumptions they make about what constitutes a single character or character variant, and these assumptions can have an effect on the statistical properties of the text. Research on the most plausible character set is ongoing, and therefore any analysis of Voynichese should take into account the particular assumptions of the given transcription system. 

\begin{figure}
\centering
\includegraphics[scale=0.3,width=.7\linewidth]{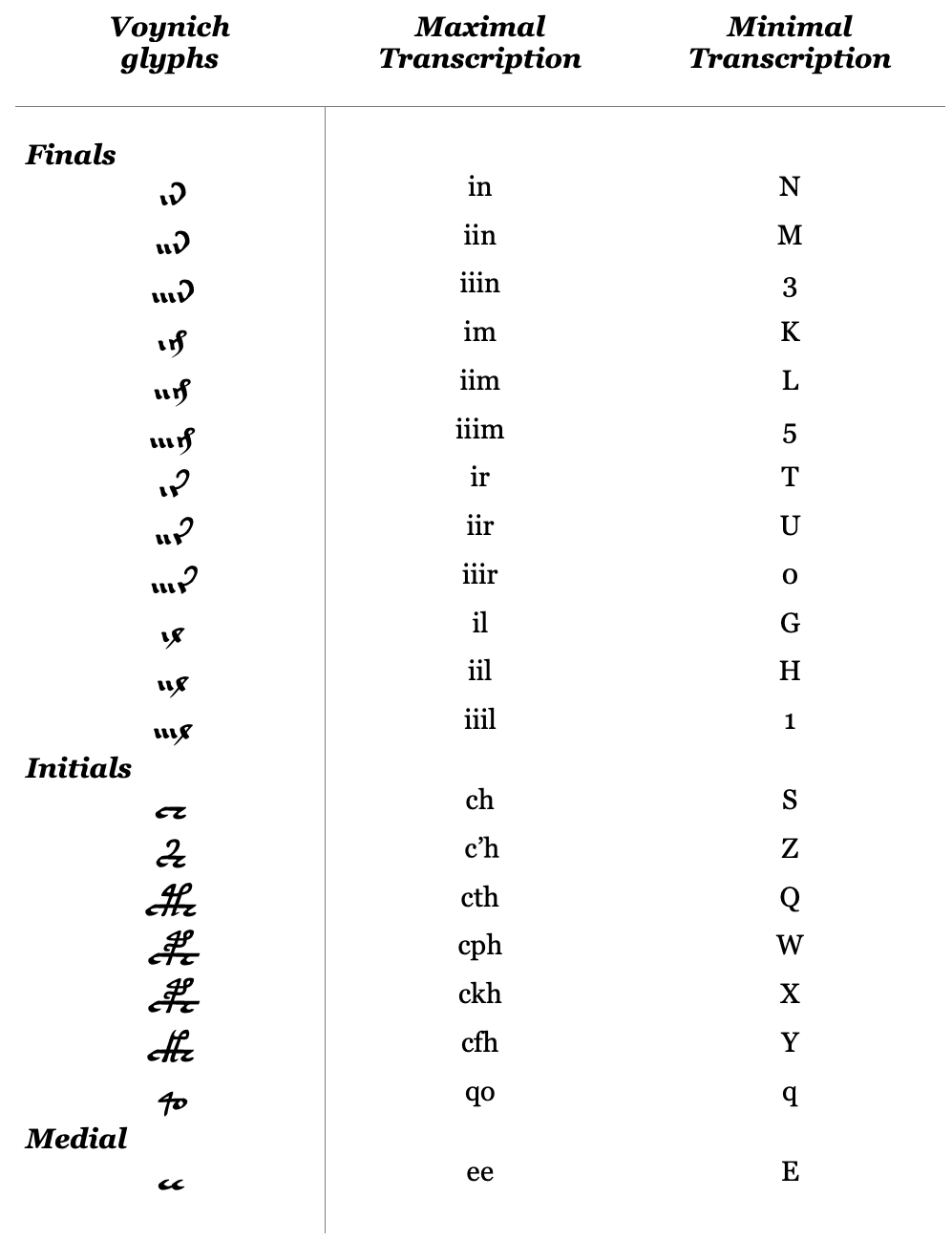}
\caption{Glyph combinations and their Minimal and Maximal transliterations. Maximal Voynich is equivalent to EVA, except that plumes are represented by apostrophes. For Minimal Voynich, we have made the substitutions given above, following Currier's schema. The last two were suggested by \citet{zandbergen2010}. The \textit{ee} combination is very common in the middle of words. The \textit{qo} combination is almost always an initial sequence, with \textit{q} being followed by \textit{o} 98\% of the time. }
\label{maxmin}
\end{figure}

The most significant assumption is whether or not certain sequences of Voynich glyphs constitute a single character or a sequence of multiple characters. Many of these common glyph sequences occur either word-finally or word-initially. They are written as sequences of multiple characters in the EVA transcription system, while the earlier Currier transcription considers them to be single characters. Transcription systems can be ranked according to whether common Voynich glyph sequences are minimally or maximally decomposed into individual characters. The EVA transcription was designed to be convertible to other major transcription systems.\footnote{The exception is V101, which makes different assumptions about many character variants. In particular, V101 assumes that many similar-looking glyphs, which in other systems are considered to be variants of the same character, are different characters. It is not easy to convert between V101 and other systems, and V101  has a larger character inventory. We are utilizing EVA because it is the system used for the interlinear files and because we believe the character inventory size is more plausible.} The characters of EVA are therefore intended to represent a lower bound on the length of characters to allow for the conversion to all possible compositions of characters in other systems. Whether or not EVA makes correct assumptions, it is the most convenient transcription system for analyzing Voynich in many cases because it allows for easy conversion into other systems. By contrast, Currier is the most minimally decomposed transcription system of the major systems. Common glyph common combinations tend to be represented as a single character rather than multiple characters.

\begin{figure}
\centering
\includegraphics[scale=0.3]{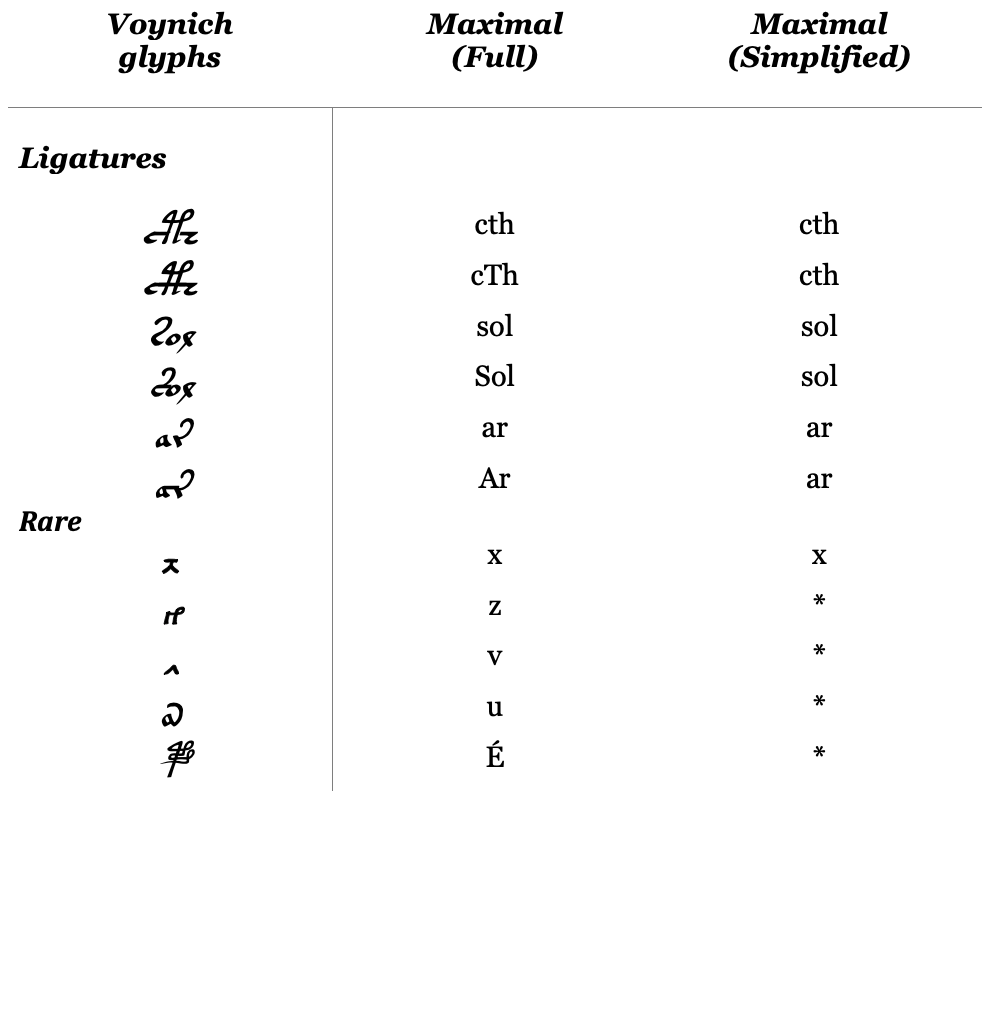}
\label{simplifiedfull}
\caption{The Simplified Maximal transcription: ligatures are ignored, and rare characters (including unusual plumes) are represented by an asterisk.}
\end{figure}

We take EVA as the basis for our most decomposed system, i.e. the Maximal transcription.\footnote{In a previous version of this paper, the Maximal transcription was identical to EVA. Here we introduce a single difference involving the treatment of plume strokes. The effect on character entropy is minimal: conditional character entropy falls 0.0-3.2\% depending on the sample text.} The only difference between our Maximal transcription and EVA is in the way that plumes are transcribed. Plumes are looping strokes which are found almost exclusively above the \textit{ch} bench characters (see the example in the Initials section of Figure \ref{simplifiedfull}). There are seven plumes found elsewhere in the document. In EVA, the \textit{c} with a plume above it is written as \textit{s}, and it is considered to be the same character as the glyph \textit{s} which is written with a single connected stroke and is found without an \textit{h} in words like \textit{sol}. For our Maximal transcription, we want the glyphs to be as decomposed as possible, and so we use the apostrophe character for all plumes. Thus for example the word which is transcribed as \textit{shedy} in EVA is transcribed as \textit{c'hedy} in the Maximal transcription. We consider the \textit{s} written with a single unconnected stroke to be a separate character.

For a Minimal transcription, we take EVA and substitute all of the glyph combinations in Currier's system. We add two additional glyph combinations based on suggestions from \citet{zandbergen2010}. The differences between the Minimal and Maximal transcriptions are outlined in Figure~\ref{maxmin}. Minimal Voynich represents our effort to create a transcription in which common glyph sequences are minimally decomposed into multiple characters. With future research, handwriting and script analysis will hopefully determine the plausibility of particular glyph decompositions with a higher degree of certainty. But for the present it is useful to compare two transcription systems which represent upper and lower bounds of compositionality.

Transcription systems also differ in the extent to which they represent ligatures and infrequent characters. Ligatures are horizontal lines that are sometimes employed to connect two characters. They may simply be the result of a fluent writing style. These are distinguished in the Full Maximal Transcription (and EVA) by capitalizing the first of the letters (see Figure \ref{simplifiedfull}). 

There are sixteen characters which appear less than fifty times each in the entire manuscript. Some of the infrequent characters are recognizable astrological symbols or are found only on cosmological diagrams, while others appear to be variants of other characters or even typographical mistakes. There are also about 250 unreadable glyphs, which are represented in EVA by an asterisk symbol. Altogether, these infrequent characters account for less than 0.15\% of the text. 

\begin{figure}
\centering
\includegraphics[scale=0.4,width=\linewidth]{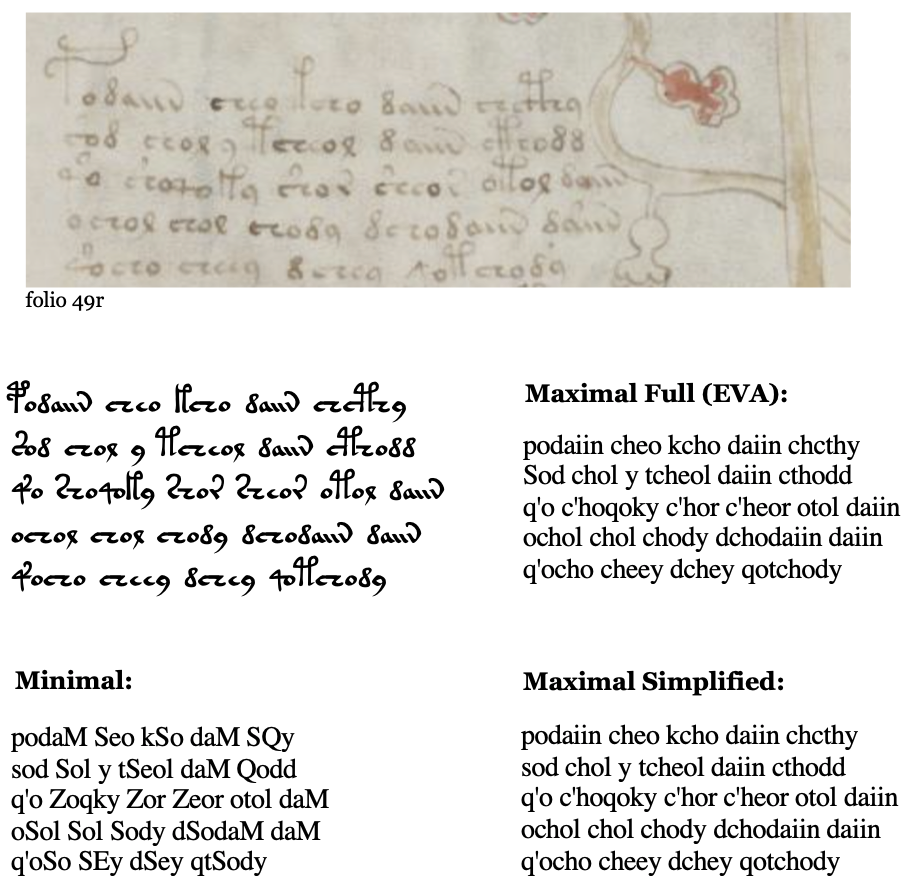}
\caption{Full Maximal, Simplified Maximal, and Minimal transcriptions of folio 49 recto, paragraph 2, lines 1-5}
\label{trans-ex}
\end{figure}

Figure~\ref{simplifiedfull} demonstrates the differences between the Full Maximal and Simplified Maximal transcriptions. Full Maximal is EVA with ligatures and rare characters included. Simplified Maximal removes the ligatures and uses an asterisk to designate all rare characters (with the exception of the \textit{x}, which is the most frequent of the rare characters and appears in both diagrams and text). The Simplified Maximal alphabet has about half as many characters (23 rather than 45), but this has a minimal effect on character statistics because of the infrequency of rare characters. Figure~\ref{trans-ex} shows the three transcriptions on a portion of folio 49 recto.

\break

\subsubsection{Voynich Document Preparation}\label{vdp}

The Voynich texts created for this analysis were derived from the Landini-Stolfi Interlinear Gloss File (LSI), which contains multiple transcriptions of the Voynich manuscript in EVA. We used Takeshi Takahashi's transcription for our corpus because it is the most complete.\footnote{A full set of transliterations can be found at \url{http://www.voynich.nu/transcr.html\#links}. There are a small number of gaps in Takahashi's transcription, including labels on the Rose pages, partial text on the foldout pages for f101, and the small amount of (non-Voynich) text on f116v, the last page.} Voynichese in the LSI is written out line-by-line and accompanied by notes and metadata. We used R to parse this code into a long table in which each word of Voynichese is associated with its precise position in the text. This consisted of deleting the notes, copying the page-level and line-level metadata, and separating the words by word breaks.\footnote{Credible word breaks in the LSI are represented by periods, and possible word breaks by commas. We followed the credible word breaks.} We then calculated the word's position from the beginning and end of the line and from the beginning and end of the paragraph. In our long table, each Voynichese word is listed sequentially along with the following metadata:

\begin{enumerate}

\item Full Maximal (EVA) transcription of the word
\item Simplified Maximal transcription of the word
\item Distance from the Beginning of the Line (1, 2, 3, etc.)
\item Distance from the End of the Line
\item Distance from the Beginning of the Paragraph
\item Distance from the End of the Paragraph
\item Paragraph/Diagram designation
\item Line Number on the page
\item Folio Number
\item Quire Number
\item Section of the Manuscript
\item Language (Currier's designation)
\item Hand (Davis' designation)
\item Transcriber

\end{enumerate}

The long table can then be consulted to create Voynich documents that focus on particular Voynich Languages, Hands, types of text, or positions within the folio, paragraph, or line. We used it to create the sixteen Voynich documents listed at the beginning of this section: \textit{Full Voynich, Full Voynich Text, Voynich A, Voynich A Text, Voynich B, Voynich B Text, Voynich 1, Voynich 1 Text, Voynich 2, Voynich 2 Text, Voynich 3, Voynich 3 Text, Voynich 4, Voynich 4 Text, Voynich 5,} and \textit{Voynich 5 Text.}

	\subsection{The Wikipedia Corpus}\label{wikicorpus}

To date, the online encyclopedia Wikipedia has versions in approximately 319 different languages. These versions are separate collaboratively edited editions which range widely in size. The English edition boasts over six million articles, although only about half of the language editions contain more than 1,000 entries.\footnote{See \url{http://meta.wikimedia.org/wiki/List\_of\_Wikipedias} for a current list of wikipedia versions by number of entries.} While it varies from language to language, a single wikipedia entry contains on average about 500 words, which means that the Voynich manuscript contains roughly the same amount of text as 75 wikipedia articles.\footnote{The entry word count for English wikipedia is much larger than most other languages because it tends to have longer entries. The average word count per entry in our English sample is over 3,700.} Our Wikipedia corpus consists of a sample of every language that has more than 100 articles. 

The primary advantage of the Wikipedia Corpus is that we can compare Voynich text with that of many different languages, language families, and scripts, and see whether Voynich falls within the range of plausible languages or language families, and if so, which languages or families it most closely resembles statistically. The conventions, motivations, and contents of modern online encyclopedias are obviously very different from that of a medieval herbal and astrological manuscript. However, both consist of discrete collections of informative text on specialized topics. The language of Voynichese should in many structural aspects be more akin to modern wikipedia entries than, for example, medieval diary entries or historical narratives. In contrast to encyclopedia entries, we would expect narratives to follow a temporal sequence throughout, for verbs to be predominantly in the past tense, and for certain names and pronouns to recur predictably. The Wikipedia corpus is thus particularly well-suited for comparison with the Voynich texts. It is also superior to more formal genres of corpora like newspaper corpora, which are written for a different purpose and contain far fewer languages.

The Wikipedia Corpus consists of 311 language samples written in thirty-four different scripts, categorized into thirty-nine major language families and seventy-three subfamilies. In most cases, the samples consist of the first 200,000 words from wikipedia entries for that language edition, listed alphabetically by headword. The corpus includes samples of many languages which are plausible candidates for Voynichese, e.g. Romance dialects like Corsican and Lombard and Germanic dialects like Bavarian and Low Saxon. There are also samples of extinct languages like Gothic, Anglo-Saxon, and Pali, as well as nine modern artificially-constructed languages including Esperanto and Lojban. 

Some language families are particularly well-represented in the Corpus, with ten or more language samples for each family. These are the Bantu, Germanic, Indic, Iranian, Malayo-Polynesian, Romance, Slavic, Tibeto-Burman, Turkic, and Uralic families.

		\subsubsection{Wikipedia Document Preparation}

The documents used for this analysis were obtained February 2021 from wikimedia dump files.\footnote{In a previous version of this paper, we described an earlier process for obtaining Wikipedia sample texts. In 2019, we processed wikidump files using the Python \texttt{Gensim} module and limited our samples to the first 500 articles in each language. Our new methodology creates dramatically cleaner texts, and the text sizes are more even because we limit the text by number of words rather than number of articles. Between 2019 and 2021, Wikipedia added the following languages: Awadhi, Balinese, Guianan Creole, Kotava, Ladin, Madurese, Mon, Moroccan Arabic, Nias, N'Ko, Saraiki, Inari Sami, and Sakizaya. The Northern Luri wikipedia was deleted. These changes have been made to the lists of languages and families given below.} These files are continually updated and available at \url{http://dumps.wikimedia.org}. We downloaded the BZIP2 compressed files titled \textit{Articles, templates, media/file descriptions, and primary meta-pages.} For most of the dump files, we ran a python script to process them into raw text documents containing the first 200,000 words from wikipedia entries for that language, including only articles that consist of 100 or more words.\footnote{The python script, along with the text files and a more in-depth description of the extraction process, is available at \url{http://www.lukelindemann.com/wiki_corpus.html}. For languages with scripts that do not use spaces for words, e.g. Thai and Chinese scripts, we instead extracted the first 100 articles. Some versions utilize two different scripts in different entries, e.g. Kashmiri has entries in either Devanagari or Arabic scripts, and we created two separate texts in this case.} 

We deleted any remaining metadata and tables by hand. We then further processed the texts by removing punctuation and capitalization, and deleted any characters with less than a .01\% occurrence in Latin, Cyrillic, Greek, Arabic, and Hebrew. In English, this filters everything but the lowercase letters (\textit{a-z}). We filtered the text by the unicode range of the particular script in order to delete irrelevant characters. 

Wikipedia versions differ widely by average article length. A small number of languages, including Cree, Cheyenne, and Inupiak, contain mostly short single-line articles with a repetitive structure. For these samples we included articles of all sizes and hand-deleted repetitive text, but this means that the resulting sample texts are quite short. Some other wikipedia versions, most notably Cebuano and Waray-Waray, have an inflated number of articles because a high percentage of entries were created by bots. Bot-created entries also tend to be short and formulaic, although this should be a minimal issue for our corpus because we deleted articles of less than 100 words.\footnote{Note that we have not filtered articles for whether they were created by bots, such as the Lsjbot (cf. \url{https://en.wikipedia.org/wiki/Lsjbot}.}

		\subsubsection{Wikipedia Languages by Family}

This is a full list of the language samples in the Wikipedia Corpus categorized by language family and sub-family. An ideal corpus would contain a large number of languages from each language family, and the sub-family categories would represent languages at an approximately equal time depth of divergence. However, the list of Wikipedia languages, while representing an impressive diversity of language families, is nevertheless skewed heavily towards European languages. 

The categorization below is an attempt to group together languages which are genetically similar while keeping the size of the categories approximately equal. Sub-families were chosen in language families with a large representation in the Corpus. If there are single languages from distinct sub-families, an ``Other'' category is used. For simplicity, extinct languages and proto-language progenitors of a family (e.g. Latin, Sanskrit, and Gothic) are also grouped into the ``Other'' category. 

There are two categories in this list that are not based upon genetic relatedness. All artificially-constructed languages are grouped under a single category. The Constructed languages that have Wikipedia versions are all international auxiliary languages meant to facilitate communication (as opposed to artistic constructed language like Klingon or Quenya). They are all heavily based on the vocabulary and grammar of European languages.\footnote{The exception is Lojban, which is a constructed logical language designed as an experiment in eliminating syntactic ambiguity.} The second category is that of Creoles, which are not the product of language divergence in a single family but rather have a complex genetic relationship with two or more language families. Here they have been subcategorized by their lexifier language, which is the language from which most of their vocabulary is drawn. 

\begin{enumerate}

	\item \textbf{Afro-Asiatic}: 
	\begin{enumerate}
		\item
		\textbf{Semitic:} Amharic, Arabic, Aramaic, Egyptian Arabic, Hebrew, Maltese, Moroccan Arabic, Tigrinya 
		\item 
		\textbf{Other families:} Hausa, Kabyle, Oromo, Somali
	\end{enumerate}
	
	\item 
	\textbf{Albanian}: Albanian
	
	\item 
	\textbf{Algonquian}: Atikamekw, Cheyenne, Cree (Canadian Syllabics), Cree (Latin)

	\item 
	\textbf{Armenian}: Armenian, Western Armenian
	
	\item 
	\textbf{Athabaskan}: Navajo
	
	\item 
	\textbf{Austroasiatic}: Khmer, Mon, Santali, Vietnamese, Banjar
	
	\item 
	\textbf{Aymara}: Aymara
	
	\item 
	\textbf{Baltic}: Latgalian, Latvian, Lithuanian, Samogitian
	
	\item 
	\textbf{Caucasian}: Abkhazian, Adyghe, Avar, Chechen, Ingush, Kabardian Circassian, Lak, Lezgian
	
	\item 
	\textbf{Celtic}: Breton, Welsh, Irish, Scottish Gaelic, Manx, Cornish
	
	\item 
	\textbf{Constructed}: Esperanto, Interlingua, Interlingue, Ido, Kotava, Lojban, Lingua Franca Nova, Novial, Volapük
	
	\item
	\textbf{Creoles}: 
	\begin{enumerate}
		\item
		\textbf{English:} Bislama, Jamaican Patois, Norfolk, Sranan, Tok Pisin
		\item
		\textbf{French:} Haitian
		\item
		\textbf{Portuguese:} Guianan Creole, Papiamentu
		\item
		\textbf{Spanish:} Zamboanga Chavacano
	\end{enumerate}
	
	\item
	\textbf{Dravidian}: Kannada, Malayalam, Tamil, Tulu, Telugu

	\item \textbf{Germanic}: 
	\begin{enumerate}
		\item
		\textbf{Anglic:} English, Scots, Simple English 
		\item
		\textbf{Dutch:} Afrikaans, Dutch, West Flemish, Zeelandic
		\item
		\textbf{Frisian:} North Frisian, West Frisian, Saterland Frisian
		\item
		\textbf{High German:} Alemannic, Bavarian, German, Ripuarian, Luxembourgisch, Palatinate German, Yiddish 
		\item
		\textbf{Low German:} Low Saxon, Dutch Low Saxon  
		\item
		\textbf{North Germanic:} Danish, Faroese, Icelandic, Norwegian (Nynorsk), Norwegian (Bokm\r{a}l), Swedish
		\item
		\textbf{Other families/proto-languages:} Anglo-Saxon, Gothic, Limburgish
	\end{enumerate}
	
	\item
	\textbf{Hellenic}: Greek, Pontic
	
	\item
	\textbf{Indic}: 
	\begin{enumerate}
		\item
		\textbf{Central:} Awadhi, Hindi, Urdu, Fiji Hindi, 
		\item 
		\textbf{Eastern:} Assamese, Bengali, Maithili, Odia, Bihari, Bishnupriya Manipuri, 
		\item 
		\textbf{Northern:} Doteli, Nepali 		
		\item 
		\textbf{Northwestern:} Sindhi, Punjabi, Saraiki, Western Punjabi
		\item
		\textbf{Southern:} Sinhalese, Divehi, Marathi, Goan Konkani
		\item
		\textbf{Western:} Romani, Gujarati
		\item
		\textbf{Other families/proto-languages:} Kashmiri (Arabic), Kashmiri (Devanagari), Sanskrit, Pali (Devanagari), Pali (Latin)	
	\end{enumerate}
	
	\item
	\textbf{Inuit}: Greenlandic, Inuktitut (Canadian Syllabics), Inuktitut (Latin), Inupiak

	\item 
	\textbf{Iranian}: Sorani, Zazaki, Persian, Gilaki, Kurdish, Mazandarani, Ossetian, Pashto, Tajik

	\item \textbf{Iroquoian}: Cherokee

	\item \textbf{Japonic}: Japanese

	\item \textbf{Kartvelian}: Georgian, Mingrelian

	\item \textbf{Koreanic}: Korean
	
	\item 
	\textbf{Malayo-Polynesian}: 
	\begin{enumerate}
		\item
		\textbf{Javanesic:} Banyumasan, Javanese
		\item
		\textbf{Malayic:} Indonesian, Malay, Minangkabau
		\item
		\textbf{Polynesian:} Tongan, Hawaiian, Maori, Samoan, Tahitian 
		\item
		\textbf{Phillipine:} Central Bicolano, Cebuano, Gorontalo, Ilokano, Pangasinan, Kapampangan, Tagalog, Waray-Waray
		\item
		\textbf{Other families:} Acehnese, Balinese, Buginese (Buginese), Buginese (Latin), Chamorro, Fijian, Madurese, Malagasy, Nauruan, Nias,  Sundanese, Tetum
	\end{enumerate}
	
	\item 
	\textbf{Mande}: Bambara, N'Ko

	\item 
	\textbf{Moksha}: Moksha

	\item 
	\textbf{Mongolic}: Buryat, Mongolian, Kalmyk

	\item 
	\textbf{Niger-Congo}: 
	\begin{enumerate} 
		\item
		\textbf{Bantu:} Kongo, Lingala, Luganda, Northern Sotho, Chichewa, Kikuyu, Kinyarwanda, Kirundi, Shona, Swati, Sesotho, Swahili, Tswana, Tsonga, Tumbuka, Twi, Venda, Xhosa, Zulu
		\item
		\textbf{Other families:} Akan, Ewe, Fula, Igbo, Kabiye, Sango, Wolof, Yoruba
	\end{enumerate}

	\item 
	\textbf{Nilotic}: Dinka

	\item 
	\textbf{Quechua}: Quechua

	\item 
	\textbf{Romance}: 
	\begin{enumerate}
		\item
		\textbf{Italo-Dalmatian:} Italian, Corsican, Sicilian, Neapolitan, Venetian, Tarantino
		\item 
		\textbf{Gallo-Romance:} Catalan, French, Franco-Provençal, Ladin, Occitan, Picard, Norman, Walloon 
		\item
		\textbf{Gallo-Italic:} Piedmontese, Ligurian, Lombard, Emilian-Romagnol 
		\item
		\textbf{Iberian:} Aragonese, Asturian, Extremaduran, Galician, Ladino, Mirandese, Portuguese, Spanish 
		\item
		\textbf{Other families/proto-languages:} Aromanian, Friulian, Romanian, Romansh, Sardinian, Latin
	\end{enumerate}
	
	\item 
	\textbf{Slavic}
	\begin{enumerate}
		\item
		\textbf{East Slavic:} Belarusian, Belarusian Taraškievica, Russian, Ukrainian 
		\item
		\textbf{West Slavic:} Czech, Kashubian, Lower Sorbian, Upper Sorbian, Polish, Rusyn, Slovak, Silesian 
		\item
		\textbf{South Slavic:} Bulgarian, Bosnian, Croatian, Macedonian, Serbo-Croatian, Serbian, Slovenian, Old Church Slavonic
	\end{enumerate}

	\item 
	\textbf{Tai}: Lao, Shan, Thai, Zhuang

	\item 
	\textbf{Tibeto-Burman}: Tibetan, Min Dong, Dzongkha, Gan, Hakka, Burmese, Newar, Wu, Chinese, Classical Chinese, Min Nan, Cantonese

	\item 
	\textbf{Tupian}: Guarani

	\item 
	\textbf{Turkic}:
	\begin{enumerate}
		\item
		\textbf{Oghuz:} Azerbaijani, Chuvash, Gagauz, South Azerbaijani, Turkmen, Turkish
		\item
		\textbf{Karluk:} Uyghur, Uzbek
		\item
		\textbf{Kipchak:}  Bashkir, Crimean Tatar, Karakalpak, Kazakh, Karachay-Balkar, Kirghiz, Tatar (Cyrillic), Tatar (Latin)
		\item
		\textbf{Siberian:} Sakha, Tuvan
	\end{enumerate}

	\item 
	\textbf{Uralic}: 
	\begin{enumerate}
		\item \textbf{Finnic:} Estonian, Finnish, Vepsian, Võro
		\item \textbf{Permic:} Komi-Permyak, Komi, Udmurt 
		\item \textbf{Mari:} Meadow Mari, Hill Mari 
		\item \textbf{Sami:} Inari Sami, Northern Sami
		\item \textbf{Other families:} Erzya, Hungarian, Livvi-Karelian, Northern Sami
	\end{enumerate}

	\item \textbf{Uto-Aztecan}: Nahuatl

	\item \textbf{Vasconic}: Basque

\end{enumerate}

	\subsection{The Historical Corpus}

The Wikipedia Corpus contains a large number of languages and language families, but it consists entirely of modern texts. It is therefore necessary to compare Voynichese to contemporaneous historical manuscripts as well, because there are important differences between modern and historical texts which are not typically addressed in statistical analyses of Voynichese. 

One important point of difference is spelling standardization, which is much higher in most modern languages than it is in medieval manuscripts. This is less of an issue for medieval Latin texts, as Latin has been standard since the Classical Period, but it is an important consideration for the many written languages which had yet to standardize by the 15th century. Spelling variation will have an effect on statistics like type-token ratio because a single word will be represented by multiple types.  

A second important difference concerns the typographical conventions of scribes. Because all literature was written and copied by hand during this period, scribes developed hundreds of abbreviations and symbols to represent frequently occurring phrases, words, and grammatical functions. This was especially prevalent in Latin texts. It introduces variability of a different type, and has an effect on statistics like the information entropy of the text. However, most modern transcriptions of historical manuscripts omit these abbreviations and conventions for readability. 

\subsubsection{Description of the Corpus}
The Historical Corpus consists of transcriptions of manuscripts written between 400 and 1600 AD. The corpus is continuously updated as we discover new sources of digitally transcribed historical manuscripts. The languages represented in the corpus currently include English, Georgian, Hebrew, Icelandic, Italian, Latin, Persian, and Spanish. The majority of the texts are in Latin and English.

In order to match the presumed contents of the Voynich manuscript, we have made an effort to include texts on magic, astrology, and alchemy. Many of the important texts in this genre, including the highly influential \textit{Secretum Secretorum}, were originally written in Arabic or Persian and were being translated into Latin and vernacular European languages during the time that the Voynich manuscript was created. We have included a Latin and English translation of the \textit{Secretum Secretorum}, an English translation of the \textit{Alphabet of Tales}, Agrippa's \textit{Three Books of Occult Philosophy}, a Spanish translation of \textit{Picatrix}, and Bruno's Latin \textit{De Magia}. We have also included Trithemius' \textit{Steganographia}, which is ostensibly about magic and spirit communication but is in fact an enciphered treatise on cryptography. In the historically related topic of Medicine, we have included the \textit{Science of Cirurgie} and the archives of Richard Napier's medical records collected by the Casebooks Project at the University of Cambridge. 

A secondary goal in the creation of the Historical Corpus is to collect manuscripts in parallel diplomatic and normalized versions. The diplomatic version of a manuscript uses special characters to faithfully replicate the original abbreviations and typographical conventions, while the normalized version does not use abbreviations and the orthography is typically modernized.\footnote{A diplomatic transcription is technically distinct from a type facsimile, which uses digital fonts to replicate the exact appearance of the text. For our purposes, we consider the most faithful available reproduction of a text to be the diplomatic transcription.} This allows us to directly compare the effect of typographical conventions on the same text, which may provide insights into the peculiar properties of Voynichese. We have included parallel diplomatic and normalized versions of three texts:  the Icelandic \textit{Codex Wormianus}, the English \textit{Medical Casebooks}, and the Latin \textit{Necrologium Lundense}. For the other Latin texts we also created our own abbreviated forms of the texts based on widespread orthographic conventions; this material will be discussed in forthcoming work.

A similar issue is found with abjad scripts like Arabic and Hebrew, which are typically written without vowels. The exclusion of vowels has an effect on the entropy statistics of a text. We have included two versions from the Tanakh: one with and one without the \textit{niqqud} diacritics which are used primarily to mark vowels. 

Table~\ref{histtable} lists the historical manuscripts in the corpus, along with their language, script, approximate date of composition, and author (or translator or scribe). 

\begin{table}[ht]\begin{tabular}{| l | l | l | l | l |}

\hline
\textbf{Name}				& \textbf{Language}	& \textbf{Script}		& \textbf{Author}	& \textbf{Date} 	\\

\hline
Medical Casebooks			& English			& Latin			& Richard Napier	 & 1597			\\

\hline
Three Books of				& English			& Latin			& Heinrich Cornelius Agrippa			& 1509			\\
Occult Philosophy			&				&				&								&			\\
\hline
Science of Cirurgie			& English			& Latin			& Lanfranc of Milan		& 1306			\\

\hline 
Secretum Secretorum 		& English			& Latin	 		& Robert Copland	& 1528				\\
						&				&				& (translator)				&					\\

\hline
Alphabet of Tales			& English			& Latin			& Etienne de Besan\c{c}on		& 1400		\\

\hline
Amiran-Darejaniani			& Georgian		& Georgian		& Mose Xoneli			& 1150		\\

\hline
Mishneh Torah				& Hebrew			& Hebrew			& Maimonides		& 1170	\\

\hline
Masoretic Tanakh 			& Hebrew			& Hebrew			& Aaron ben Moses	& 1008		 \\
						&				&				& ben Asher (scribe)	&			\\	

\hline
Codex Wormianus			& Icelandic		& Latin			& Unknown			& 1350			\\

\hline
La Rettorica				& Italian			& Latin			& Brunetto Latini		& 1261	\\

\hline
Necrologium Lundense		& Latin			& Latin			& Unknown		& 1123						\\

\hline
De Ortu Et Tempo Antichristi	& Latin			& Latin			& Adso Deruensis	& 900						\\

\hline
Historia Hierosylmitanae 		 & Latin 		& Latin			& Albert of Aix		& 1125						\\
Expeditionis				&			&				&				&						\\
\hline
De Magia					& Latin			& Latin			& Giordano Bruno	& 1590							\\

\hline
Secretum Secretorum		& Latin			& Latin			& Philip of Tripoli 			& 1270			\\
						&				&				& (translator)				&				\\

\hline
Steganographia			& Latin			& Latin			& Johannes Trithemius				& 1499			\\

\hline
Sindbad-Name				& Persian			& Arabic			& Zahiri Samarqandi		& 1362		\\

\hline
Picatrix					& Spanish			& Latin			& pseudo-Majriti				& 1256		\\

\hline

\end{tabular}
\caption{Details of historical manuscripts}\label{histtable}
\end{table}


\subsubsection{Historical Document Preparation}

The transcribed texts were obtained from multiple sources. The Georgian, Italian, and Persian texts come from the TITUS Project at the University of Frankfurt.\footnote{\url{http://titus.uni-frankfurt.de}} The Icelandic text comes from the Medieval Nordic Text Archive.\footnote{\url{http://clarino.uib.no/menota/page}} The Hebrew Masoretic Tanakh comes from Sacred Texts\footnote{\url{http://www.sacred-texts.com/bib/tan/index.htm}} and the Christian Classics Ethereal Library\footnote{\url{http://www.ccel.org/a/anonymous/hebrewot/home.html}}, while the \textit{Mishneh of Maimonidies} was obtained here.\footnote{\url{http://kodesh.snunit.k12.il/i/0.htm}}

Of the Latin texts, the \textit{Secretum Secretorum} comes from the Corpus Corporum of the University of Zurich\footnote{\url{http://mlat.uzh.ch/MLS/}}, the \textit{Necrologium Lundense} comes from the Necrologium Lundense Online\footnote{\url{http://notendur.hi.is/mjm7/}}, \textit{De Ortu et Tempo Antichristi} and \textit{Historia Hierosylmitanae Expeditionis} come from the Latin Library\footnote{\url{http://www.thelatinlibrary.com}}, and \textit{De Magia} and \textit{Steganographia} come from the Twilit Grotto.\footnote{\url{http://esotericarchives.com}}

The English \textit{Alphabet of Tales} and \textit{Science of Cirurgie} come the Corpus of Middle English Prose and Verse at the University of Michigan.\footnote{\url{http://quod.lib.umich.edu}} The \textit{Medical Casebooks} come from the Casebooks Project of the University of Cambridge.\footnote{\url{http://casebooks.lib.cam.ac.uk}} The English translation of the \textit{Secretum Secretorum} comes from Colour Country,\footnote{\url{http://www.colourcountry.net/secretum/}} and Agrippa's \textit{Three Books of Occult Philosophy} are from the Twilit Grotto.\footnote{\url{http://esotericarchives.com}}

Some of these texts are much longer than the Voynich manuscript, and so we have included only a portion of the entire text. We restricted the Masoretic Tanakh to the \textit{Bereshit}, i.e. the Book of Genesis. We included the introduction and first two books of the Mishneh, the first sixty pages of the \textit{Codex Wormianus}, and the first three books of the \textit{Science of Cirurgie}. The Necrologium Lundense currently has normalized and diplomatic transcriptions of three folios (f124v, f125r, f125v), but they are substantive enough that we included them here. For the Medical Casebooks, we copied the first fifty chronologically sorted consultations taken by Richard Napier and written in his hand. 

As with the Wikipedia Corpus, the historical documents were cleaned by removing capitalization and punctuation, as well as notes made by transcribers. For texts in Latin scripts, symbols with a less than .01\% occurrence were removed. Texts written in non-Latin scripts were filtered by unicode range. For the diplomatic texts, special characters -- including character variants and astrological symbols -- were left intact. 


\section{Conditional Character Entropy in Voynichese}\label{sec:h2}

In this section, we demonstrate the usage of the Historical and Wikipedia corpora by examining the character-level properties of Voynichese. As discussed in \citet{bowernlindemann20}, we are particularly interested in the metric of conditional character entropy, or second-order character entropy (\textit{h2}). Conditional character entropy is a measure of the overall predictability of characters in a text. In his 1976 book on computational applications to scientific and engineering problems, Yale physicist William \citeauthor{bennett1976scientific} Jr. used a transcription of Voynichese to illustrate the concept of information entropy in language and its application to cryptography. He found the conditional character entropy of Voynichese to be surprisingly low compared to a sample of European plain texts and ciphers. This means that Voynichese characters are unusually predictable compared to most European languages. We discuss the definition of conditional character entropy and the history of its application to the Voynich manuscript more thoroughly in \citet{bowernlindemann20}. 

This conditional character entropy value of a text, \textit{h2}, is dependent upon the conventions of the script in which it is written. For example, \citet{bennett1976scientific} found that the \textit{h2} of Voynichese was roughly equivalent to that of a Hawaiian text. \citet{stallings1998understanding} pointed out that Bennett's Hawaiian sample used a simplified orthography that did not contain glottal stops or distinguish between long and short vowels, and this has the effect of making the Hawaiian text look more predictable (and more like Voynichese). The following factors potentially have an effect on the \textit{h2} value of a text:

\begin{enumerate}

	\item Document Length
	\item Character set size (total number of characters in the alphabet) 
	\item Type of script (alphabet, syllabary, abugida, abjad, etc.)
	\item Abbreviations and other typographical conventions
	\item Encoding process (if the text is a cipher)

\end{enumerate}

We discuss the first four of these factors in detail below (the fifth will be discussed in a forthcoming paper). We find that Voynich A and B are of a sufficient length that the \textit{h2} values are reliable, and that the exclusion of rare characters (Maximal Simplified as opposed to the Full Maximal transcription) has a negligible effect on entropy. 

The type of script (Maximal as opposed to Minimal) has a more appreciable effect on entropy, but all transcriptions of Voynich are significantly lower than any other text in the corpora. An analysis of script types in the Wikipedia and Historical corpora shows that Voynich most closely resembles an alphabetic script rather than an abjad, abugida, or syllabary.

We compare the parallel diplomatic and normalized versions of historical texts, as well as the forms of Hebrew with and without vowels, and conclude that the unusual character entropy of Voynich is not attributable to conventional scholarly abbreviations or the absence of characters that represent vowels. 

At the character level, Voynichese most closely resembles tonal languages written in the Latin script and other languages in which there is a restricted set of word-final characters. This is likely the result of an encoding process, and may suggest that Voynichese simplifies the phonemic distinctions of the language it represents. 

\subsection{Character Frequency Distribution}

In its simplest characteristics, Voynichese does not appear very different from other texts in the Historical and Wikipedia corpora. The character set size for both the Maximal transcription (42 characters) and Minimal transcription (45 characters) is well within the general range for alphabets: 25-92 characters.\footnote{The Simplified Maximal transcription has only 23 characters, but it excludes rare characters and therefore represents an absolute lowest estimate.} As discussed in Section~\ref{sec:wikicomp}, the Voynichese character set size is small compared to non-alphabetic script types like abugidas and syllabaries, but it is the right size for an alphabet.

\begin{figure}[h!]
\includegraphics[scale=0.15,width=\linewidth]{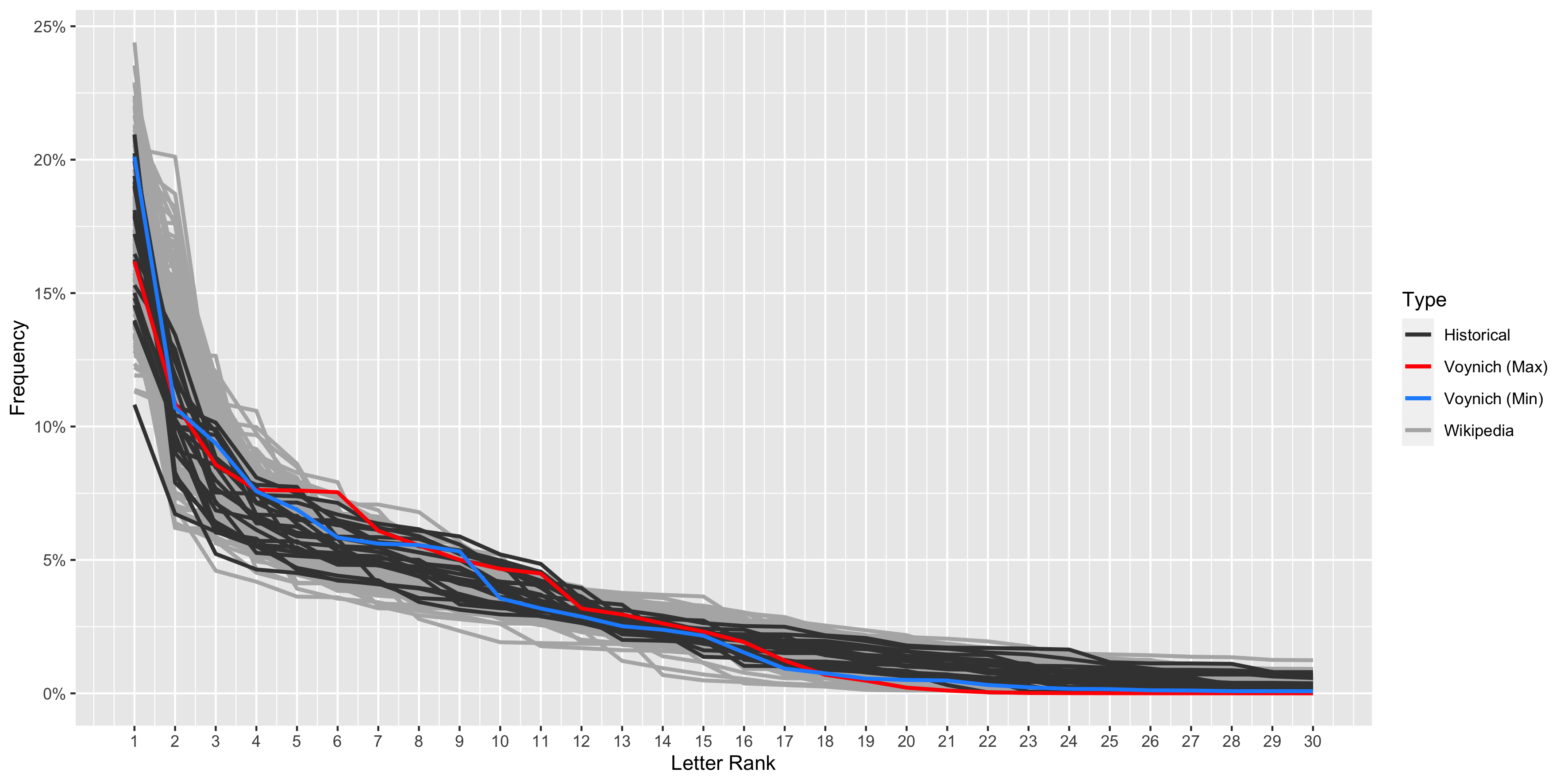}\centering
\caption{Proportional Frequency of the thirty most frequent characters in the Historical Corpus texts, the (alphabetic) Wikipedia Corpus texts, and Voynichese (Full Maximal and Minimal transcriptions). Simplified Maximal is identical to Full Maximal with regards to frequent characters. The first ranked character for each language is a space, and its frequency indicates average word length.}
\label{char-dist}
\end{figure}

Secondly, the character frequency distribution of Voynichese is fairly typical. This is demonstrated in Figure~\ref{char-dist}, which displays the ranked proportional frequencies of the thirty most frequent characters in Voynichese compared with those of texts in the Historical and Wikipedia corpora. 

Character frequency distribution is related to and reflected in the metric of unigram character entropy (\textit{H1}), which measures character-level predictability irrespective of position within the text. Here again, Voynichese is not unusual: \textit{H1} is 3.94 for Minimal Voynich and 3.91 for Maximal Voynich, while the overall range for alphabets in the corpora is from 3.57-4.82 bits. 

The Voynich text only begins to look unusual when we factor in the position of a character within the text. Conditional character entropy (\textit{h2}) measures the predictability of a character given the character that precedes it. This is the metric that \citet{bennett1976scientific} found to produce unusually low values in Voynichese, and on which we focus this analysis. 

\subsection{Entropy Variance and Document Length}

Two texts written in the same language and script may have slightly different \textit{h2} values due to differences in content and stylistic variation. With a large enough text sample, this variation is minimal. If a text sample is very short, there will not be enough data to obtain a reliable \textit{h2} result, and there will be more variation. 

The Voynich manuscript contains roughly 38 thousand words, of which 11 thousand are in Voynich A and 23 thousand are in Voynich B. We need to know whether these lengths are sufficient for obtaining reasonably certain \textit{h2} values, and what sort of variance can be expected. 

\begin{figure}[h!]
\includegraphics[scale=0.15,width=\linewidth]{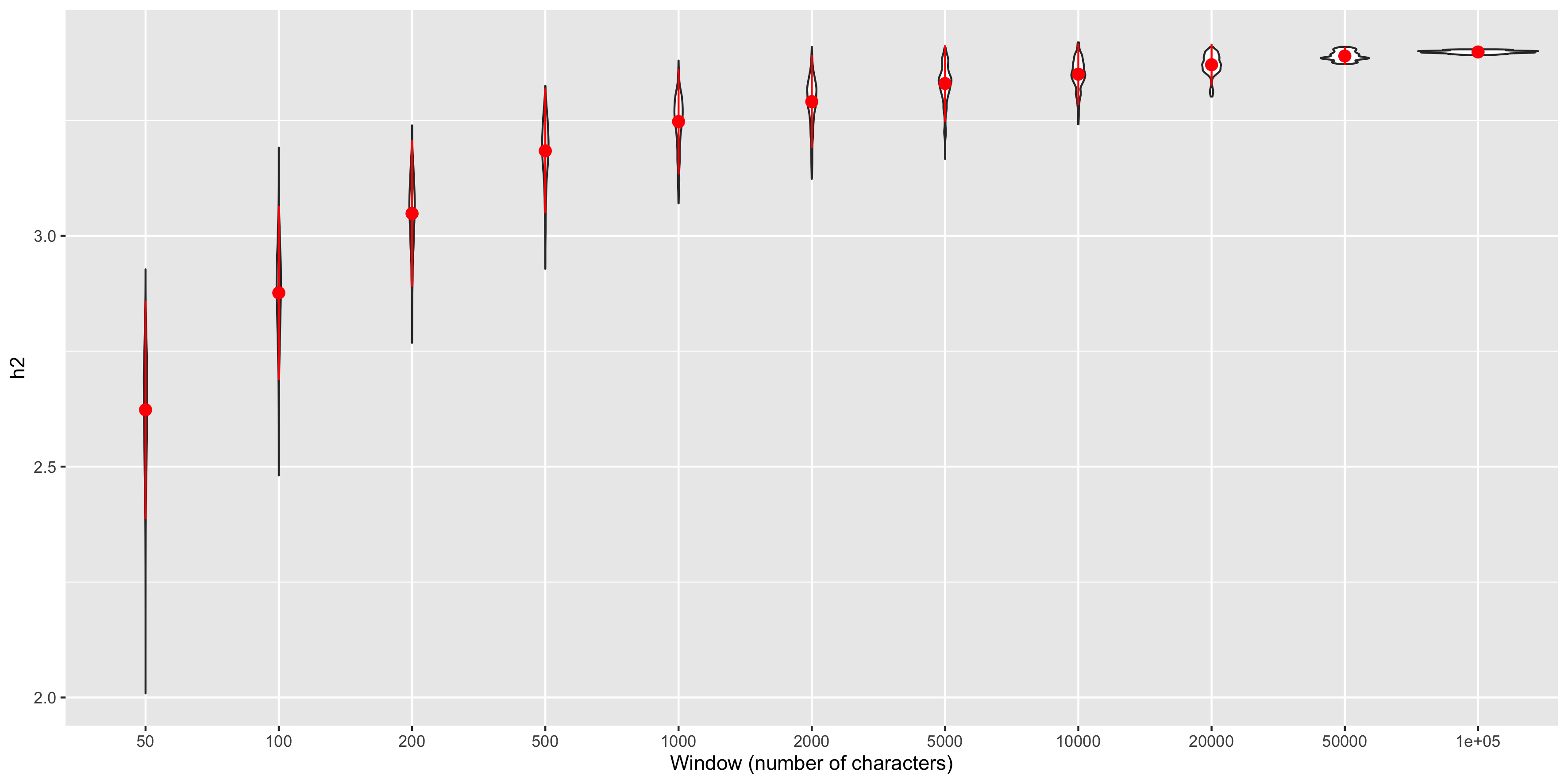}\centering
\caption{Density plot of \textit{h2} values for random samples of English at multiple word lengths from 50-100,000. Each window size is sampled 1,000 times. The red dot is the average and the line indicates one standard deviation from the mean.}
\label{en-win-h2}
\end{figure}

We tested \textit{h2} variance using the English wikipedia sample, which consists of 199,564 words. We calculated the \textit{h2} values samples of randomly selected sequences of text at various word lengths. The results are in Figure~\ref{en-win-h2}.

With a window of only 50 words, the average \textit{h2} is 2.62, there is a wide range of 2.0-2.9, and the standard deviation is 0.12. The average is much lower than the text's overall \textit{h2} value of 3.40. As the window size increases, the variance tightens and the averages converge on the overall value. With a window of 10,000 words, the average is 3.35, the range is 3.22-3.42, and the standard deviation is 0.033. This means that 95\% of the samples are within 0.066 bits of the average, and the average is within 0.05 bits of the text's overall \textit{h2} value. For documents of around 10,000 words we should therefore reasonably expect \textit{h2} to be accurate to about one-tenth of a bit. 

When running the same procedure on Voynich A and Voynich B, the \textit{h2} variance is comparable to that of English. The standard deviation at 50 words is 0.13 for A and 0.15 for B (compared to 0.13 in English), and at 5,000 words it is 0.049 for A and 0.072 for B (compared to 0.048 in English). 

This means that the Voynich A and Voynich B sample are large enough to obtain reasonable entropy calculations. However, an analysis at the level of sections, scribal hands, or folios will be somewhat less reliable.

\subsection{Entropy in Voynichese}

\begin{figure}[h!]
\includegraphics[scale=0.2,width=\linewidth]{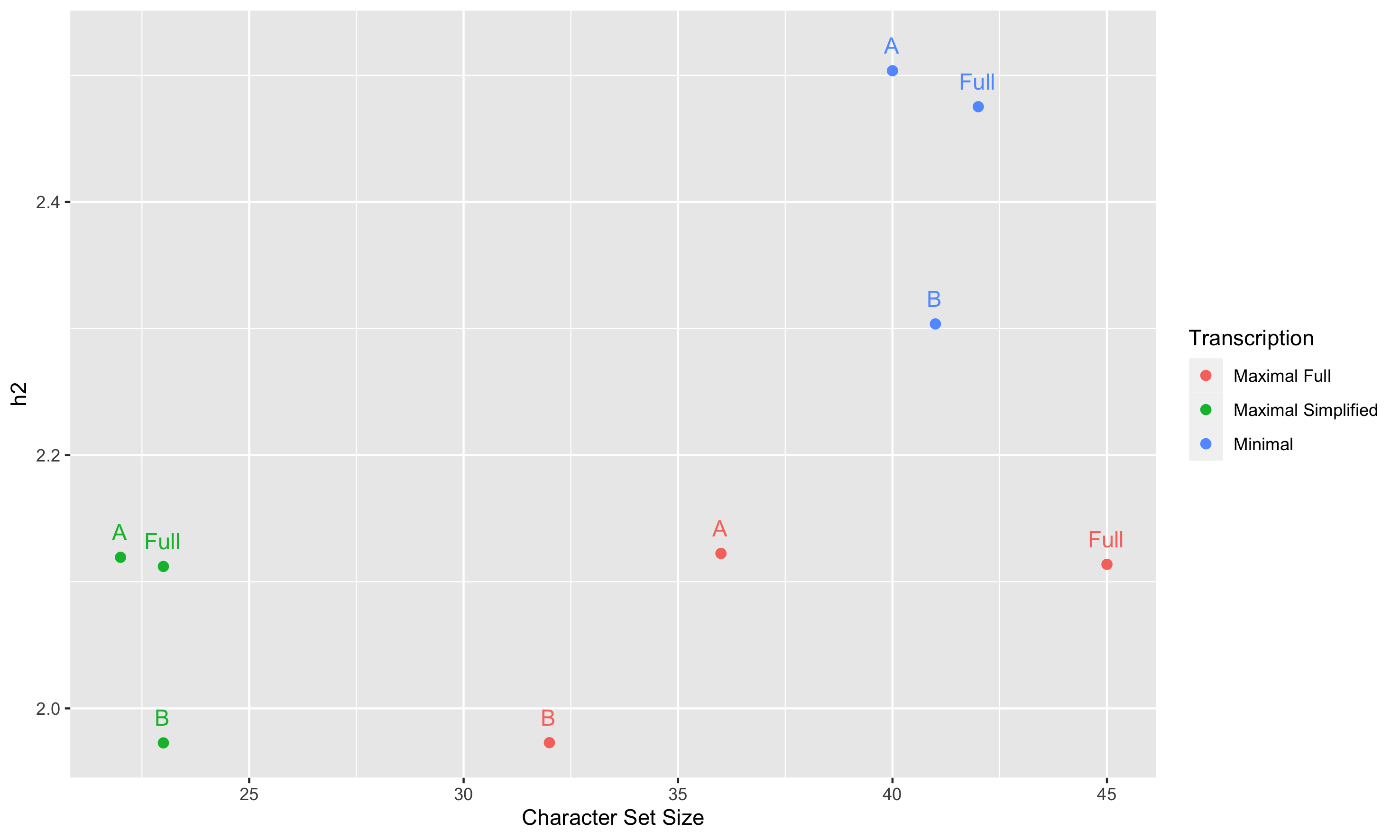}\centering
\caption{Character set size is plotted against conditional character entropy (\textit{h2}). Lower \textit{h2} values indicate more predictability in character bigrams. }
\label{voyh2}
\end{figure}

For Voynichese, Language A and Language B pattern differently. Conditional character entropy in Language B is lower regardless of the transcription system, while in Language A it is only slightly higher than in the combined Full text (see Figure \ref{voyh2}). The compositionality of the transcription system has an effect on \textit{h2}. Maximal Voynich has a lower \textit{h2} than Minimal Voynich, because glyph compositions are based upon common glyph sequences (making the text appear more predictable).

The \textit{h2} values for running paragraph text (excluding labels on diagrams) is slightly lower than text with labels included. Voynichese in labels and diagrams has a higher \textit{h2}. Scribe Hand 1, which is used to write Language A, has a nearly identical \textit{h2}. Scribe Hands 2 and 3, which are used to write Language B, have values quite similar to B. Scribe Hands 4 and 5 have values slightly higher than Language B, as they are mostly employed in writing labels and diagrams. These values are listed in Appendix \ref{voytables}. 

In all cases, the character set size for Voynichese is between 21 and 45 characters, and the conditional character entropy ranges from 1.91 to 2.56 bits.

\subsubsection{Character Set Size}

The Full Maximal and Maximal Simplified transcriptions have nearly identical \textit{h2} values despite a significant difference in the size of the character sets. With the Full Voynich text, the Simplified Maximal transcription has twenty-two fewer characters, but there is only a .08\% difference in conditional character entropy (2.114 to 2.112). 

This is significant, because conditional character entropy can be affected by character set size (as noted in \citealt{stallings1998understanding}). It is potentially important because the upper bound for conditional character entropy is determined by the value of \textit{H0}, calculated as the logarithm of character set size. For example, a text that uses an alphabet of 16 characters will have a maximum conditional character entropy of 4 bits, while an alphabet of 49 characters will have a maximum \textit{h2} of 7 bits. 

However, if the additional characters are rare, the overall effect on \textit{h2} is slight. The process of cleaning a text by removing highly infrequent characters does not have an appreciable effect on entropy. After removing the capitalization, numerals, and punctuation from the raw English sample, there are 75 remaining characters. These includes characters which are very rare in English or exist only in foreign words, such as \textit{ü} and \textit{ç}. Processing the sample involves removing characters which appear with a frequency less than 0.01\%, after which only 27 characters remain. However, the difference in \textit{h2} between the filtered and unfiltered English sample is only 0.08\% (3.406 to 3.403). 

On the other hand, there is an appreciable difference in entropy between the Minimal and Maximal transcriptions, despite the fact that the character set size is roughly similar. This demonstrates that decisions about the composition of high-frequency glyph sequences have a greater effect on entropy than decisions about the inclusion of low-frequency characters.

\subsubsection{Languages A and B}

Voynich Language A and Language B are similar at the character level. Despite the fact that they have different distributions at the word level, the most frequent character sequences are roughly the same in both languages. There are two exceptions, which illustrate the difference in entropy between the two texts. 

The \textit{-edy} glyph sequence found at the end of words is eighty-six times more common in Voynich B (one out of five words in Voynich B end with this sequence). Secondly, the \textit{qo-} sequence at the beginning of a word is about twice as common in Voynich B (also found in one in five words). The frequency of these two sequences alone substantially increase the predictability of the Voynich B text, and this is the main source of the differences in conditional character entropy between A and B. If the two sequences are removed from both texts, then the \textit{h2} value for Language A and Language B come within about 1\% of each other.\footnote{The frequency of this common glyph sequence is partially attributable to a single word \textit{chedy}, which is the most common word in B and almost entirely absent from A. But even when this word is disregarded, \textit{-edy} is significantly more common in B.}

		\subsection{Comparison to the Wikipedia Corpus}\label{sec:wikicomp}

Of the 311 wikipedia language samples represented in the Wikipedia Corpus, none of them have an \textit{h2} comparable to Voynichese. Voynichese has lower values, meaning that its text is more predictable. While the Minimal Voynich transcription is slightly higher (with an average \textit{h2} of 2.48 rather than 2.11), this is still lower than the \textit{h2} range in the Wikipedia Corpus, from 2.77-6.14.

\begin{figure}[h!]
\includegraphics[scale=0.2,width=\linewidth]{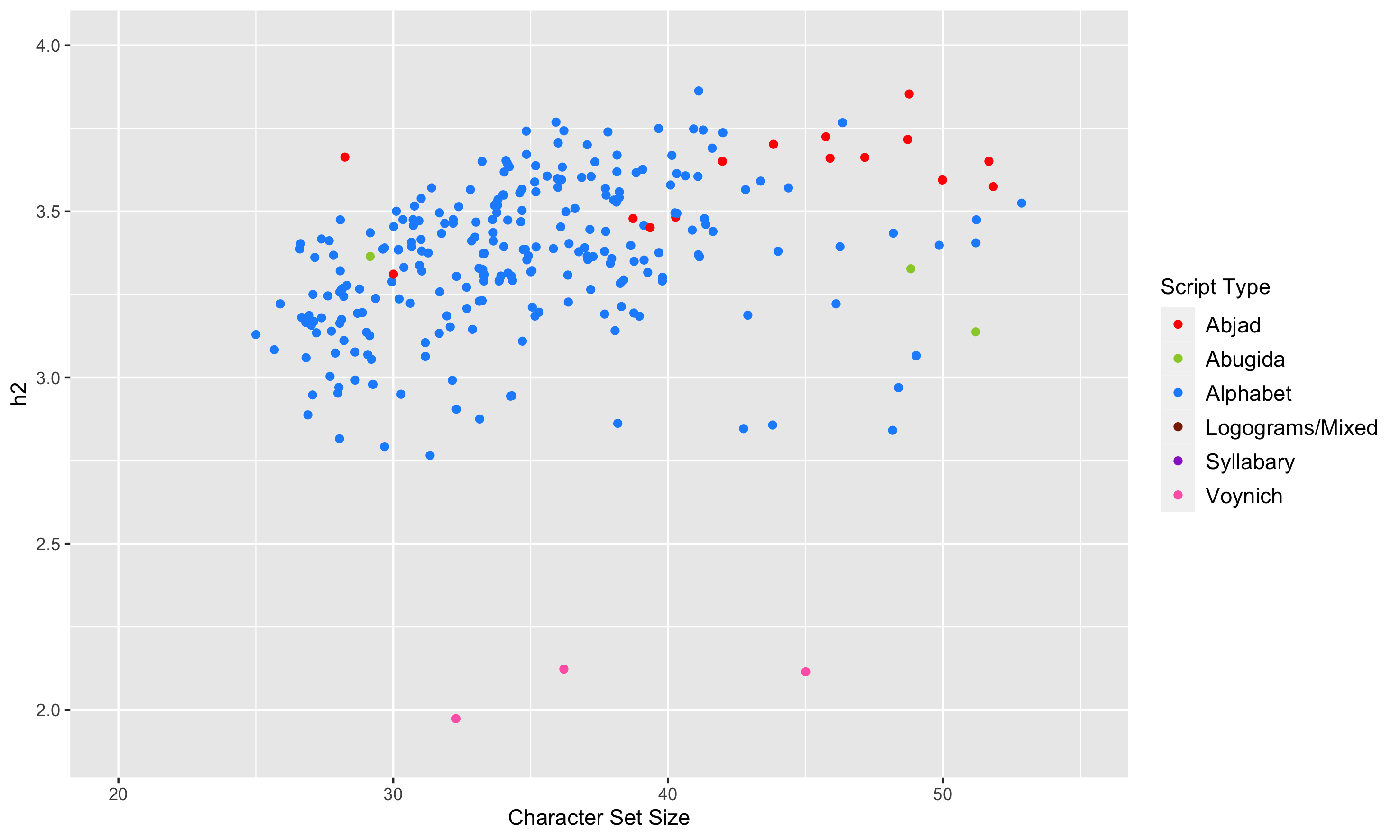}\centering
\caption{Conditional Character Entropy and Character Size for Wikipedia Languages from 20-55 characters. Full Maximal Voynich (A, B, and Full) is shown at the bottom. The languages with the closest \textit{h2} values are Hawaiian (2.77), Venda (2.79), Tswana (2.82), Min Dong (2.84), Tahitian (2.85), Hakka (2.86), and Sango (2.86).}
\label{wiki-h2-1}

\end{figure}

Figure~\ref{wiki-h2-1} depicts the character set size and conditional character entropy for texts that use between 20 and 55 characters and have an \textit{h2} range from 2.5 to 4. This is the range of most of the alphabets in the Corpus. The majority of Wikipedia versions (202 languages) are written in the Latin script, although the Corpus also includes samples of Cyrillic, Georgian, Gothic, Greek, Ol Chiki and N'Ko. Languages written in Cyrillic, which are mostly Slavic and Turkic, tend to have a somewhat higher \textit{h2}. 

Most of the abjads are also in this range. Abjads are writing systems in which consonants are written and vowels are (mostly) not represented. The abjads in the Corpus include Arabic and Hebrew. They are used to write Afro-Asiatic (specifically Semitic), Indic, Iranian, and Turkic languages, as well as the Germanic language Yiddish. They also have a somewhat higher \textit{h2} on average.

The conditional character entropy of texts written in Latin scripts ranges from 2.8-3.8, and includes all of the languages with the lowest \textit{h2} values. The languages with \textit{h2} values closest to Voynichese are Hawaiian, Venda, Tswana, and Min Dong. Hakka and Min Dong are Tibeto-Burman languages, while Venda and Tswana are Bantu (Niger-Congo). It is noteworthy that all except Hawaiian are tonal languages that use a Latin script for their orthography. That is, they systematically collapse suprasegmental distinctions.

\begin{figure}[h!]
\includegraphics[scale=0.2,width=\linewidth]{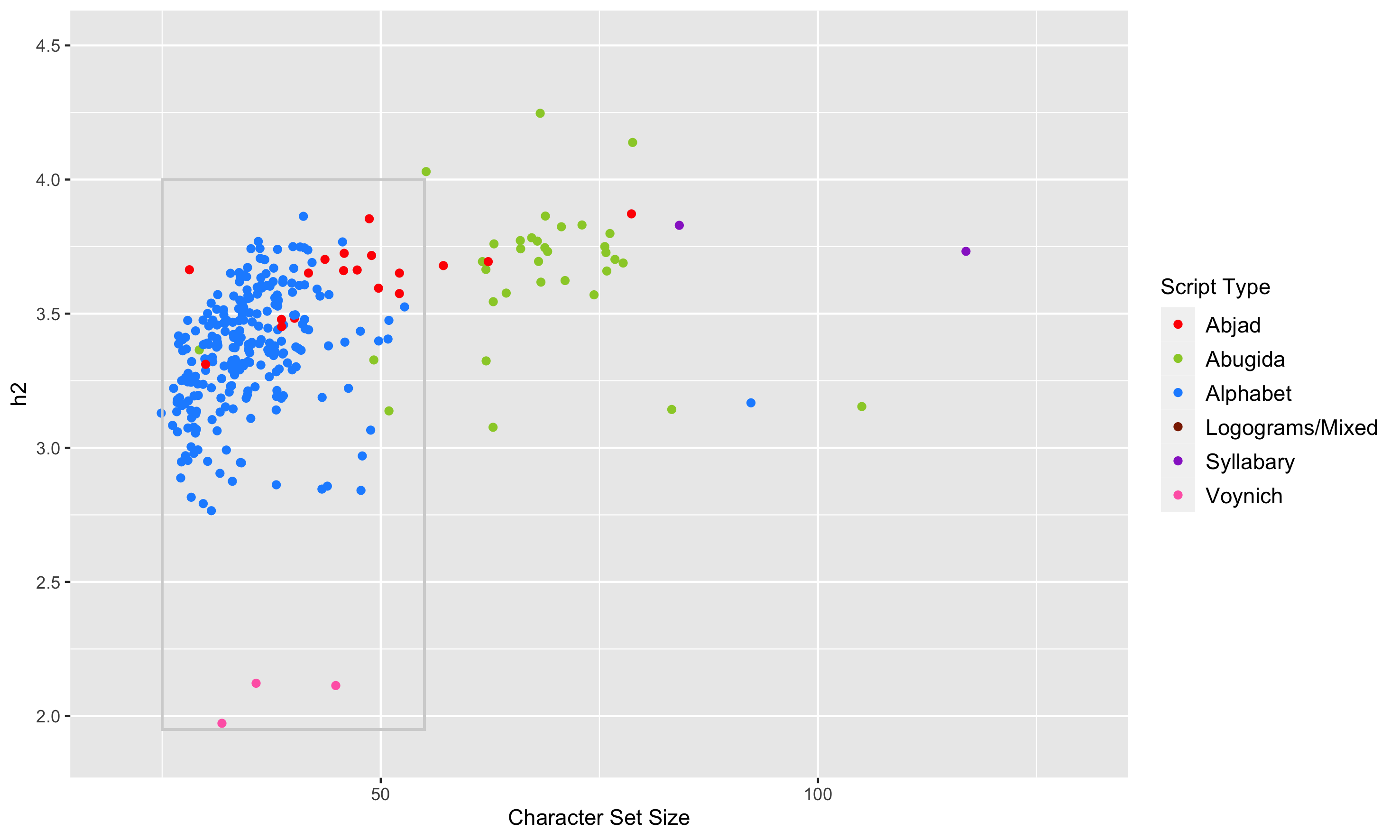}\centering
\caption{Conditional Character Entropy and Character Size for Wikipedia Languages from 20-130 characters. Most of the Abugidas are in this range, of which the Tibeto-Burman and Indic languages tend to have lower \textit{h2} values and the Austroasiatic and Tai languages have higher \textit{h2} values. The grey box indicates the range of the previous graph.}
\label{wiki-h2-2}
\end{figure}

Expanding outward, Figure~\ref{wiki-h2-2} includes languages that contain up to 130 characters and have a conditional entropy between 2.5 and 4.5. These primarily consist of the abugidas. Abugidas are writing systems in which consonant-vowel sequences are written as a unit, with consonants as the primary symbol and vowels added to it. The abugidas in the corpus include Bengali, Buginese, Devanagari, Gujarati, Gurmukhi, Kannada, Khmer, Lao, Malayalam, Myanmar, Odia, Sinhala, Tamil, Telugu, Thaana, Thai, and Tibetan. They are all derived from the Brahmi script, and are used to write Austroasiatic, Dravidian, Indic, Tai, and Tibeto-Burman languages. The abugidas usually have many more characters but tend to have only a slightly higher entropy. 

\begin{figure}[t!]
\includegraphics[scale=0.2,width=\linewidth]{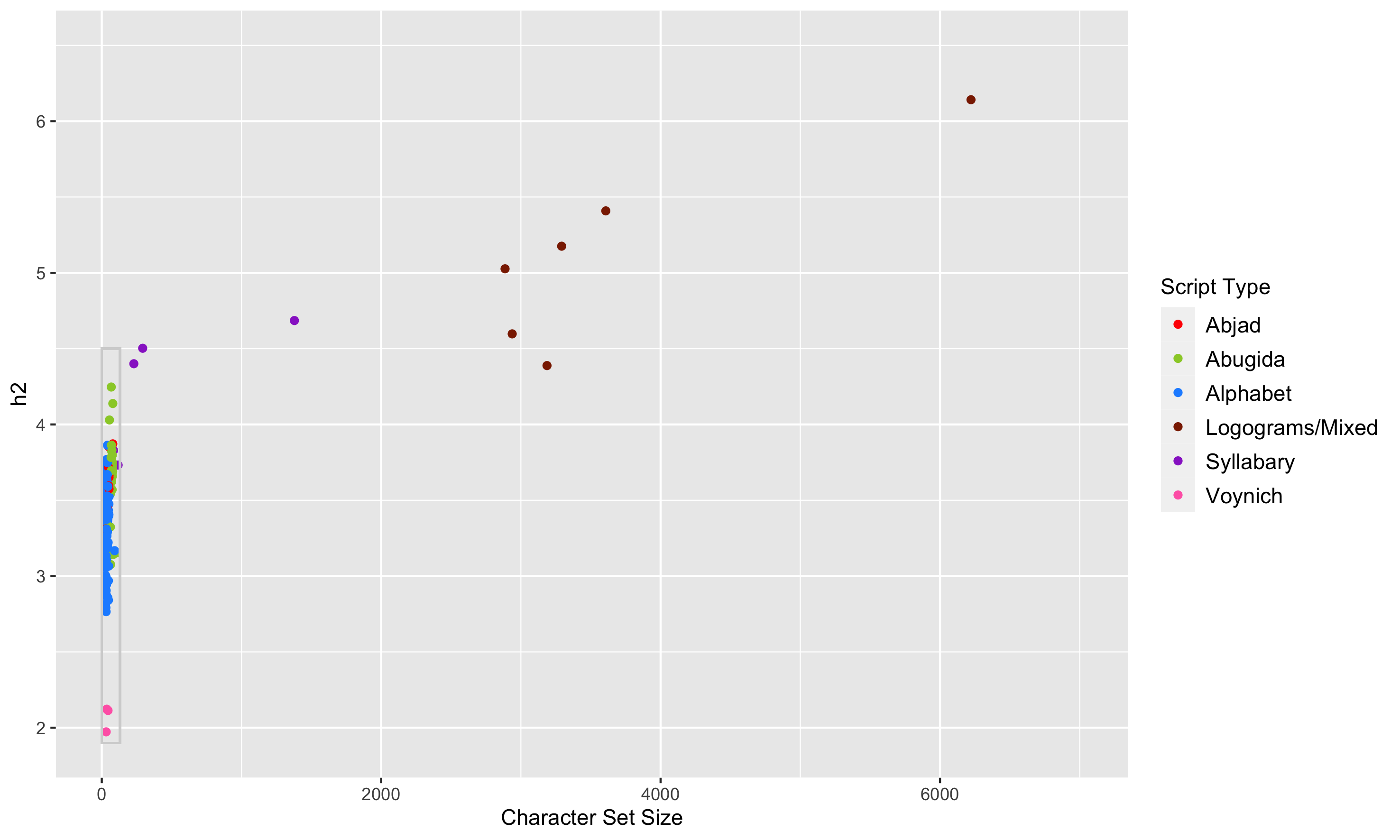}\centering
\caption{Conditional Character Entropy and Character Size for Wikipedia Languages from 20-7000 characters. The grey box indicates the range of the previous graph.}
\label{wiki-h2-3}
\end{figure}

Expanding even further, Figure~\ref{wiki-h2-3} includes languages that contain up to 7000 characters and have a conditional entropy up to 6.5. The languages that are written with logograms have the highest character set sizes and highest entropy values. For these languages, each character represents a morpheme or word, and thus character entropy is approximately equal to word entropy. Chinese logograms are used to write varieties of Chinese and other Tibeto-Burman languages: Cantonese, Chinese, Classical Chinese, Gan, and Wu. Also included under this category is Japanese, which uses a mixed writing system with logograms and two syllabaries, and ranges between syllabaries and logograms in character set size and \textit{h2}.

Between this extreme and the abugidas are syllabaries, in which a single character denotes a syllable. The syllabaries include scripts designed for the Iroquoian language Cherokee and the Inuit language Inuktitut. They also include the Hangul script used for Korean and the Ethiopic (Ge'ez) script used to write Amharic and Tigrinya. Technically, Ethiopic is an abugida and Hangul is an alphabet, but both are represented in unicode by separate codes for each full syllable rather than with combining characters. Thus they have the character set size and conditional entropy in the syllabary range. The full table of values for each language may be found in Appendix \ref{wikitables}. 

The Voynichese script patterns most closely with the alphabets in the Wikipedia Corpus. The character set size of Full Maximal and Minimal Voynichese are quite similar to most of the alphabets in the sample, but the \textit{h2} values are lower than we see with any of the languages in the Wikipedia Corpus. The abjads are also similar, with an equivalent character set size but a slightly higher \textit{h2}. Voynichese clearly falls outside of the range of most abugidas, syllabaries, and logograms.

	\subsection{Comparison to the Historical Corpus}

\begin{figure}[t!]
\includegraphics[scale=0.2,width=\linewidth]{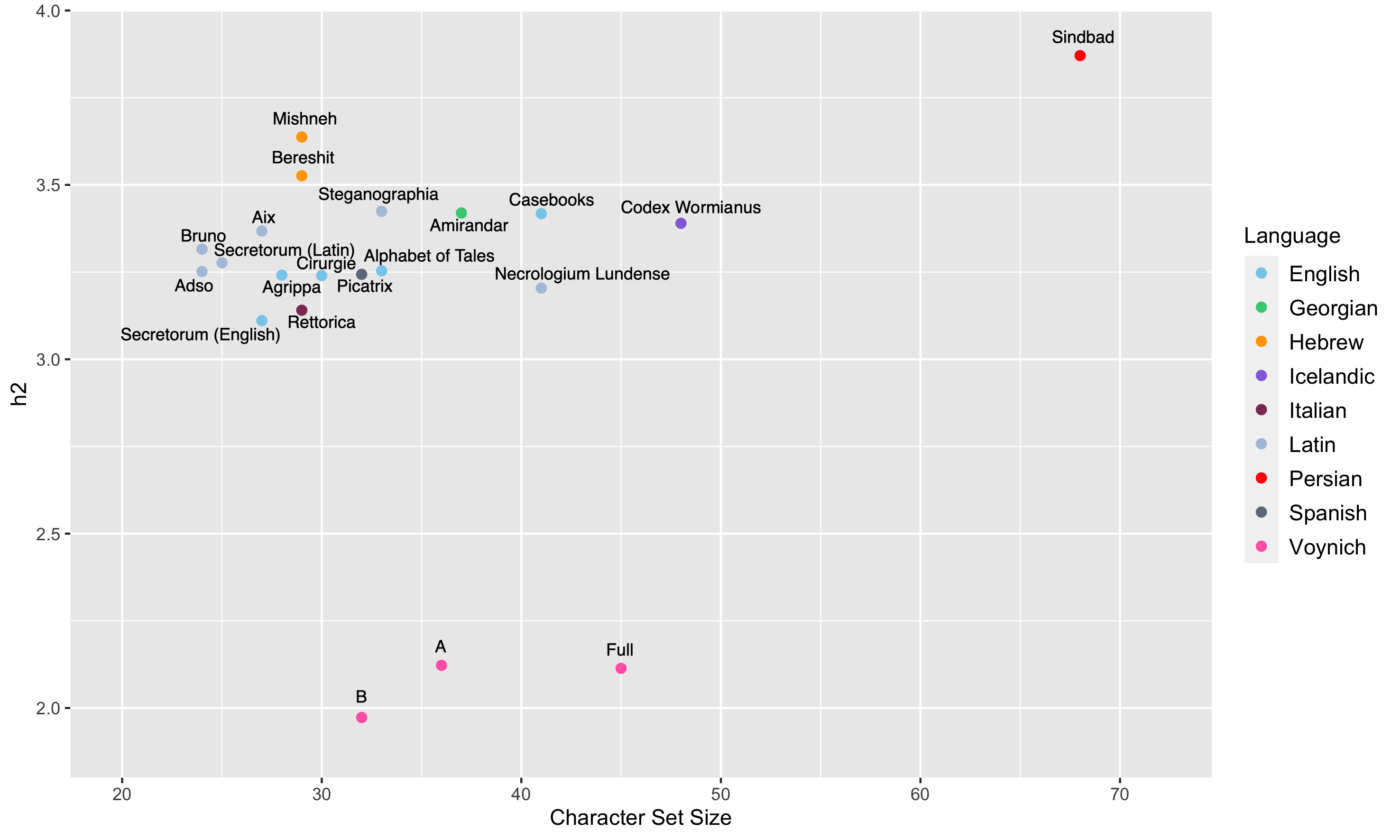}\centering
\caption{Conditional Character Entropy and Character Size for Historical Texts. The Voynich transcription is Full Maximal, and the historical scripts are normalized and unabbreviated. Hebrew is calculated without vowel markings. The alphabets range from 3-3.5 and the abjads range from 3.5-4.}
\label{hist-h2-1}
\end{figure}

Figure~\ref{hist-h2-1} shows the range for the texts in the Historical Corpus. The texts written with alphabets (Latin and Georgian) have an \textit{h2} range between 3 and 3.5, while the abjads (Hebrew and Arabic scripts) range from 3.5 to 4. The five English and five Latin texts demonstrate the variability in \textit{h2} and character set size within the same language. Some of these differences are attributable to script variation. The English Medical Casebooks, like the Icelandic Codex Wormianus and Necrologium Lundense, has a larger alphabet because it contains somewhat more characters in the normalized versions as well as the diplomatic versions.

\subsubsection{Parallel Diplomatic and Normalized texts}
\begin{figure}[t!]
\includegraphics[scale=0.2,width=\linewidth]{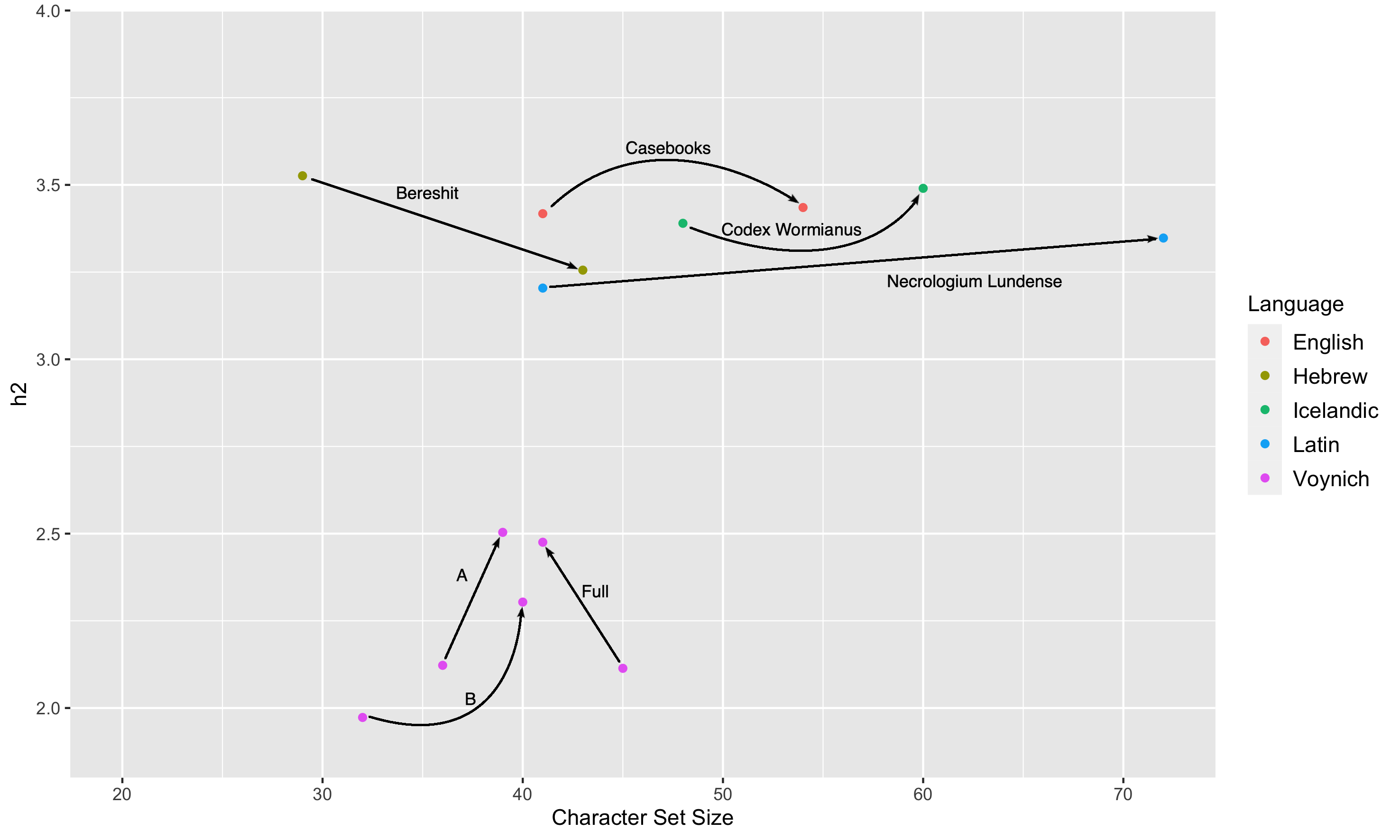}\centering
\caption{Differences between Parallel Comparison Texts: abbreviated Latin has higher conditional character entropy than unabbreviated Latin, and the diplomatic transcriptions of Icelandic and medical English have higher conditional character entropy than the normalized transcriptions. The Minimal transcription of Voynichese has higher conditional character entropy than the Full Maximal transcription. }
\label{hist-h2-2}
\end{figure}

Figure~\ref{hist-h2-2} shows how conditional entropy varies between parallel versions of the same text when different forms of transliteration are employed. For the English \textit{Casebooks}, Latin \textit{Necrologium Lundense}, and Icelandic \textit{Codex Wormianus}, this consists of the normalized and diplomatic versions. For the Hebrew Bereshit, this consists of the text with or without \textit{niqqud} vowel-marking diacritics. We also compare the Full Maximal and Minimal transcriptions of the Voynich texts.

The graph illustrates the variation in the conditional entropy among the versions of historical texts, which relates primarily to character set size rather than to the conditional entropy values. The biggest difference in \emph{h2} is from the Hebrew Bereshit text, where including \textit{niqqud} lowers the values of \emph{h2}  by 0.25 bits. Note that the variation between Voynich hands and characters is of a similar order, but all the Voynich measurements are substantially lower than the historical samples.

The usage of abbreviations and special characters has the effect of \textit{raising} the conditional character entropy of the English, Icelandic, and Latin texts and taking them further from the values we find for Voynichese. The Minimal transcription of Voynich has a slightly higher conditional character entropy, but it is clear that the extremely low conditional entropy of Voynichese is not simply attributable to a particular Voynich transcription system or the kinds of abbreviations and typographical conventions that were common in European manuscripts. 

\citet{reddy2011we} argue from the statistical distribution of letters and words that Voynichese most closely resembles an abjad. Many Voynichese characters are only found at the beginning or end of a word, which resembles the positional variants of letters in the Arabic script. Comparison with the Wikipedia and Historical corpora demonstrates that the abjad hypothesis is not, however, an explanation for the low conditional entropy in Voynichese. The abjads in our corpora have a higher conditional character entropy than the alphabets, and adding the vowels back in with \textit{niqqud} (essentially turning an abjad into an alphabet) lessens conditional character entropy substantially. If Voynichese is an abjad, it is a highly unusual one.\footnote{\citet{reddy2011we} also argue that the Voynich Manuscript might be written in an abjad because of the results from their two-state HMM investigations. The HMM deduces a word formula of A*B rather than picking out a class of consonants and one of vowels; this pattern was also found with their Arabic tests. However, this does not necessarily mean that there are no vowels represented in the text, rather that the regularity and singularity of word-final items is swamping other possible groupings.}

\subsection{What makes Voynichese unique?}

\subsubsection{Entropy and Bigram Frequency}

Conditional character entropy tells us about the predictability of a text at the character level. One way of thinking about it is this: if you look at any character in a text, how certain can you be, on average, that you will be able to guess the character that follows it? In the English sample from the Wikipedia corpus, for example, the letter \textit{q} is followed by the letter \textit{u} 96\% of the time.\footnote{Some of the exceptions include the words \textit{FAQs}, \textit{Iraqi}, and \textit{qi}.} In other words, the conditional frequency of the bigram \textit{qu} is 96\%. So if we see a \textit{q} in an English text, we can be reasonably certain that we know what the next letter is. 

\begin{figure}[t!]

\includegraphics[scale=0.12,width=\linewidth]{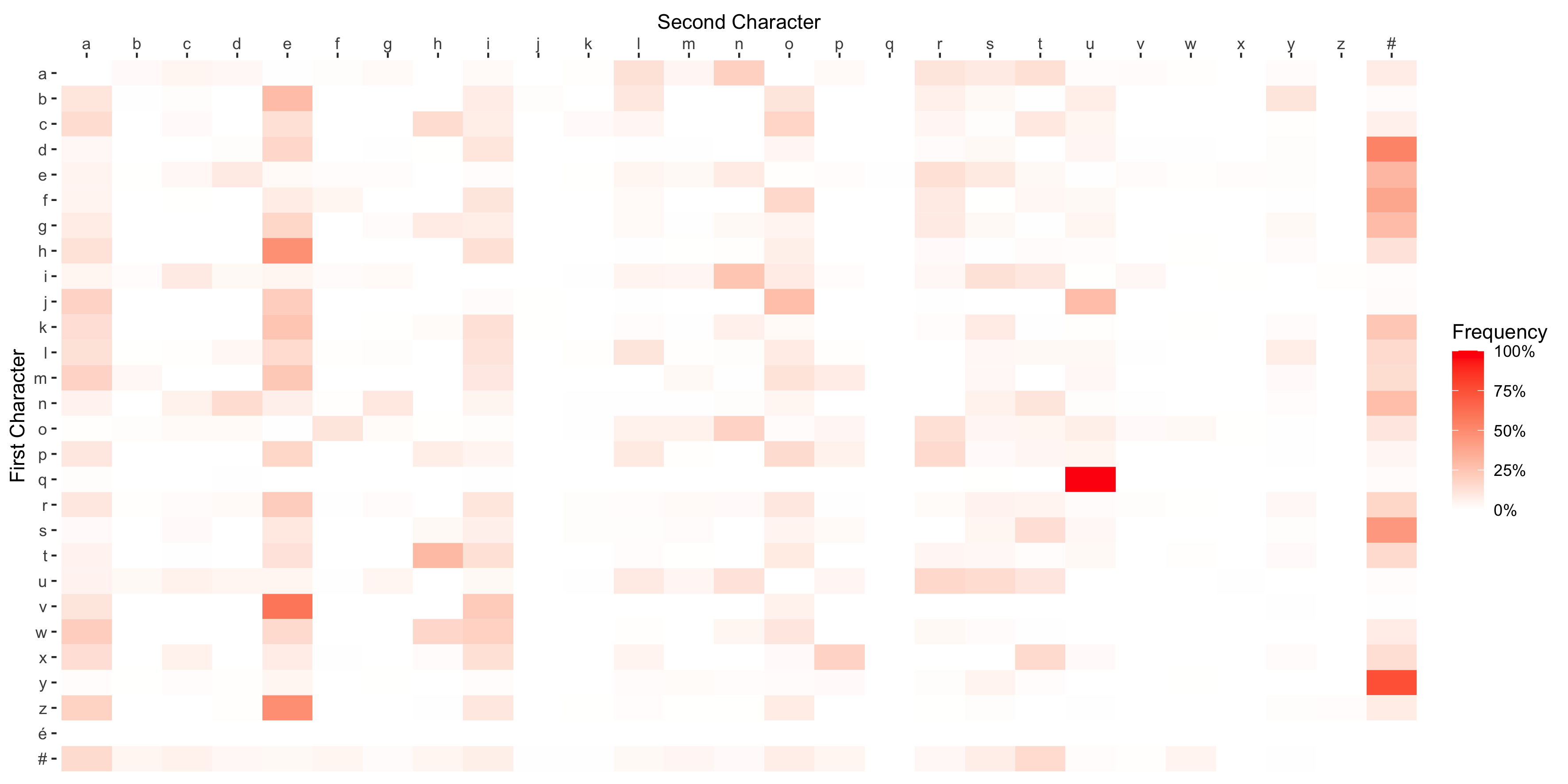}\centering
\caption{Conditional Frequency of English Bigrams}
\label{en-cond-freq}

\includegraphics[scale=0.12,width=\linewidth]{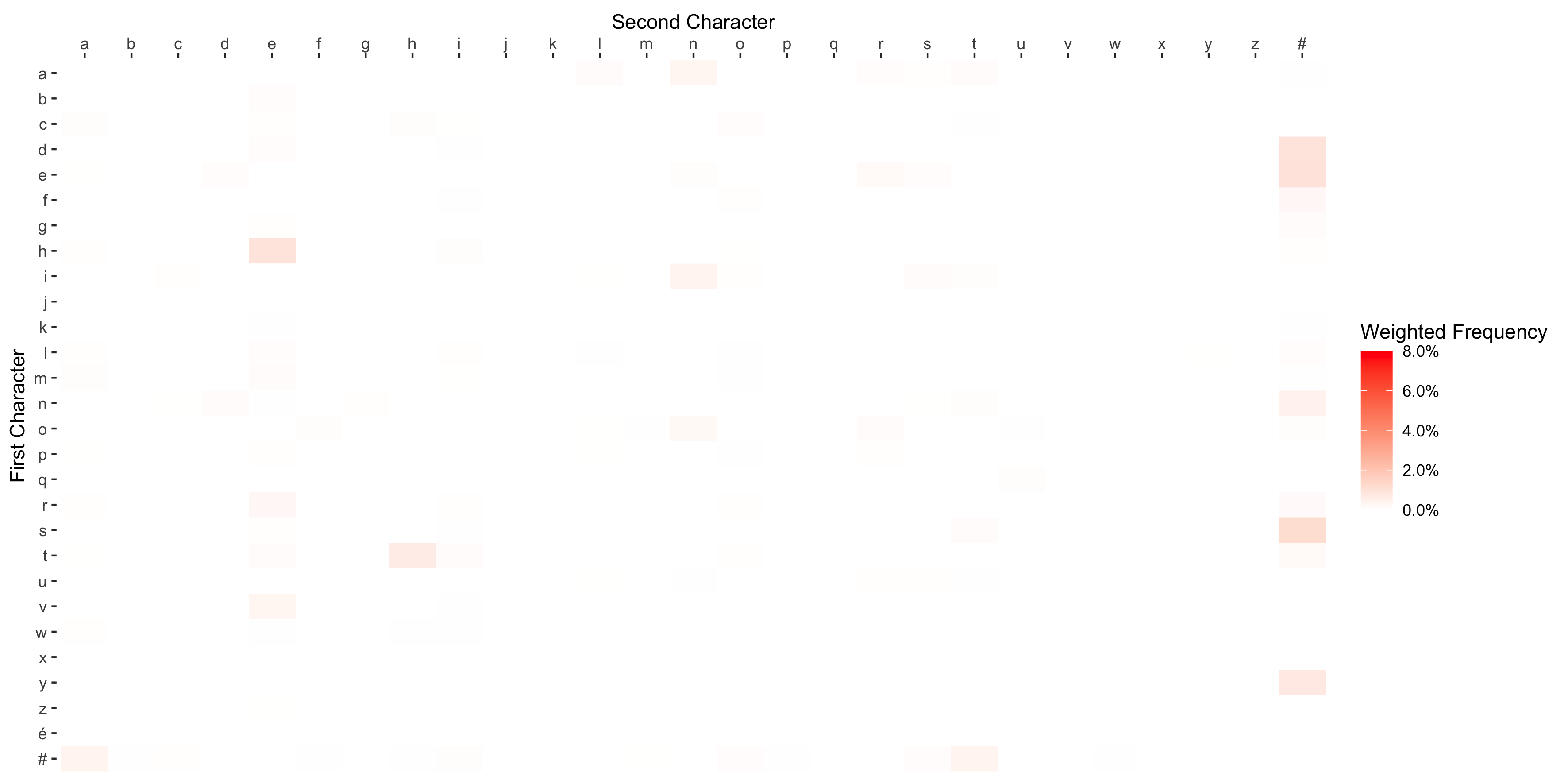}\centering
\caption{Conditional Frequency of English Bigrams Weighted by Overall Bigram Frequency.}
\label{en-cond-freq-wt}

\end{figure}

However, all the other letters in English are much less predictable. For example, the most common letter to follow \textit{p} is \textit{e}, but this only happens 17\% of the time. Conditional character entropy is the average for all letters weighted by their frequency, and this gives us a measure of the overall level of disorder in the text. A text consisting of randomly generated  characters has a higher conditional character entropy than a text that contains meaningful and ordered natural language. Voynichese, however, has starkly lower values than we see with any natural language, meaning many of its letters are like the English \textit{q} rather than \textit{p}.

Figures (\ref{en-cond-freq}) and (\ref{en-cond-freq-wt}) consist of two heatmaps of the English bigram space. The top map is simply the conditional frequency of each bigram. Bright spots indicate bigrams with particularly high conditional frequencies: \textit{qu}, \textit{y\#} (the \# symbol indicates a space, i.e., \textit{y} at the end of a word), \textit{ve} and \textit{d\#}. In the bottom map, each of these values have been weighted by the overall frequency of the bigram itself. This weighting gives a much better picture of which bigrams contribute to the conditional character entropy of the text as a whole. 

So while \textit{qu} is a highly predictable pairing, the letter \textit{q} itself is fairly rare, and therefore it has a negligible effect on the predictability of the text as a whole. Note from the right side of this map that certain letters at the end of the word (\textit{d\#, s\#, y\#}) make a contribution as a result of the frequent English morphological suffixes (\textit{-ed}, \textit{-s}, and \textit{-y}). Other bigrams like \textit{th}, \textit{he}, \textit{\#a}, and \textit{in} in the most frequently used words in English (\textit{the}, \textit{a}, \textit{in}). The bigram \textit{th} is a digraph, meaning that two characters are used to represent a single phoneme. Common digraphs can contribute to low conditional character entropy values because the two letters together share the information load of a single phoneme.  


\subsubsection{Compositionality, Word-finals, and Syllable Structure}

This is related to the issue of compositionality in the transcription of unknown texts like Voynichese. If certain glyphs occur primarily in a particular sequence, this may be evidence that the sequence of glyphs represents a single character. Thus Full Maximal Voynich (EVA), which is maximally decomposed, has a lower conditional entropy than Minimal Voynich, for which common glyph sequences are taken to be single characters. But even Minimal Voynich is much more predictable than any of the European languages. 

\begin{figure}[t!]

\includegraphics[scale=0.12,width=\linewidth]{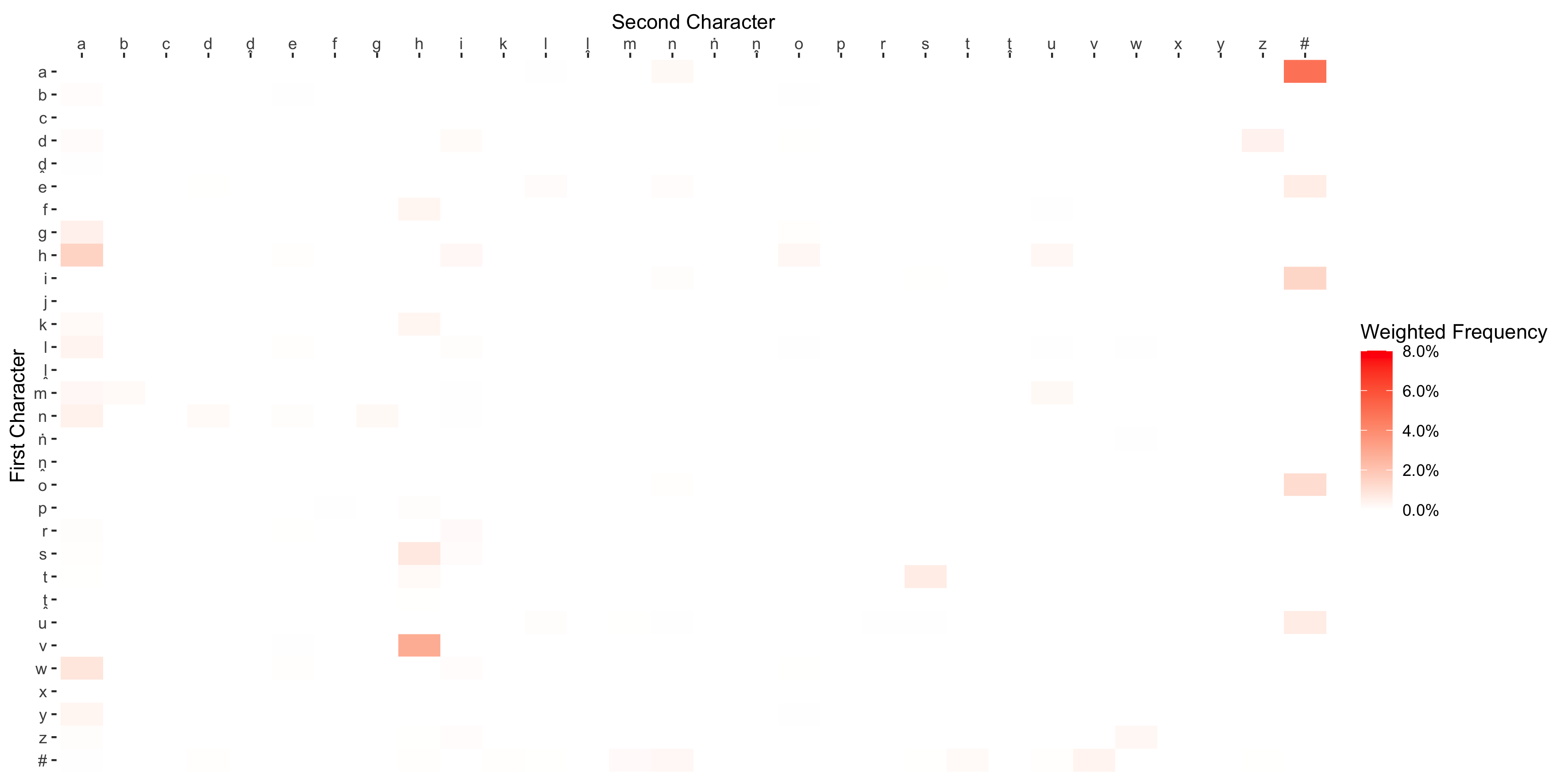}\centering
\caption{Conditional Frequency of Venda Bigrams Weighted by Overall Bigram Frequency}
\label{venda-cond-freq-wt}

\includegraphics[scale=0.12,width=\linewidth]{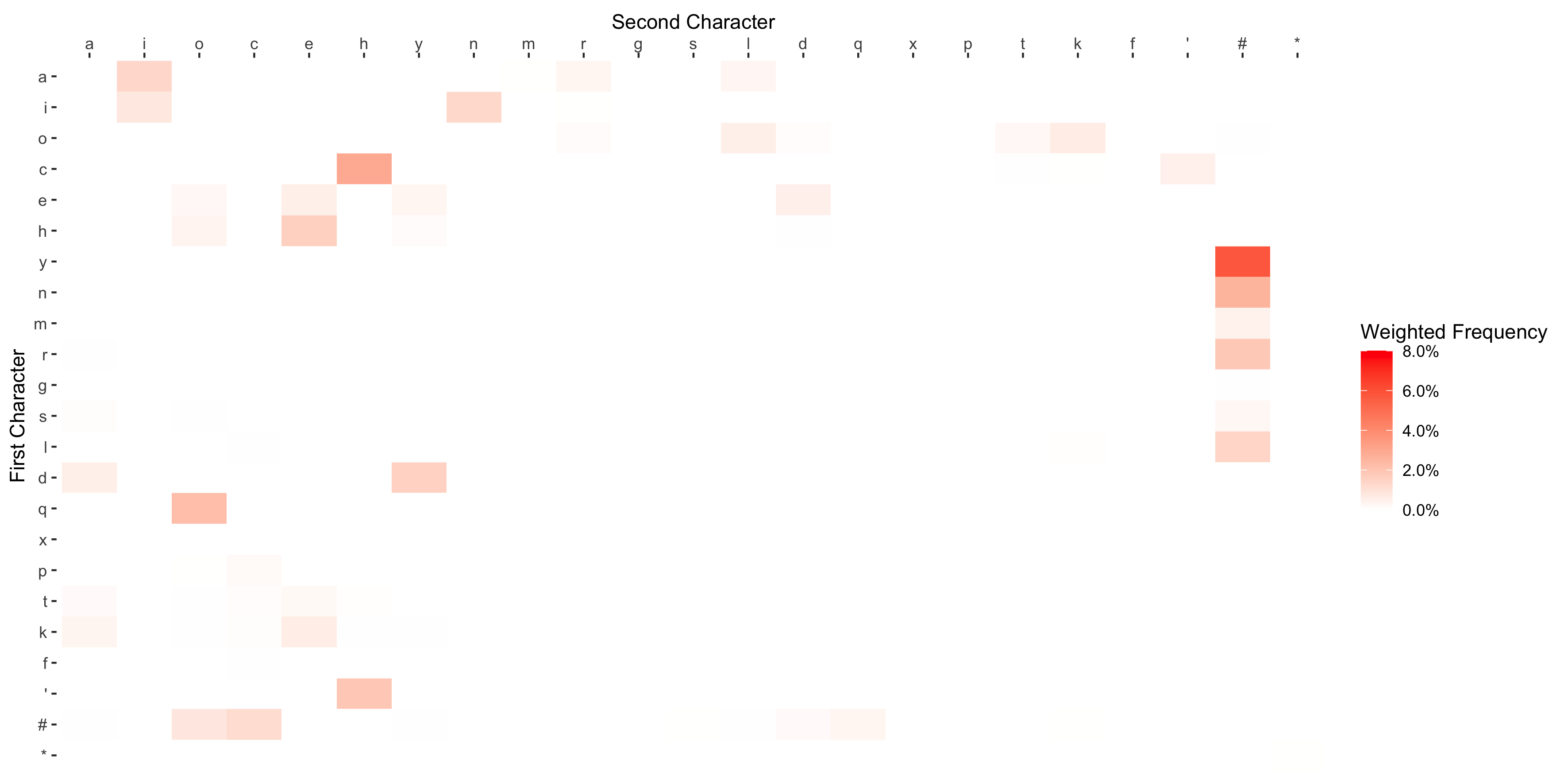}\centering
\caption{Conditional Frequency of (Simple Maximal) Voynich Bigrams Weighted by Overall Bigram Frequency}
\label{max-voy-cond-freq-wt}

\end{figure}

Compare the weighted heatmap for English at the bottom of Figure~\ref{en-cond-freq} with the weighted heatmap for Venda (a Southern Bantu language of South Africa) in Figure~\ref{venda-cond-freq-wt}. The Venda language has the second-lowest conditional character entropy of any non-Voynichese text in the corpora, and this added predictability is visible in the overall reddening of the heatmap. There are more bigrams in Venda which are both highly predictable and extremely frequent in the text. The \textit{vh} bigram is a digraph found in some of the most common words in the language. Most notably, the letter \textit{a} is very common at the end of a word, as are the other vowels. In fact, 97\% of all words in Venda end with a vowel, and nearly half (48\%) of all words end with an \textit{a}. 

This dramatic restriction of possible letters at the end of the word is common for languages with low conditional character entropy. For example, 98\% of words in the Hawaiian sample end with a vowel, and in Min Dong 95\% of words end with either a vowel, \textit{k}, or \textit{g}. This is due to restrictions on the possible syllable structures in these languages. While many Indo-European languages have fairly complex syllables with consonant clusters that can occur both at the beginning or end of the syllable, many languages of the Tibeto-Burman, Malayo-Polynesian, and Niger-Congo families do not allow syllables to end with a consonant, and they disallow many of the consonant clusters found in Indo-European languages. More complex syllables increase conditional character entropy because each consonant can be followed by a much larger number of possible consonants and vowels. Even among European languages, those that have more complex syllables, such as Slavic languages, tend to have slightly higher conditional character entropy. Abjads have even higher conditional character entropy because they lack written vowels altogether. 

Secondly, Venda, like most of the other lowest conditional-entropy languages, is tonal. Words in tonal languages that differ only by tone may have distinct meanings, and so most orthographies of these languages have a means of indicating tone. In tonal languages written with the Latin script (like Vietnamese), diacritics over the vowels are often employed to indicate tone. The orthography employed for the Venda sample, however, does not distinguish tone at all. This has the effect of collapsing distinctions that are present in the spoken language: two words which are pronounced differently may be spelled the same, and this makes the text more predictable. 

Voynichese has a lower conditional entropy than these other languages because it has even more frequent, highly predictable bigrams. Figure~\ref{max-voy-cond-freq-wt} maps the weighted conditional bigrams for (Simple) Maximal Voynich. As with Venda, certain characters are usually found at the end of words: 41\% of words end with \textit{y}, and 93\% of words end with either \text{y, n, l, r, m} or \textit{s}. Many other bigrams are prominent in other parts of the word: the \textit{ch} bench characters are usually found together at the beginning of the word (but sometimes have an intervening gallows character), \textit{qo} is found at the beginning of words, \textit{dy} is a very common sequence at the end of words, \textit{o} is almost always followed by \textit{l} or a gallows symbol (usually \textit{t} or \textit{k}), and \textit{i} is usually part of a word final sequence of \textit{in} or \textit{iin}. All characters are heavily restricted in whether they can appear at the beginning, middle, or end of the word, and which characters can come before or after. 

The unusual predictability of Voynichese cannot be entirely attributed to the compositionality of the transcription system. The Minimal transcription of Voynichese lacks many of these highly predictable bigrams, because common sequences like \textit{ch}, \textit{iin}, and \textit{qo} are represented as single characters (cf. Figure \ref{maxmin}). But most characters are still restricted to certain positions in the word: \textit{S, Z, Q, W, X} and \textit{Y} at the beginning, \textit{a, E, e, i, t, k, p} and \textit{f} in the middle, and \textit{N, M, 3, K, L, 5, T, U, 0, G H} and \textit{1} at the end. Thus one cannot simply assume that the low character entropy is due to our over-splitting of characters; even when they are grouped together, Voynichese is still unusual compared to other language samples.

\subsubsection{Bigrams with High Conditional Probability}

Another way to investigate the relationship between script properties and entropy is to measure the percentage of a text that contains bigrams with high conditional probability. Most texts have relatively few bigrams with a conditional probability greater than 50\%. In the English Wikipedia text, there are only four: \textit{qu} has 96\% conditional probability, \textit{y\#} has 75\%, \textit{ve} has 59\% conditional probability, and \textit{d\#} has 54\% conditional probability. Because most of these bigrams are relatively infrequent, they make up only 3.3\% of the text as a whole. By contrast, Voynich in the Simple Maximum transcription contains 12 bigrams with high conditional probability, and they are much more frequent, making up 29.3\% of the text. With the Minimal transcription, there are 23 bigrams with high conditional probability, and they make up 23.9\% of the text.

\begin{figure}[t!]
\includegraphics[scale=0.15,width=\linewidth]{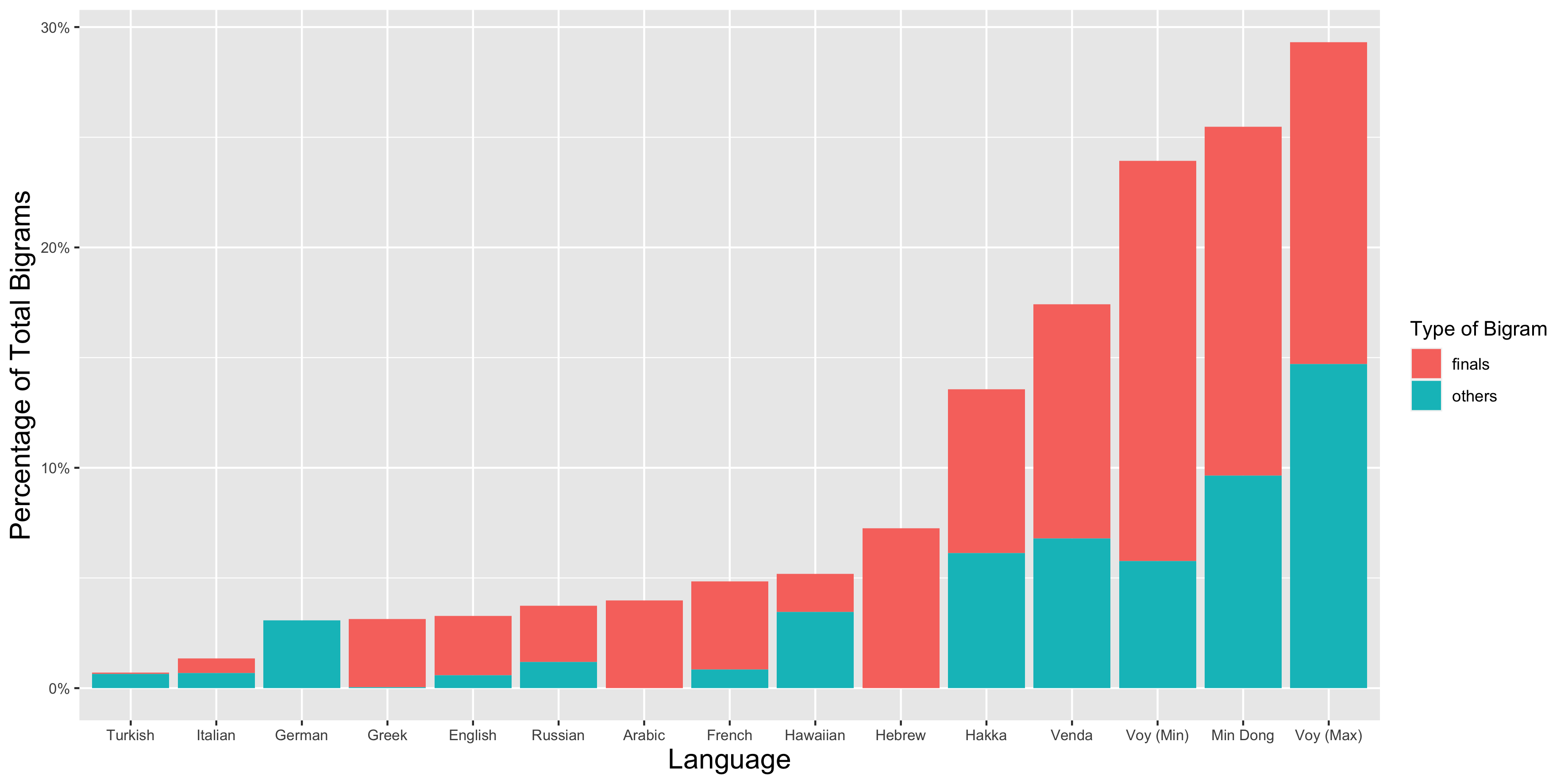}\centering
\caption{Prevalence of Bigrams with High Conditional Probability ($>$50\%) for Various Languages. Red indicates the percentage of word-final bigrams (e.g. \textit{y\#})}
\label{pred-bigrams}
\end{figure}

Figure~\ref{pred-bigrams} shows for various languages the proportion of the text which consists of bigrams with high conditional probability. The only other texts that have a comparable amount of high conditional bigrams are those of Min Dong, Hakka, Venda, and Hawaiian, the four languages in the Wikipedia corpus with the lowest conditional character entropy. Min Dong is between the two Voynich transcriptions. 

The red coloring denotes word-finals, i.e., characters which are followed by a space. In most languages of the sample, the majority of predictable bigrams are at the ends of words, whether or not their script contains characters which are found exclusively at the end of words (as with the positional character variants of Arabic, Hebrew, and Greek). This is true for English, because \textit{y\#} and \textit{d\#} are much more common than \textit{qu} and \textit{ve}. Voynichese, like Hakka and Min Dong, has a great deal of word-final bigrams in addition to other types of bigrams.

\subsection{Summary}

At the character level, Voynichese is strikingly different from any other text in the Wikipedia and Historical corpora. The character set size and frequency of characters is conventional, but the characters are combined in an extremely predictable way, as indicated by an unusual conditional character entropy that is distinctly lower than any of the 316 comparison texts. 

This discrepancy is not attributable to the transcription system used to encode Voynich, although decisions about the compositionality of glyph sequences can have a significant effect on entropy. Nor is it the result of conventional scholarly abbreviations of the historical period or the absence of written vowels. 

Rather, it is largely the result of common characters which are heavily restricted to certain positions within the word. Voynichese most closely resembles tonal languages written in the Latin script and languages with relatively limited syllabic inventories. 

We do not take this as evidence that the language underlying Voynich is likely to be from the Niger-Congo, Malayo-Polynesian, or Tibeto-Burman families. A more reasonable scenario is that the script ignores certain sound distinctions that are made in the underlying language, as Bennett's sample of Hawaiian did not include vowel distinctions and the Venda wikipedia text does not mark tone. Whatever method was used to generate the Voynichese script, it created written words that are highly constrained in form. 

We have described the creation of a corpus of Voynich texts and two comparison corpora for analyzing Voynichese from a broad typological perspective. In addition to the character-level analysis given here, we have used these corpora for our overview in \citet{bowernlindemann20} and will continue to use them in subsequent analyses. 

One line of inquiry that will be addressed in forthcoming work is the effect of historical encipherment methods on character properties. Here we will make a few brief observations. If Voynichese is encoded text, it must be more complex than a simple substitution cipher. A simple monoalphabetic substitution cipher will have absolutely no effect on conditional character entropy, because the same characters simply shift places with one another. The addition of null (meaningless) characters and multiple variants of high-frequency characters were used to make it harder to identify uniquely frequent characters. This can have an effect on character entropy (as can more complex polyalphabetic ciphers), but we would then expect the character frequency distribution of Voynichese to be atypical. They also affect the distribution of minimal pairs (if null-insertion operates on an otherwise monoalphabetic cipher). We can also rule out most types of polyalphabetic cipher, since such ciphers would disrupt the regular encoding of identical words across pages, as well as increasing (rather than decreasing) \emph{h2}.

Voynichese is a clear outlier at the character level, and we might be tempted to conclude from this that the text is meaningless. However, the Voynich text conceals sophisticated layers of structure. Subsequent work will examine the Voynich text at the word level, for which Voynichese is much more typical.

\pagebreak

\appendix

The following appendices provide some general information about the datasets used in this paper.

\section{Voynich Corpus Statistics}\label{voytables}

This table displays basic statistics for each of the Voynich sample texts in the Maximal transcription, broken down by running paragraph text and labels. \\

\noindent\begin{tabular}{| l | c | c | c | c | c | c |}

\hline
					& Character		& Character		& Word			& Word  		&	\textit{h2}	\\
					& Count			& Set Size			& Count			& Set Size 	&		\\
\hline\hline

\textbf{Full Voynich Text}		& 234,404		& 45				& 37,940			& 8,172		& 2.072		\\
\indent Paragraphs			& 205,014		& 36				& 33,111			& 6,936		& 2.117		\\
\indent Labels				& 29,389		& 36				& 4,829			& 2,283	 	& 2.309		\\

\hline\hline

\textbf{Voynich A}			& 68,612		& 36				& 11,415			& 3,460		& 2.122		\\
\indent Paragraphs			& 66,477		& 33				& 11,081			& 3,281		& 2.101		\\
\indent Labels				& 2,134		& 25				& 334			& 289		& 2.425	\\
\hline
\textbf{Voynich B}			& 145,745		& 32				& 23,226			& 4,947		& 1.973		\\
\indent Paragraphs			& 136,046		& 30				& 21,632			& 4,661		& 1.964		\\
\indent Labels				& 9,698		& 26				& 1,594			& 778		& 2.044		\\

\hline\hline

\textbf{Hand  1}				& 64,747		& 42				& 10,877			& 3,260		& 2.122		\\
\indent Paragraphs			& 61,963		& 33				& 10,352			& 3,032		& 2.083		\\
\indent Labels				& 2,783		& 31				& 525			& 365		& 2.572	\\
\hline
\textbf{Hand 2}				& 67,929		& 27				& 11,070			& 2,590		& 1.921		\\
\indent Paragraphs			& 61,698		& 27				& 10,054			& 2,367		& 1.910		\\
\indent Labels				& 6,230		& 21				& 1,016			& 531	 	& 1.975		\\
\hline
\textbf{Hand  3}				& 75,182		& 30				& 11,755			& 3,419		& 1.999		\\
\indent Paragraphs			& 72,550		& 30				& 11,328			& 3,302		& 1.991		\\
\indent Labels				& 2,631		& 22				& 427			& 294	 	& 2.086		\\
\hline
\textbf{Hand 4}				& 17,850		& 25				& 2,864			& 1,548		& 2.279		\\
\indent Paragraphs			& 2,219		& 21				& 353			& 268		& 2.083		\\
\indent Labels				& 15,630		& 25				& 2,511			& 1,399		& 2.284		\\
\hline
\textbf{Hand  5}				& 5,774		& 26				& 930			& 563		& 2.111		\\
\indent Paragraphs			& 3,662		& 22				& 580			& 387		& 2.079		\\
\indent Labels				& 2,111		& 26				& 350 			& 255		& 2.055		\\
\hline

\end{tabular} \\ \\

The following table compares the character set size and conditional character entropy (\textit{h2}) for each of the Voynich texts in each of the three transcription systems: Maximal, Maximal Simplified, and Minimal. \\

\noindent\begin{tabular}{| l | c | c | c |}

\hline
					& \textbf{Maximal}		& \textbf{Maximal Simplified}	& \textbf{Minimal}			 \\
\hline\hline

\textbf{Full Voynich}		& 45 / 2.114				& 23 / 2.112				& 41 / 2.475			 \\
\hline\hline

\textbf{Language A}		&  36 / 2.122				& 22 / 2.119				& 39 / 2.504				 \\
\hline
\textbf{Language B}		& 32 / 1.973				& 23 / 1.973				& 40 / 2.304 				 \\

\hline\hline

\textbf{Hand 1}		&  42 / 2.122				& 23 / 2.117					& 40 / 2.506			 \\
\hline
\textbf{Hand 2}		& 27 / 1.921				& 23 / 1.921					& 39 / 2.219 				 \\
\hline
\textbf{Hand 3}		&  30 / 1.999				& 23 / 1.999					& 39 / 2.338		 \\
\hline
\textbf{Hand 4}		& 25 / 2.279				& 22 / 2.279					& 36 / 2.558 				 \\
\hline
\textbf{Hand 5}		& 26 / 2.111				& 23 / 2.112					& 31 / 2.319			 \\
\hline
\end{tabular}\\\\

\section{Wikipedia Corpus Statistics}\label{wikitables}

The following table gives basic statistics for each of the sample languages in the Wikipedia Corpus:

\begin{scriptsize}
\setlength\LTleft{-1.5cm}
\renewcommand{\arraystretch}{1.5} 

\begin{longtable}{| l | c | c | c | c | c | c | c | c |}

\hline
\textbf{Language} & \textbf{Wikicode} & \textbf{Family} & \textbf{Script} & \textbf{Character} & \textbf{Character} & \textbf{Word} & \textbf{Word}  & \textbf{\textit{h2}} \\
 & & & & \textbf{Count} & \textbf{Set Size} & \textbf{Count} & \textbf{Set Size}  & \\
\hline\hline

\endfirsthead

\hline
\textbf{Language} & \textbf{Wikicode} & \textbf{Family} & \textbf{Script} & \textbf{Character} & \textbf{Character} & \textbf{Word} & \textbf{Word} & \textbf{\textit{h2}} \\
 & & & & \textbf{Count} & \textbf{Set Size} & \textbf{Count} & \textbf{Set Size} & \\
\hline\hline

\endhead
Abkhazian & ab & Caucasian & Cyrillic & 474,421 & 44 & 58,100 & 19,109 & 3.571 \\
\hline
Acehnese & ace & Malayo-Polynesian & Latin & 566,068 & 33 & 90,025 & 11,400 & 3.208 \\
\hline
Adyghe & ady & Caucasian & Cyrillic & 86,867 & 36 & 10,791 & 5,212 & 3.403 \\
\hline
Afrikaans & af & Germanic & Latin & 1,228,386 & 31 & 198,349 & 22,349 & 3.375 \\
\hline
Akan & ak & Niger-Congo & Latin & 409,746 & 31 & 82,731 & 9,578 & 3.321 \\
\hline
Albanian & sq & Albanian & Latin & 1,162,617 & 30 & 193,060 & 29,512 & 3.387 \\
\hline
Alemannic & als & Germanic & Latin & 1,197,089 & 38 & 197,012 & 39,254 & 3.542 \\
\hline
Amharic & am & Afro-Asiatic & Ethiopic & 940,578 & 293 & 193,388 & 49,646 & 4.503 \\
\hline
Anglo Saxon & ang & Germanic & Latin & 679,186 & 41 & 109,182 & 26,219 & 3.605 \\
\hline
Arabic & ar & Afro-Asiatic & Arabic & 1,164,523 & 44 & 195,629 & 38,372 & 3.702 \\
\hline
Aragonese & an & Romance & Latin & 1,170,029 & 34 & 210,697 & 23,412 & 3.293 \\
\hline
Aramaic & arc & Afro-Asiatic & Syriac & 52,659 & 40 & 9,236 & 3,853 & 3.483 \\
\hline
Armenian & hy & Armenian & Armenian & 1,521,970 & 40 & 194,344 & 36,578 & 3.376 \\
\hline
Aromanian & roa\_rup & Romance & Latin & 89,929 & 42 & 15,126 & 5,248 & 3.440 \\
\hline
Assamese & as & Indic & Bengali & 1,203,527 & 63 & 189,182 & 32,598 & 3.760 \\
\hline
Asturian & ast & Romance & Latin & 1,235,937 & 36 & 206,485 & 24,907 & 3.308 \\
\hline
Atikamekw & atj & Algonquian & Latin & 245,152 & 28 & 34,454 & 7,904 & 2.953 \\
\hline
Avar & av & Caucasian & Cyrillic & 705,663 & 37 & 88,140 & 25,138 & 3.355 \\
\hline
Awadhi & awa & Indic & Devanagari & 438,119 & 69 & 84,403 & 14,615 & 3.746 \\
\hline
Aymara & ay & Aymara & Latin & 202,890 & 39 & 23,822 & 11,645 & 3.185 \\
\hline
Azerbaijani & az & Turkic & Latin & 1,492,347 & 34 & 199,917 & 35,295 & 3.644 \\
\hline
Balinese & ban & Malayo-Polynesian & Latin & 1,317,387 & 27 & 199,768 & 25,568 & 3.180 \\
\hline
Bambara & bm & Mande & Latin & 122,358 & 38 & 25,163 & 4,806 & 3.191 \\
\hline
Banjar & bjn & Austronesian & Latin & 1,371,576 & 28 & 199,761 & 27,426 & 3.175 \\
\hline
Banyumasan & map\_bms & Malayo-Polynesian & Latin & 1,383,372 & 28 & 198,810 & 23,499 & 3.257 \\
\hline
Bashkir & ba & Turkic & Cyrillic & 1,412,766 & 42 & 194,138 & 37,466 & 3.691 \\
\hline
Basque & eu & Vasconic & Latin & 1,540,698 & 27 & 199,658 & 37,568 & 3.250 \\
\hline
Bavarian & bar & Germanic & Latin & 1,178,419 & 34 & 197,795 & 41,284 & 3.550 \\
\hline
Belarusian & be & Slavic & Cyrillic & 1,383,017 & 33 & 192,440 & 39,472 & 3.566 \\
\hline
Belarusian (Taraškievica) & be\_x\_old & Slavic & Cyrillic & 1,379,366 & 35 & 189,806 & 39,306 & 3.567 \\
\hline
Bengali & bn & Indic & Bengali & 1,301,543 & 62 & 191,876 & 28,481 & 3.665 \\
\hline
Bihari & bh & Indic & Devanagari & 940,050 & 68 & 189,977 & 18,150 & 3.618 \\
\hline
Bishnupriya Manipuri & bpy & Indic & Bengali & 517,177 & 63 & 79,490 & 16,408 & 3.545 \\
\hline
Bislama & bi & Creole & Latin & 16,041 & 29 & 3,145 & 798 & 3.126 \\
\hline
Bosnian & bs & Slavic & Latin & 1,305,485 & 33 & 195,997 & 37,659 & 3.468 \\
\hline
Breton & br & Celtic & Latin & 1,095,236 & 35 & 206,119 & 20,998 & 3.322 \\
\hline
Buginese (Buginese) & bug & Malayo-Polynesian & Buginese & 11,978 & 29 & 2,158 & 1,153 & 3.365 \\
\hline
Buginese (Latin) & bug & Malayo-Polynesian & Latin & 49,167 & 28 & 7,126 & 2,956 & 3.278 \\
\hline
Bulgarian & bg & Slavic & Cyrillic & 1,251,467 & 31 & 194,249 & 29,953 & 3.476 \\
\hline
Burmese & my & Tibeto-Burman & Myanmar & 1,002,038 & 62 & 75,527 & 44,643 & 3.324 \\
\hline
Buryat & bxr & Mongolic & Cyrillic & 1,418,044 & 35 & 192,042 & 38,617 & 3.503 \\
\hline
Cantonese & zh\_yue & Tibeto-Burman & Chinese & 322,674 & 3,608 & 31,888 & 28,439 & 5.409 \\
\hline
Catalan & ca & Romance & Latin & 1,189,313 & 41 & 209,944 & 23,100 & 3.370 \\
\hline
Cebuano & ceb & Malayo-Polynesian & Latin & 1,198,584 & 32 & 200,597 & 23,993 & 3.186 \\
\hline
Central Bicolano & bcl & Malayo-Polynesian & Latin & 1,197,161 & 35 & 197,619 & 23,546 & 3.212 \\
\hline
Chamorro & ch & Malayo-Polynesian & Latin & 37,146 & 34 & 6,547 & 1,903 & 3.291 \\
\hline
Chechen & ce & Caucasian & Cyrillic & 1,344,890 & 35 & 195,158 & 43,064 & 3.672 \\
\hline
Cherokee & chr & Iroquoian & Cherokee & 57,838 & 84 & 11,835 & 3,777 & 3.829 \\
\hline
Cheyenne & chy & Algonquian & Latin & 2,812 & 39 & 318 & 225 & 3.194 \\
\hline
Chichewa & ny & Niger-Congo & Latin & 689,446 & 28 & 99,634 & 17,337 & 3.246 \\
\hline
Chinese & zh & Tibeto-Burman & Chinese & 1,234,309 & 6,222 & 124,650 & 109,341 & 6.142 \\
\hline
Chuvash & cv & Turkic & Cyrillic & 1,184,848 & 38 & 188,153 & 35,546 & 3.620 \\
\hline
Classical Chinese & zh\_classical & Tibeto-Burman & Chinese & 80,593 & 3,292 & 12,778 & 11,912 & 5.176 \\
\hline
Cornish & kw & Celtic & Latin & 749,995 & 30 & 139,379 & 18,315 & 3.385 \\
\hline
Corsican & co & Romance & Latin & 1,134,720 & 33 & 207,593 & 28,600 & 3.232 \\
\hline
Cree & cr & Algonquian & Latin & 3,127 & 29 & 352 & 229 & 2.992 \\
\hline
Crimean Tatar & crh & Turkic & Latin & 397,729 & 35 & 54,339 & 16,048 & 3.589 \\
\hline
Croatian & hr & Slavic & Latin & 1,293,468 & 32 & 194,591 & 39,602 & 3.464 \\
\hline
Czech & cs & Slavic & Latin & 1,330,126 & 41 & 196,446 & 40,391 & 3.745 \\
\hline
Danish & da & Germanic & Latin & 1,258,211 & 30 & 196,664 & 28,753 & 3.476 \\
\hline
Dinka & din & Nilotic & Latin & 331,605 & 35 & 64,389 & 12,093 & 3.469 \\
\hline
Divehi & dv & Indic & Thaana & 1,732,631 & 51 & 186,755 & 41,357 & 3.138 \\
\hline
Doteli & dty & Indic & Devanagari & 1,170,968 & 71 & 187,576 & 38,443 & 3.824 \\
\hline
Dutch & nl & Germanic & Latin & 1,263,624 & 30 & 197,198 & 24,939 & 3.332 \\
\hline
Dutch Low Saxon & nds\_nl & Germanic & Latin & 1,191,824 & 35 & 199,215 & 31,151 & 3.393 \\
\hline
Dzongkha & dz & Tibeto-Burman & Tibetan & 93,231 & 63 & 24,147 & 1,735 & 3.076 \\
\hline
Egyptian Arabic & arz & Afro-Asiatic & Arabic & 1,078,766 & 47 & 195,437 & 36,930 & 3.663 \\
\hline
Emilian-Romagnol & eml & Romance & Latin & 784,411 & 53 & 162,961 & 23,869 & 3.525 \\
\hline
English & en & Germanic & Latin & 1,216,023 & 27 & 199,564 & 19,843 & 3.403 \\
\hline
Erzya & myv & Uralic & Cyrillic & 1,353,230 & 34 & 180,837 & 44,893 & 3.518 \\
\hline
Esperanto & eo & Constructed & Latin & 1,219,760 & 34 & 197,285 & 30,975 & 3.305 \\
\hline
Estonian & et & Uralic & Latin & 1,487,423 & 34 & 192,800 & 48,756 & 3.536 \\
\hline
Ewe & ee & Niger-Congo & Latin & 94,902 & 43 & 18,632 & 4,058 & 3.188 \\
\hline
Extremaduran & ext & Romance & Latin & 1,175,298 & 35 & 200,577 & 31,832 & 3.354 \\
\hline
Faroese & fo & Germanic & Latin & 1,181,388 & 34 & 191,824 & 29,506 & 3.523 \\
\hline
Fiji Hindi & hif & Indic & Latin & 706,696 & 34 & 128,450 & 15,253 & 3.411 \\
\hline
Fijian & fj & Malayo-Polynesian & Latin & 226,281 & 27 & 42,360 & 4,345 & 2.888 \\
\hline
Finnish & fi & Uralic & Latin & 1,728,271 & 28 & 196,573 & 55,335 & 3.475 \\
\hline
Franco-Provençal & frp & Romance & Latin & 314,875 & 41 & 58,163 & 12,054 & 3.461 \\
\hline
French & fr & Romance & Latin & 1,371,315 & 39 & 233,025 & 23,791 & 3.354 \\
\hline
Friulian & fur & Romance & Latin & 1,052,533 & 38 & 198,007 & 21,733 & 3.284 \\
\hline
Fula & ff & Niger-Congo & Latin & 238,729 & 37 & 40,601 & 10,899 & 3.364 \\
\hline
Gagauz & gag & Turkic & Latin & 484,481 & 38 & 67,724 & 19,070 & 3.570 \\
\hline
Galician & gl & Romance & Latin & 1,225,239 & 34 & 197,864 & 23,310 & 3.314 \\
\hline
Gan & gan & Tibeto-Burman & Chinese & 48,003 & 2,938 & 6,514 & 5,837 & 4.597 \\
\hline
Georgian & ka & Kartvelian & Georgian & 1,590,023 & 43 & 190,379 & 42,981 & 3.566 \\
\hline
German & de & Germanic & Latin & 1,411,555 & 31 & 197,197 & 32,028 & 3.394 \\
\hline
Gilaki & glk & Iranian & Arabic & 916,037 & 49 & 180,675 & 31,070 & 3.854 \\
\hline
Goan Konkani & gom & Indic & Devanagari & 1,317,991 & 76 & 197,428 & 48,971 & 3.659 \\
\hline
Gorontalo & gor & Malayo-Polynesian & Latin & 359,794 & 32 & 53,579 & 10,964 & 3.258 \\
\hline
Gothic & got & Germanic & Gothic & 63,136 & 28 & 9,352 & 3,598 & 3.321 \\
\hline
Greek & el & Hellenic & Greek & 1,270,133 & 34 & 192,045 & 26,770 & 3.519 \\
\hline
Greenlandic & kl & Inuit & Latin & 213,944 & 29 & 18,203 & 9,980 & 3.055 \\
\hline
Guarani & gn & Tupian & Latin & 1,388,881 & 48 & 198,012 & 32,287 & 3.435 \\
\hline
Guianan Creole & gcr & Creole & Latin & 610,232 & 31 & 115,926 & 12,096 & 3.224 \\
\hline
Gujarati & gu & Indic & Gujarati & 1,173,177 & 69 & 195,421 & 36,078 & 3.732 \\
\hline
Haitian & ht & Creole & Latin & 995,666 & 35 & 188,176 & 20,106 & 3.385 \\
\hline
Hakka & hak & Tibeto-Burman & Latin & 369,395 & 44 & 84,588 & 3,607 & 2.857 \\
\hline
Hausa & ha & Afro-Asiatic & Latin & 1,116,212 & 31 & 200,918 & 20,315 & 3.105 \\
\hline
Hawaiian & haw & Malayo-Polynesian & Latin & 561,237 & 31 & 112,870 & 5,728 & 2.765 \\
\hline
Hebrew & he & Afro-Asiatic & Hebrew & 1,123,178 & 28 & 194,947 & 38,482 & 3.664 \\
\hline
Hill Mari & mrj & Uralic & Cyrillic & 630,450 & 37 & 91,189 & 22,380 & 3.701 \\
\hline
Hindi & hi & Indic & Devanagari & 1,035,585 & 71 & 195,809 & 20,001 & 3.623 \\
\hline
Hungarian & hu & Uralic & Latin & 1,437,916 & 35 & 198,457 & 47,571 & 3.742 \\
\hline
Icelandic & is & Germanic & Latin & 1,237,880 & 36 & 194,966 & 33,368 & 3.596 \\
\hline
Ido & io & Constructed & Latin & 1,178,602 & 33 & 197,718 & 20,822 & 3.329 \\
\hline
Igbo & ig & Niger-Congo & Latin & 984,726 & 44 & 188,440 & 20,250 & 3.380 \\
\hline
Ilokano & ilo & Malayo-Polynesian & Latin & 1,172,160 & 28 & 198,139 & 20,266 & 3.163 \\
\hline
Inari Sami & smn & Uralic & Latin & 260,928 & 36 & 34,021 & 10,928 & 3.769 \\
\hline
Indonesian & id & Malayo-Polynesian & Latin & 1,446,218 & 26 & 201,308 & 17,949 & 3.222 \\
\hline
Ingush & inh & Caucasian & Cyrillic & 191,553 & 37 & 27,865 & 10,432 & 3.649 \\
\hline
Interlingua & ia & Constructed & Latin & 1,195,958 & 28 & 195,759 & 19,769 & 3.244 \\
\hline
Interlingue & ie & Constructed & Latin & 1,076,710 & 38 & 187,169 & 18,211 & 3.380 \\
\hline
Inuktitut (Canadian Syllabics) & iu & Inuit & Canadian Syllabics & 13,614 & 117 & 1,923 & 1,193 & 3.732 \\
\hline
Inuktitut (Latin) & iu & Inuit & Latin & 17,119 & 25 & 1,415 & 1,097 & 3.129 \\
\hline
Inupiak & ik & Inuit & Latin & 6,536 & 39 & 692 & 500 & 3.350 \\
\hline
Irish & ga & Celtic & Latin & 1,130,156 & 31 & 199,466 & 20,054 & 3.338 \\
\hline
Italian & it & Romance & Latin & 1,268,739 & 33 & 204,004 & 26,338 & 3.272 \\
\hline
Jamaican Patois & jam & Creole & Latin & 535,308 & 27 & 95,784 & 13,421 & 3.387 \\
\hline
Japanese & ja & Japonic & Japanese & 1,015,563 & 2,887 & 74,348 & 60,520 & 5.027 \\
\hline
Javanese & jv & Malayo-Polynesian & Latin & 1,351,976 & 29 & 199,101 & 27,774 & 3.238 \\
\hline
Kabardian Circassian & kbd & Caucasian & Cyrillic & 693,947 & 35 & 91,720 & 30,563 & 3.367 \\
\hline
Kabiye & kbp & Niger-Congo & Latin & 1,104,123 & 40 & 206,498 & 22,038 & 3.302 \\
\hline
Kabyle & kab & Afro-Asiatic & Latin & 1,130,890 & 39 & 210,877 & 30,148 & 3.458 \\
\hline
Kalmyk & xal & Mongolic & Cyrillic & 149,170 & 41 & 23,437 & 9,615 & 3.863 \\
\hline
Kannada & kn & Dravidian & Kannada & 1,633,412 & 66 & 198,531 & 57,666 & 3.773 \\
\hline
Kapampangan & pam & Malayo-Polynesian & Latin & 1,207,315 & 31 & 193,277 & 23,199 & 3.063 \\
\hline
Karachay-Balkar & krc & Turkic & Cyrillic & 1,138,595 & 34 & 145,979 & 31,559 & 3.476 \\
\hline
Karakalpak & kaa & Turkic & Latin & 958,773 & 37 & 121,182 & 32,916 & 3.509 \\
\hline
Kashmiri (Arabic) & ks & Indic & Arabic & 34,438 & 79 & 6,201 & 2,922 & 3.872 \\
\hline
Kashmiri (Devanagari) & ks & Indic & Devanagari & 22,493 & 69 & 3,641 & 2,093 & 3.864 \\
\hline
Kashubian & csb & Slavic & Latin & 792,280 & 40 & 120,564 & 37,608 & 3.750 \\
\hline
Kazakh & kk & Turkic & Cyrillic & 1,478,373 & 40 & 195,972 & 38,033 & 3.496 \\
\hline
Khmer & km & Austroasiatic & Khmer & 654,815 & 79 & 28,173 & 21,384 & 4.138 \\
\hline
Kikuyu & ki & Niger-Congo & Latin & 62,285 & 32 & 10,016 & 3,532 & 3.152 \\
\hline
Kinyarwanda & rw & Niger-Congo & Latin & 840,081 & 29 & 126,721 & 20,888 & 3.136 \\
\hline
Kirghiz & ky & Turkic & Cyrillic & 1,477,252 & 37 & 194,186 & 37,921 & 3.605 \\
\hline
Kirundi & rn & Niger-Congo & Latin & 218,081 & 28 & 32,282 & 8,974 & 3.074 \\
\hline
Komi & kv & Uralic & Cyrillic & 952,576 & 36 & 139,004 & 30,271 & 3.706 \\
\hline
Komi-Permyak & koi & Uralic & Cyrillic & 547,852 & 36 & 82,138 & 18,841 & 3.743 \\
\hline
Kongo & kg & Niger-Congo & Latin & 126,751 & 38 & 21,763 & 4,324 & 3.141 \\
\hline
Korean & ko & Koreanic & Hangul & 817,046 & 1,379 & 203,285 & 65,252 & 4.685 \\
\hline
Kotava & avk & Constructed & Latin & 1,160,948 & 30 & 190,260 & 33,067 & 3.454 \\
\hline
Kurdish & ku & Iranian & Latin & 1,097,852 & 35 & 201,217 & 29,586 & 3.386 \\
\hline
Ladin & lld & Romance & Latin & 597,615 & 46 & 114,922 & 18,314 & 3.394 \\
\hline
Ladino & lad & Romance & Latin & 1,115,534 & 32 & 193,345 & 26,514 & 3.305 \\
\hline
Lak & lbe & Caucasian & Cyrillic & 170,141 & 36 & 22,564 & 5,981 & 3.454 \\
\hline
Lao & lo & Tai & Lao & 343,956 & 55 & 21,063 & 13,620 & 4.030 \\
\hline
Latgalian & ltg & Baltic & Latin & 299,177 & 36 & 41,482 & 13,350 & 3.606 \\
\hline
Latin & la & Romance & Latin & 1,430,965 & 28 & 193,514 & 42,260 & 3.412 \\
\hline
Latvian & lv & Baltic & Latin & 1,408,947 & 36 & 193,369 & 36,694 & 3.634 \\
\hline
Lezgian & lez & Caucasian & Cyrillic & 1,426,123 & 33 & 194,005 & 32,609 & 3.411 \\
\hline
Ligurian & lij & Romance & Latin & 1,108,522 & 51 & 214,470 & 37,687 & 3.475 \\
\hline
Limburgish & li & Germanic & Latin & 1,193,438 & 34 & 201,569 & 27,862 & 3.436 \\
\hline
Lingala & ln & Niger-Congo & Latin & 383,704 & 46 & 64,511 & 11,314 & 3.222 \\
\hline
Lingua Franca Nova & lfn & Constructed & Latin & 1,022,406 & 28 & 196,299 & 15,063 & 3.111 \\
\hline
Lithuanian & lt & Baltic & Latin & 1,477,695 & 36 & 194,671 & 43,340 & 3.573 \\
\hline
Livvi-Karelian & olo & Uralic & Latin & 987,848 & 31 & 126,766 & 33,422 & 3.571 \\
\hline
Lojban & jbo & Constructed & Latin & 288,970 & 28 & 59,390 & 6,453 & 3.140 \\
\hline
Lombard & lmo & Romance & Latin & 1,036,268 & 41 & 208,682 & 29,793 & 3.444 \\
\hline
Low Saxon & nds & Germanic & Latin & 1,177,767 & 30 & 199,049 & 27,399 & 3.390 \\
\hline
Lower Sorbian & dsb & Slavic & Latin & 863,845 & 39 & 131,759 & 30,568 & 3.626 \\
\hline
Luganda & lg & Niger-Congo & Latin & 1,427,630 & 27 & 204,897 & 31,783 & 3.158 \\
\hline
Luxembourgisch & lb & Germanic & Latin & 1,254,155 & 33 & 200,807 & 28,191 & 3.423 \\
\hline
Macedonian & mk & Slavic & Cyrillic & 1,302,905 & 31 & 199,719 & 30,368 & 3.407 \\
\hline
Madurese & mad & Malayo-Polynesian & Latin & 312,666 & 34 & 46,468 & 8,560 & 3.394 \\
\hline
Maithili & mai & Indic & Devanagari & 1,105,898 & 73 & 186,573 & 34,716 & 3.831 \\
\hline
Malagasy & mg & Malayo-Polynesian & Latin & 1,245,462 & 34 & 186,592 & 21,030 & 2.944 \\
\hline
Malay & ms & Malayo-Polynesian & Latin & 1,406,260 & 27 & 198,741 & 18,313 & 3.186 \\
\hline
Malayalam & ml & Dravidian & Malayalam & 797,436 & 77 & 78,073 & 35,542 & 3.702 \\
\hline
Maltese & mt & Afro-Asiatic & Latin & 1,418,795 & 32 & 241,726 & 24,892 & 3.515 \\
\hline
Manx & gv & Celtic & Latin & 1,166,926 & 33 & 208,179 & 18,495 & 3.373 \\
\hline
Maori & mi & Malayo-Polynesian & Latin & 417,506 & 33 & 88,263 & 6,548 & 2.875 \\
\hline
Marathi & mr & Indic & Devanagari & 1,297,866 & 78 & 188,409 & 42,414 & 3.689 \\
\hline
Mazandarani & mzn & Iranian & Arabic & 993,445 & 46 & 193,871 & 24,773 & 3.660 \\
\hline
Meadow Mari & mhr & Uralic & Cyrillic & 1,371,590 & 38 & 199,570 & 34,921 & 3.549 \\
\hline
Min Dong & cdo & Tibeto-Burman & Latin & 430,592 & 48 & 101,336 & 2,796 & 2.841 \\
\hline
Min Nan & zh\_min\_nan & Tibeto-Burman & Latin & 1,267,107 & 49 & 309,129 & 8,318 & 3.066 \\
\hline
Minangkabau & min & Malayo-Polynesian & Latin & 1,083,848 & 27 & 158,742 & 19,545 & 3.181 \\
\hline
Mingrelian & xmf & Kartvelian & Georgian & 1,452,488 & 40 & 187,555 & 49,075 & 3.580 \\
\hline
Mirandese & mwl & Romance & Latin & 1,210,157 & 33 & 203,845 & 22,829 & 3.309 \\
\hline
Moksha & mdf & Moksha & Cyrillic & 254,066 & 34 & 34,186 & 12,397 & 3.474 \\
\hline
Mon & mnw & Austroasiatic & Myanmar & 510,675 & 74 & 42,496 & 24,496 & 3.570 \\
\hline
Mongolian & mn & Mongolic & Cyrillic & 1,321,073 & 35 & 193,476 & 29,282 & 3.559 \\
\hline
Moroccan Arabic & ary & Afro-Asiatic & Arabic & 792,354 & 46 & 143,786 & 29,576 & 3.725 \\
\hline
N'Ko & nqo & Mande & N'Ko & 1,122,166 & 48 & 196,488 & 22,518 & 2.970 \\
\hline
Nahuatl & nah & Uto-Aztecan & Latin & 508,194 & 36 & 62,805 & 15,605 & 3.227 \\
\hline
Nauruan & na & Malayo-Polynesian & Latin & 59,243 & 51 & 9,774 & 2,819 & 3.405 \\
\hline
Navajo & nv & Athabaskan & Latin & 202,428 & 37 & 28,125 & 7,125 & 3.366 \\
\hline
Neapolitan & nap & Romance & Latin & 1,031,893 & 40 & 185,831 & 32,261 & 3.291 \\
\hline
Nepali & ne & Indic & Devanagari & 1,170,229 & 68 & 179,096 & 36,637 & 3.770 \\
\hline
Newar & new & Tibeto-Burman & Devanagari & 770,077 & 76 & 122,917 & 27,822 & 3.750 \\
\hline
Nias & nia & Malayo-Polynesian & Latin & 293,580 & 29 & 52,144 & 7,274 & 3.077 \\
\hline
Norfolk & pih & Creole & Latin & 114,354 & 34 & 21,202 & 5,008 & 3.497 \\
\hline
Norman & nrm & Romance & Latin & 877,907 & 41 & 165,642 & 20,754 & 3.364 \\
\hline
North Frisian & frr & Germanic & Latin & 1,073,028 & 36 & 191,321 & 26,674 & 3.499 \\
\hline
Northern Sami & se & Uralic & Latin & 793,557 & 41 & 100,856 & 26,797 & 3.607 \\
\hline
Northern Sotho & nso & Niger-Congo & Latin & 137,660 & 30 & 25,626 & 4,583 & 2.950 \\
\hline
Norwegian (Bokma\r{a}l) & no & Germanic & Latin & 1,247,199 & 32 & 196,357 & 29,556 & 3.475 \\
\hline
Norwegian (Nynorsk) & nn & Germanic & Latin & 1,185,458 & 32 & 193,537 & 27,119 & 3.496 \\
\hline
Novial & nov & Constructed & Latin & 220,177 & 33 & 38,318 & 8,423 & 3.325 \\
\hline
Occitan & oc & Romance & Latin & 1,238,392 & 38 & 208,924 & 20,667 & 3.358 \\
\hline
Old Church Slavonic & cu & Slavic & Cyrillic & 68,106 & 50 & 10,108 & 3,258 & 3.398 \\
\hline
Oriya & or & Indic & Odia & 1,249,524 & 66 & 184,928 & 32,844 & 3.742 \\
\hline
Oromo & om & Afro-Asiatic & Latin & 1,625,327 & 29 & 233,367 & 40,522 & 3.193 \\
\hline
Ossetian & os & Iranian & Cyrillic & 1,198,686 & 33 & 194,122 & 34,753 & 3.651 \\
\hline
Palatinate German & pfl & Germanic & Latin & 1,250,014 & 38 & 197,124 & 42,618 & 3.440 \\
\hline
Pali (Devanagari) & pi & Indic & Devanagari & 7,512 & 62 & 1,157 & 686 & 3.694 \\
\hline
Pali (Latin) & pi & Indic & Latin & 56,922 & 32 & 5,971 & 1,953 & 2.905 \\
\hline
Pangasinan & pag & Malayo-Polynesian & Latin & 176,803 & 27 & 31,387 & 6,904 & 3.166 \\
\hline
Papiamentu & pap & Creole & Latin & 1,049,584 & 38 & 195,235 & 20,194 & 3.344 \\
\hline
Pashto & ps & Iranian & Arabic & 892,610 & 52 & 196,694 & 25,578 & 3.651 \\
\hline
Pennsylvania German & pdc & Germanic & Latin & 226,954 & 31 & 37,682 & 8,436 & 3.381 \\
\hline
Persian & fa & Iranian & Arabic & 1,024,245 & 42 & 197,880 & 19,807 & 3.651 \\
\hline
Picard & pcd & Romance & Latin & 543,922 & 39 & 102,478 & 17,963 & 3.397 \\
\hline
Piedmontese & pms & Romance & Latin & 1,062,707 & 35 & 218,355 & 19,308 & 3.318 \\
\hline
Polish & pl & Slavic & Latin & 1,410,675 & 35 & 195,119 & 44,911 & 3.638 \\
\hline
Pontic & pnt & Hellenic & Greek & 127,501 & 38 & 20,894 & 6,214 & 3.534 \\
\hline
Portuguese & pt & Romance & Latin & 1,235,214 & 39 & 198,672 & 22,429 & 3.316 \\
\hline
Punjabi & pa & Indic & Gurmukhi & 970,590 & 64 & 195,713 & 22,171 & 3.577 \\
\hline
Quechua & qu & Quechua & Latin & 335,139 & 35 & 40,329 & 14,685 & 3.197 \\
\hline
Ripuarian & ksh & Germanic & Latin & 1,110,517 & 35 & 198,243 & 34,554 & 3.556 \\
\hline
Romani & rmy & Indic & Latin & 91,925 & 38 & 15,487 & 4,116 & 3.294 \\
\hline
Romanian & ro & Romance & Latin & 1,348,034 & 31 & 210,279 & 28,217 & 3.416 \\
\hline
Romansh & rm & Romance & Latin & 1,203,902 & 34 & 208,626 & 18,734 & 3.306 \\
\hline
Russian & ru & Slavic & Cyrillic & 1,500,801 & 34 & 200,538 & 40,697 & 3.619 \\
\hline
Rusyn & rue & Slavic & Cyrillic & 1,244,127 & 38 & 187,752 & 48,078 & 3.740 \\
\hline
Saaraiki & skr & Indic & Arabic & 927,476 & 57 & 197,187 & 23,313 & 3.679 \\
\hline
Sakha & sah & Turkic & Cyrillic & 1,488,350 & 39 & 191,813 & 44,495 & 3.617 \\
\hline
Sakizaya & szy & Austronesian & Latin & 1,067,950 & 27 & 194,641 & 13,539 & 2.948 \\
\hline
Samoan & sm & Malayo-Polynesian & Latin & 399,484 & 32 & 83,345 & 6,083 & 2.992 \\
\hline
Samogitian & bat\_smg & Baltic & Latin & 719,074 & 36 & 105,660 & 34,440 & 3.599 \\
\hline
Sango & sg & Niger-Congo & Latin & 17,054 & 38 & 3,746 & 640 & 2.862 \\
\hline
Sanskrit & sa & Indic & Devanagari & 1,507,088 & 76 & 177,498 & 60,631 & 3.728 \\
\hline
Santali & sat & Austroasiatic & Ol Chiki & 1,013,781 & 37 & 187,788 & 19,498 & 3.391 \\
\hline
Sardinian & sc & Romance & Latin & 1,157,097 & 33 & 204,144 & 32,653 & 3.230 \\
\hline
Saterland Frisian & stq & Germanic & Latin & 1,146,938 & 30 & 186,897 & 25,801 & 3.384 \\
\hline
Scots & sco & Germanic & Latin & 1,122,338 & 27 & 199,515 & 22,474 & 3.417 \\
\hline
Scottish Gaelic & gd & Celtic & Latin & 1,137,203 & 33 & 206,651 & 18,580 & 3.145 \\
\hline
Serbian & sr & Slavic & Cyrillic & 1,241,960 & 31 & 190,415 & 38,120 & 3.458 \\
\hline
Serbo-Croatian & sh & Slavic & Latin & 1,271,863 & 31 & 192,522 & 40,828 & 3.472 \\
\hline
Sesotho & st & Niger-Congo & Latin & 303,739 & 28 & 55,756 & 7,713 & 3.004 \\
\hline
Shan & shn & Tai & Myanmar & 494,159 & 105 & 31,048 & 22,064 & 3.154 \\
\hline
Shona & sn & Niger-Congo & Latin & 626,613 & 27 & 76,706 & 22,057 & 3.169 \\
\hline
Sicilian & scn & Romance & Latin & 1,198,370 & 38 & 204,500 & 29,579 & 3.214 \\
\hline
Silesian & szl & Slavic & Latin & 891,117 & 46 & 132,556 & 42,331 & 3.767 \\
\hline
Simple English & simple & Germanic & Latin & 1,128,909 & 27 & 198,481 & 14,313 & 3.362 \\
\hline
Sindhi & sd & Indic & Arabic & 875,905 & 62 & 184,506 & 20,142 & 3.694 \\
\hline
Sinhalese & si & Indic & Sinhala & 1,182,228 & 76 & 188,511 & 34,716 & 3.799 \\
\hline
Slovak & sk & Slavic & Latin & 1,343,586 & 42 & 195,932 & 42,027 & 3.737 \\
\hline
Slovenian & sl & Slavic & Latin & 1,039,854 & 31 & 161,193 & 29,227 & 3.516 \\
\hline
Somali & so & Afro-Asiatic & Latin & 1,245,451 & 28 & 194,843 & 30,491 & 3.268 \\
\hline
Sorani & ckb & Iranian & Arabic & 1,220,535 & 39 & 194,694 & 38,589 & 3.452 \\
\hline
South Azerbaijani & azb & Turkic & Arabic & 1,325,401 & 49 & 194,657 & 50,268 & 3.717 \\
\hline
Spanish & es & Romance & Latin & 1,251,204 & 33 & 202,983 & 23,801 & 3.291 \\
\hline
Sranan & srn & Creole & Latin & 61,506 & 35 & 12,032 & 2,520 & 3.185 \\
\hline
Sundanese & su & Malayo-Polynesian & Latin & 703,485 & 28 & 103,181 & 14,482 & 3.368 \\
\hline
Swahili & sw & Niger-Congo & Latin & 1,233,964 & 26 & 195,587 & 21,695 & 3.084 \\
\hline
Swati & ss & Niger-Congo & Latin & 201,303 & 29 & 23,888 & 10,554 & 3.195 \\
\hline
Swedish & sv & Germanic & Latin & 1,301,628 & 31 & 198,498 & 32,104 & 3.539 \\
\hline
Tagalog & tl & Malayo-Polynesian & Latin & 1,213,045 & 27 & 196,218 & 19,488 & 3.059 \\
\hline
Tahitian & ty & Malayo-Polynesian & Latin & 16,927 & 43 & 3,858 & 694 & 2.846 \\
\hline
Tajik & tg & Iranian & Cyrillic & 1,248,236 & 40 & 190,743 & 34,147 & 3.614 \\
\hline
Tamil & ta & Dravidian & Tamil & 1,783,419 & 49 & 192,511 & 58,671 & 3.327 \\
\hline
Tarantino & roa\_tara & Romance & Latin & 1,148,683 & 35 & 205,377 & 21,844 & 3.109 \\
\hline
Tatar (Cyrillic) & tt & Turkic & Cyrillic & 940,273 & 38 & 128,935 & 27,004 & 3.670 \\
\hline
Tatar (Latin) & tt & Turkic & Latin & 429,020 & 41 & 60,991 & 18,904 & 3.749 \\
\hline
Telugu & te & Dravidian & Telugu & 1,551,065 & 67 & 193,017 & 56,921 & 3.783 \\
\hline
Tetum & tet & Malayo-Polynesian & Latin & 660,727 & 37 & 108,003 & 14,469 & 3.446 \\
\hline
Thai & th & Tai & Thai & 953,229 & 68 & 38,480 & 27,031 & 4.247 \\
\hline
Tibetan & bo & Tibeto-Burman & Tibetan & 1,655,843 & 83 & 428,495 & 6,127 & 3.143 \\
\hline
Tigrinya & ti & Afro-Asiatic & Ethiopic & 96,980 & 230 & 21,081 & 7,467 & 4.401 \\
\hline
Tok Pisin & tpi & Creole & Latin & 87,118 & 27 & 16,046 & 2,136 & 3.135 \\
\hline
Tongan & to & Malayo-Polynesian & Latin & 167,431 & 34 & 36,884 & 3,972 & 2.945 \\
\hline
Tsonga & ts & Niger-Congo & Latin & 522,298 & 28 & 88,346 & 11,540 & 2.970 \\
\hline
Tswana & tn & Niger-Congo & Latin & 1,068,169 & 28 & 203,475 & 16,097 & 2.816 \\
\hline
Tulu & tcy & Dravidian & Kannada & 1,243,356 & 68 & 168,376 & 51,885 & 3.695 \\
\hline
Tumbuka & tum & Niger-Congo & Latin & 62,879 & 29 & 9,480 & 2,803 & 2.979 \\
\hline
Turkish & tr & Turkic & Latin & 1,503,312 & 34 & 202,601 & 39,777 & 3.549 \\
\hline
Turkmen & tk & Turkic & Latin & 1,510,318 & 37 & 202,877 & 43,688 & 3.603 \\
\hline
Tuvan & tyv & Turkic & Cyrillic & 1,451,621 & 40 & 208,827 & 34,787 & 3.494 \\
\hline
Twi & tw & Niger-Congo & Latin & 173,715 & 30 & 35,061 & 5,612 & 3.289 \\
\hline
Udmurt & udm & Uralic & Cyrillic & 539,349 & 40 & 75,154 & 19,440 & 3.669 \\
\hline
Ukrainian & uk & Slavic & Cyrillic & 1,401,545 & 34 & 193,133 & 40,936 & 3.653 \\
\hline
Upper Sorbian & hsb & Slavic & Latin & 1,272,611 & 38 & 190,832 & 32,500 & 3.559 \\
\hline
Urdu & ur & Indic & Arabic & 912,459 & 52 & 195,713 & 15,390 & 3.575 \\
\hline
Uyghur & ug & Turkic & Arabic & 1,257,845 & 39 & 164,208 & 31,147 & 3.479 \\
\hline
Uzbek & uz & Turkic & Latin & 1,548,096 & 29 & 197,038 & 37,078 & 3.436 \\
\hline
Venda & ve & Niger-Congo & Latin & 99,874 & 30 & 17,566 & 3,452 & 2.792 \\
\hline
Venetian & vec & Romance & Latin & 1,046,029 & 37 & 200,507 & 30,156 & 3.265 \\
\hline
Vepsian & vep & Uralic & Latin & 1,366,394 & 30 & 179,658 & 23,467 & 3.501 \\
\hline
Vietnamese & vi & Austroasiatic & Latin & 901,129 & 92 & 200,641 & 6,639 & 3.168 \\
\hline
Volapük & vo & Constructed & Latin & 1,001,117 & 30 & 163,340 & 14,480 & 3.237 \\
\hline
Võro & fiu\_vro & Uralic & Latin & 829,521 & 34 & 122,791 & 31,833 & 3.635 \\
\hline
Walloon & wa & Romance & Latin & 1,006,614 & 37 & 197,108 & 20,966 & 3.379 \\
\hline
Waray-Waray & war & Malayo-Polynesian & Latin & 1,170,812 & 32 & 200,277 & 19,247 & 3.133 \\
\hline
Welsh & cy & Celtic & Latin & 1,125,686 & 32 & 204,242 & 19,707 & 3.465 \\
\hline
West Flemish & vls & Germanic & Latin & 1,123,656 & 32 & 200,520 & 30,198 & 3.434 \\
\hline
West Frisian & fy & Germanic & Latin & 1,127,112 & 33 & 194,755 & 21,690 & 3.374 \\
\hline
Western Armenian & hyw & Armenian & Armenian & 1,406,809 & 41 & 196,090 & 42,235 & 3.479 \\
\hline
Western Punjabi & pnb & Indic & Arabic & 922,341 & 50 & 196,296 & 18,042 & 3.595 \\
\hline
Wolof & wo & Niger-Congo & Latin & 983,997 & 34 & 203,888 & 11,051 & 3.292 \\
\hline
Wu & wuu & Tibeto-Burman & Chinese & 54,268 & 3,187 & 6,303 & 5,755 & 4.388 \\
\hline
Xhosa & xh & Niger-Congo & Latin & 1,064,361 & 29 & 130,727 & 41,754 & 3.267 \\
\hline
Yiddish & yi & Germanic & Hebrew & 1,133,439 & 30 & 199,605 & 18,652 & 3.311 \\
\hline
Yoruba & yo & Niger-Congo & Latin & 683,230 & 43 & 153,063 & 18,010 & 3.591 \\
\hline
Zamboanga Chavacano & cbk\_zam & Creole & Latin & 1,147,970 & 33 & 195,097 & 24,341 & 3.310 \\
\hline
Zazaki & diq & Iranian & Latin & 1,093,565 & 38 & 198,835 & 35,611 & 3.528 \\
\hline
Zeelandic & zea & Germanic & Latin & 1,135,510 & 36 & 204,440 & 29,827 & 3.388 \\
\hline
Zhuang & za & Tai & Latin & 28,961 & 29 & 4,509 & 1,599 & 3.069 \\
\hline
Zulu & zu & Niger-Congo & Latin & 866,234 & 28 & 102,758 & 33,905 & 3.262 \\
\hline
\end{longtable}

\end{scriptsize}

\newpage

\section{Historical Corpus Statistics}\label{histtables}

The following table gives basic statistics for each of the Historical Texts, along with separate statistics for parallel texts with normalized and diplomatic versions. \\

\noindent\begin{tabular}{| l | c | c | c | c | c | c | c |}

\hline
Text		& Language & Script & Character		& Character		& Word			& Word  		&	\textit{h2}	\\
		&		&	& Count			& Set Size			& Count			& Set Size 	&		\\
\hline\hline
\textit{Medical Casebooks} & English & Latin &  &  &  &  &  \\
\indent Normalized & 	 &  & 3,057 & 884 & 15,195 & 41 & 3.418 \\
\indent Diplomatic &  &  & 3,069 & 901 & 14,646 & 54 & 3.435 \\
\hline
\textit{Three Books of} & English & Latin & 56,914 & 6,675 & 314,259 & 28 & 3.241 \\
\textit{Occult Philosophy} &  &  &  &  &  &  &  \\
\hline
\textit{Science of Cirurgie} & English & Latin & 97,949 & 7,443 & 483,823 & 30 & 3.240 \\
\hline
\textit{Secretum Secretorum} & English & Latin & 18,350 & 2,754 & 96,373 & 27 & 3.111 \\
\hline
\textit{Alphabet of Tales} & English & Latin & 177,763 & 14,083 & 891,000 & 33 & 3.254 \\
\hline
\textit{Amiran-Darejaniani} & Georgian & Georgian & 45,169 & 12,888 & 336,149 & 37 & 3.420 \\
\hline
\textit{Mishneh Torah} & Hebrew & Hebrew & 27,261 & 7,857 & 143,516 & 29 & 3.637 \\
\hline
\textit{Masoretic Tanakh (Bereshit)}  & Hebrew & Hebrew &  &  &  &  &  \\
\indent Without \textit{niqqud} &  &  & 17,802 & 6,327 & 99,024 & 29 & 3.526 \\
\indent With \textit{niqqud} &  &  & 17,802 & 7,091 & 164,569 & 43 & 3.256 \\
\hline
\textit{Codex Wormianus} & Icelandic & Latin &  &  &  &  &  \\
\indent Normalized  &  &  & 31,592 & 8,393 & 162,890 & 48 & 3.390 \\
\indent Diplomatic &  &  & 31,442 & 8,637 & 150,374 & 60 & 3.490 \\
\hline
\textit{La Rettorica} & Italian & Latin & 32,230 & 6,789 & 198,138 & 29 & 3.141 \\
\hline
\textit{Necrologium Lundense} & Latin & Latin &  &  &  &  &  \\
 \indent Normalized &  &  & 309 & 222 & 1,723 & 72 & 3.348 \\
 \indent Diplomatic &  &  & 314 & 194 & 2,046 & 41 & 3.204 \\
\hline
\textit{De Ortu Et Tempo} & Latin & Latin & 1,939 & 975 & 13,078 & 24 & 3.252 \\
\textit{Antichristi}		&		&	&	&	&	&	&	\\
\hline
\textit{Historia Hierosylmitanae} & Latin & Latin & 125,987 & 20,082 & 900,781 & 27 & 3.368 \\
\textit{Expeditionis}			&	&		&		&		&		&	&	\\
\hline
\textit{De Magia} & Latin & Latin & 11,790 & 4,067 & 81,028 & 24 & 3.315 \\
\hline
\textit{Secretum Secretorum} & Latin & Latin & 39,349 & 9,294 & 262,206 & 25 & 3.277 \\
\hline
\textit{Steganographia} & Latin & Latin & 21,529 & 7,559 & 154,739 & 33 & 3.424 \\
\hline
\textit{Sindbad-Name} & Persian & Arabic & 19,751 & 8,672 & 103,002 & 68 & 3.871 \\
\hline
\textit{Picatrix} & Spanish & Latin & 110,684 & 19,420 & 642,787 & 32 & 3.244 \\
\hline

\end{tabular}

\newpage
\section{The Sukhotin Algorithm for Vowel Detection}

The Sukhotin Algorithm is a procedure for determining which characters of an encoded text are vowels (\citealt{sukhotin1962eksperimental}, \citealt{guy1991vowel}). \citet{guy1991statistical} applied the algorithm to two folios of the Voynich text. For further discussion of the Sukhotin algorithm and its usage in Voynich analysis, see \citet{bowernlindemann20}.

The vowel determination for Voynichese is slightly different between Voynich languages, and is heavily dependent upon the transcription system. Here we present the vowel determination results of the Sukhotin Algorithm for multiple languages and scripts in comparison to Voynichese. We include two results: one in which spaces between words are included as a separate character (designated by the \# symbol), and one in which they are ignored. If spaces are included, they are almost always identified first by the algorithm, and the overall results are better. In most cases, the exclusion of spaces produces a similar result, but may include consonants that are `vowel-like' (e.g. \textit{y}) or tend to be found at the beginning or end of words.

\begin{figure}[h!]
\includegraphics[scale=0.4,width=\linewidth]{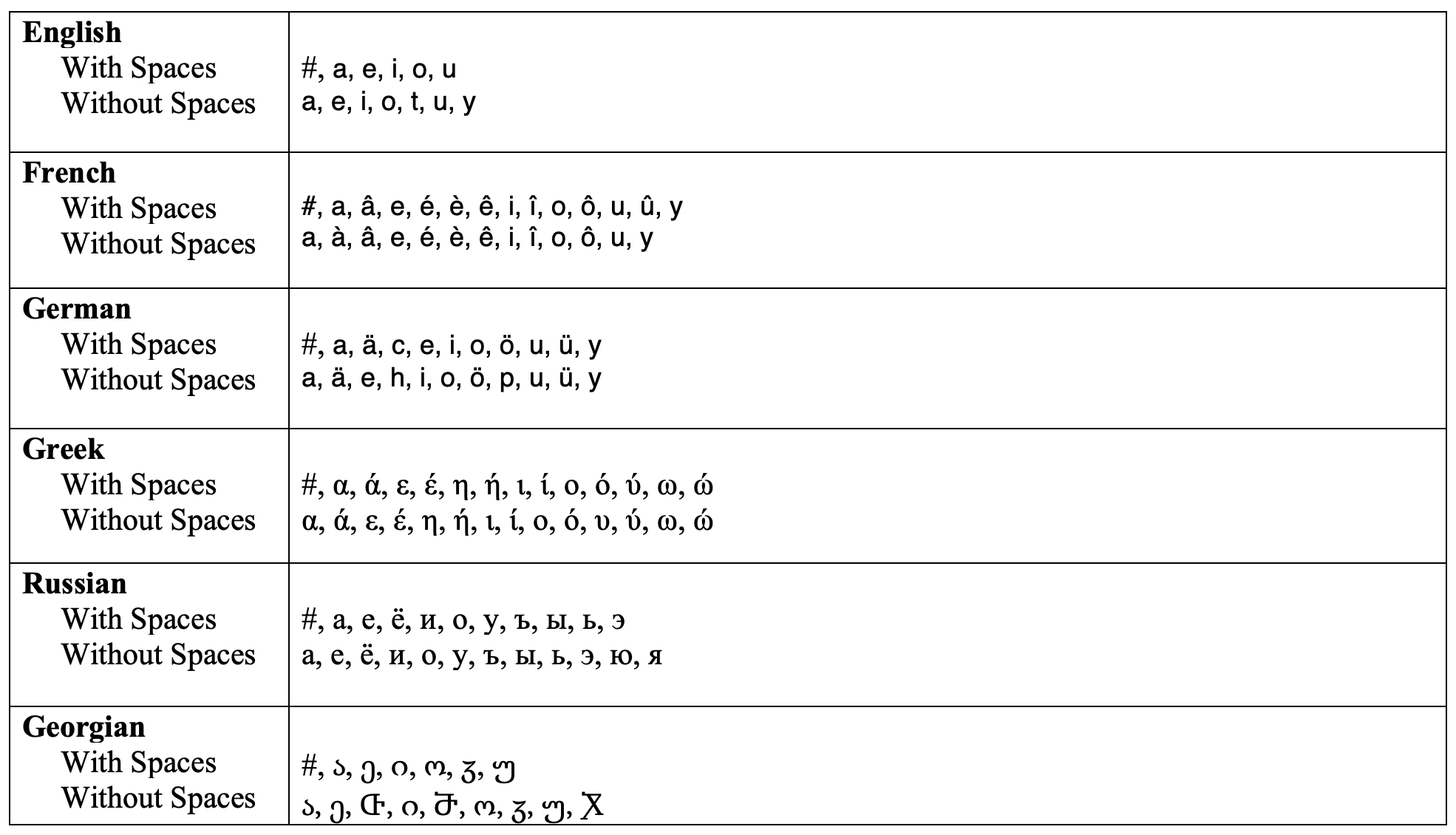}\centering
\caption{Results of the Sukhotin Algorithm for Wikipedia Samples in the Latin alphabet (English, French, German), Greek alphabet (Greek), and Georgian alphabet (Georgian).}
\label{wiki-alpha-sukh}
\end{figure}

\begin{figure}[h!]
\includegraphics[scale=0.4,width=\linewidth]{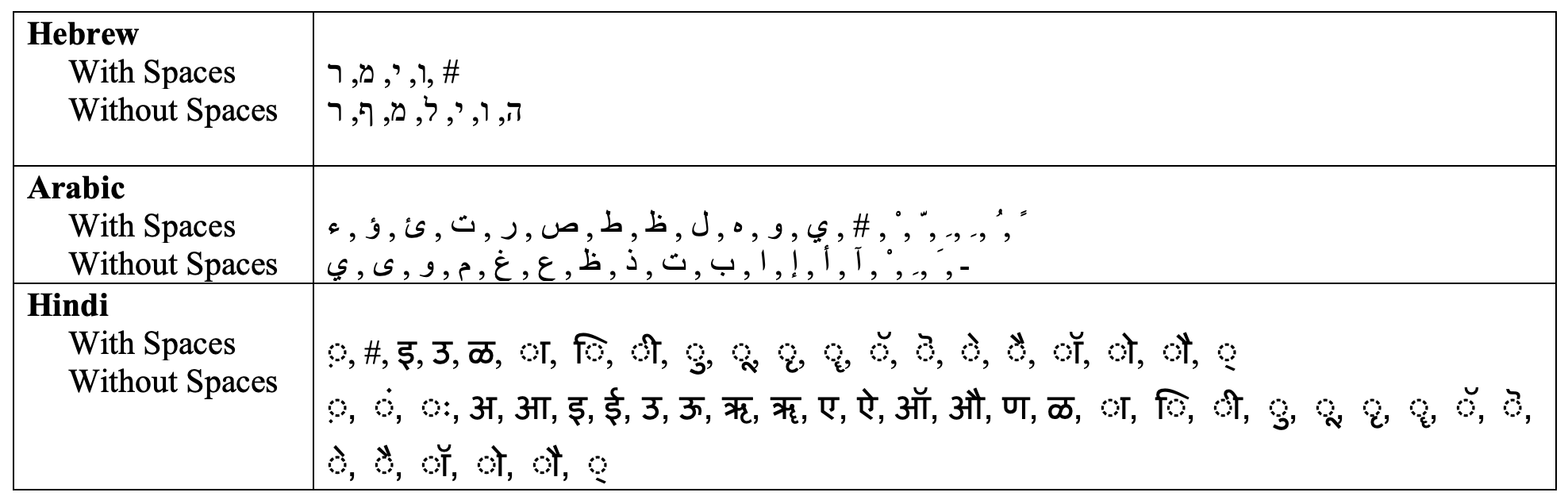}\centering
\caption{Results of the Sukhotin Algorithm for Wikipedia Samples in the Hebrew and Arabic abjads and the Devanagari abugida (Hindi).}
\label{wiki-script-sukh}
\end{figure}

Figure~\ref{wiki-alpha-sukh} shows the results for the Wikipedia samples of English, French, German, Greek, Russian, and Georgian, which are written in the Latin, Greek, Cyrillic, and Georgian alphabets. Running the algorithm with spaces does a fairly good job of picking out the vowels exhaustively and exclusively. Without spaces, the algorithm picks out \textit{t} in English, which is the first letter of the most common word. German also includes the characters \textit{c} and \textit{h}, which are commonly used in digraphs. For Georgian, the results are mixed: including spaces, the algorithm correctly identifies the vowels but includes some consonants as well, and it does worse if spaces are excluded. 

Figure~\ref{wiki-script-sukh} shows the results of the Sukhotin algorithm for Wikipedia samples of languages in three non-alphabetic scripts: Hebrew, Arabic, and Devanagari (run on the Hindi sample). Hebrew and Arabic are abjads which do not include vowels.\footnote{For Hebrew and Arabic the ordering of results is given from right to left to reflect the order in which these characters are read. Devanagari is ordered from left to right.} The algorithm identifies some consonants which double as vowels, but otherwise has a tendency to identify the final or initial forms of consonants, particularly if spaces are excluded. In Arabic the \textit{harakat} vowel marking is identified first. The algorithm is somewhat successful for Hindi, written in an abugida (alphasyllabary). Without spaces, it initially identifies the freestanding forms of the vowels, and then the separate combining forms of the vowels. With spaces, it only includes the combining forms.
\begin{figure}[p]
\includegraphics[scale=0.4,width=\linewidth]{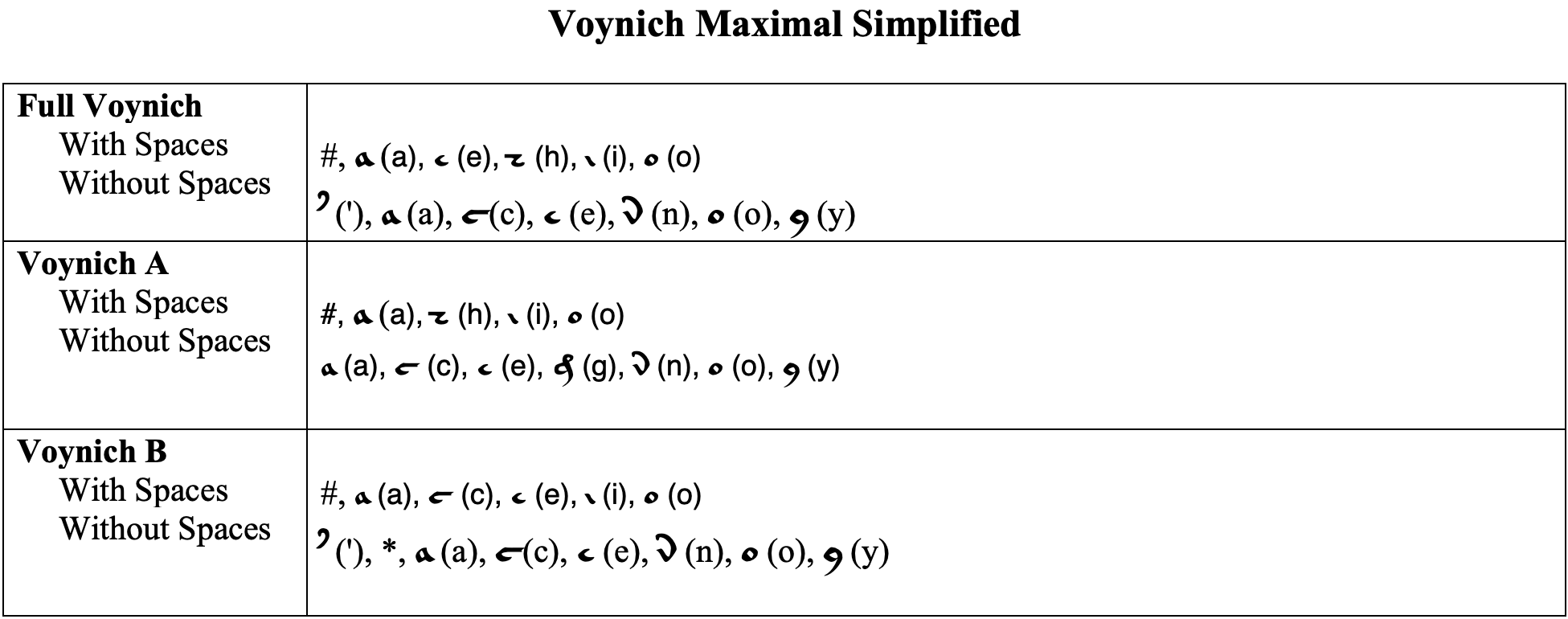}\centering

\includegraphics[scale=0.4,width=\linewidth]{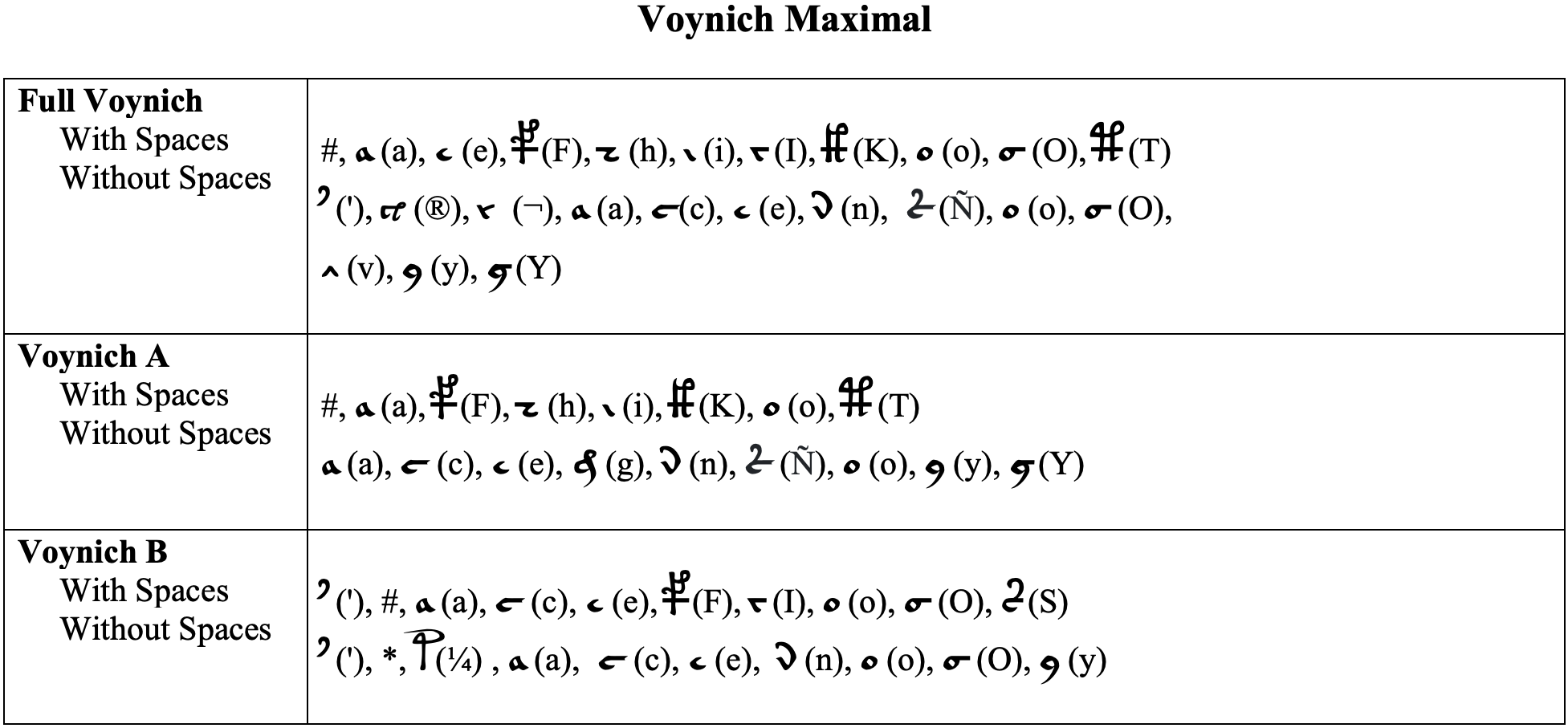}\centering

\includegraphics[scale=0.4,width=\linewidth]{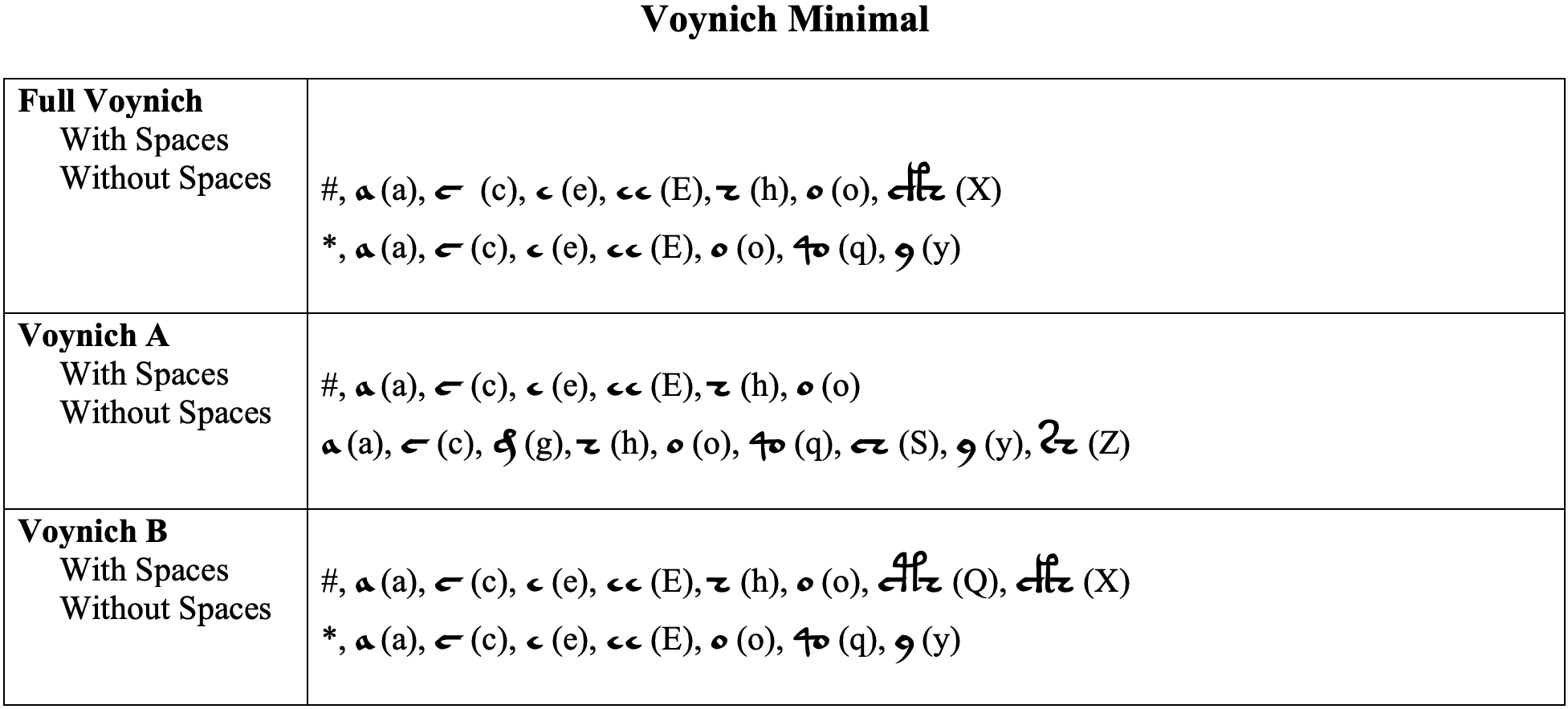}\centering
\caption{Results of the Sukhotin Algorithm for Voynichese in the Maximal Simplified, Full Maximal, and Minimal transcriptions.}
\label{voy-sukh}
\end{figure}

Figure~\ref{voy-sukh} shows the results of the Sukhotin algorithm for Full Voynich, A, and B in each of the three transcriptions. In the Maximal Simplified transcription, the algorithm identifies common characters which are found primarily in the middle of words and which bear a resemblance to Latin script vowels (\textit{a}, \textit{c}, \textit{h}, \textit{i}, and \textit{o}). The only difference between A and B is whether \textit{c} and \textit{h} are included. Without spaces, the algorithm additionally identifies word-finals: \textit{n}, \textit{y}, and \textit{g}. For Voynich B it also identifies the rare character symbol, and the \textit{f} gallows character which marks the beginning of a paragraph.

In the full Maximal transcription, the algorithm identifies the same characters as vowels, but the results are swamped by the addition of the ligature forms of consonants and vowels, and, when spaces are excluded, other extremely rare characters. Similarly, in the Minimal transcription the same characters are identified as vowels, with the exception of \textit{i} and the inclusion of the digraph \textit{ee}. Characters which are only found at the beginning of the words are also included: \textit{q}, \textit{X}, \textit{Q}, and \textit{S} and \textit{C}.

The Sukhotin algorithm is based upon the observation that vowels are more likely to be adjacent to consonants than to other vowels \citep{guy1991vowel}. This generalization is ultimately a reflection of universal properties of sonority in the world's languages, but its validity for any one language is subject to phonotactics (language-particular rules about possible phoneme sequences) and the peculiarities of the script. In addition to identifying vowels, it tends to identify rare characters and word-initial/final characters.

\bibliography{Voynich}

\begin{thebibliography}{}

\bibitem[Amancio et~al., 2013]{amancio2013probing}
Amancio, D.~R., Altmann, E.~G., Rybski, D., Oliveira~Jr, O.~N., and Costa, L.
  d.~F. (2013).
\newblock Probing the statistical properties of unknown texts: application to
  the voynich manuscript.
\newblock {\em PLoS One}, 8(7).

\bibitem[Barlow, 1986]{barlow1986}
Barlow, M. (1986).
\newblock The {Voynich} {Manuscript} - {By} {Voynich}?
\newblock {\em Cryptologia}, 10(4):210--216.
\newblock Citation Key Alias: barlow1986a.

\bibitem[Bennett, 1976]{bennett1976scientific}
Bennett, W.~R. (1976).
\newblock {\em {Scientific and engineering problem-solving with the computer}},
  chapter 4. Language, pages 103--198.
\newblock Prentice Hall series in automatic computation. Prentice Hall,
  Englewood Cliffs, N.J.

\bibitem[Bowern and Lindemann, 2020]{bowernlindemann20}
Bowern, C. and Lindemann, L. (2020).
\newblock {The Linguistics of the Voynich Manuscript}.
\newblock {\em Annual Review of Linguistics}.

\bibitem[Currier, 1976]{currier1976papers}
Currier, P. (1976).
\newblock {Papers on the Voynich Manuscript}.
\newblock In D'Imperio, M., editor, {\em New Research on the Voynich
  Manuscript}, Washington, DC.

\bibitem[Davis, 2020]{davis20}
Davis, L.~F. (2020).
\newblock How many glyphs and how many scribes: Digital paleography and the
  voynich manuscript.
\newblock {\em Manuscript Studies}, V(1):162--178.

\bibitem[Guy, 1991a]{guy1991statistical}
Guy, J.~B. (1991a).
\newblock {Statistical properties of two folios of the Voynich Manuscript}.
\newblock {\em Cryptologia}, 15(3):207--218.

\bibitem[Guy, 1991b]{guy1991vowel}
Guy, J.~B. (1991b).
\newblock {Vowel identification: an old (but good) algorithm}.
\newblock {\em Cryptologia}, 15(3):258--262.

\bibitem[Landini, 2001]{landini2001evidence}
Landini, G. (2001).
\newblock {Evidence of linguistic structure in the Voynich manuscript using
  spectral analysis}.
\newblock {\em Cryptologia}, 25(4):275--295.

\bibitem[Reddy and Knight, 2011]{reddy2011we}
Reddy, S. and Knight, K. (2011).
\newblock {What we know about the Voynich manuscript}.
\newblock In {\em Proceedings of the 5th ACL-HLT workshop on language
  technology for cultural heritage, social sciences, and humanities}, pages
  78--86. Association for Computational Linguistics.

\bibitem[Rugg, 2004]{rugg2004elegant}
Rugg, G. (2004).
\newblock An elegant hoax? a possible solution to the voynich manuscript.
\newblock {\em Cryptologia}, 28(1):31--46.

\bibitem[Stallings, 1998]{stallings1998understanding}
Stallings, D. (1998).
\newblock {Understanding the second-order entropies of Voynich text}.
\newblock \url{http://ixoloxi.com/voynich/mbpaper.htm}.

\bibitem[Sukhotin, 1962]{sukhotin1962eksperimental}
Sukhotin, B. (1962).
\newblock {Eksperimental’noe vydelenie klassov bukv s pomoscju evm}.
\newblock {\em Problemy strukturnoj lingvistiki}, 234:189--206.

\bibitem[Timm and Schinner, 2020]{TimmSchinner2020}
Timm, T. and Schinner, A. (2020).
\newblock {A possible generating algorithm of the Voynich manuscript}.
\newblock {\em Cryptologia}, 44(1):1--19.

\bibitem[Timm and Schinner, 2021]{TimmSchinner2021}
Timm, T. and Schinner, A. (2021).
\newblock Review of the linguistics of the voynich manuscript by claire bowern
  and luke lindemann.
\newblock {\em Cryptologia}, 0(0):1–5.

\bibitem[Zandbergen, 2010]{zandbergen2010}
Zandbergen, R. (2010).
\newblock Voynich ms.
\newblock \url{http://www.voynich.nu/index.html}.
\newblock [Online; accessed 24-April-2018].

\bibitem[Zandbergen, 2021]{Zandbergen2021}
Zandbergen, R. (2021).
\newblock The cardan grille approach to the voynich ms taken to the next level.
\newblock {\em arXiv:2104.12548 [cs]}.
\newblock arXiv: 2104.12548.

\end{thebibliography}

\end{document}